\documentclass[manuscript,screen]{acmart}

\AtBeginDocument{%
  \providecommand\BibTeX{{%
    \normalfont B\kern-0.5em{\scshape i\kern-0.25em b}\kern-0.8em\TeX}}}

\usepackage{url}
\usepackage{graphicx}
\usepackage{amsmath,amsfonts,xfrac}
\usepackage{algorithmic,algorithm}
\usepackage{subfig}
\usepackage{booktabs,multirow}
\usepackage{color}
\usepackage{enumerate}

\begin{document}

\title{FedGL: Federated Graph Learning Framework with Global Self-Supervision}

\author{Chuan Chen}
\email{chenchuan@mail.sysu.edu.cn}
\author{Weibo Hu}
\email{huwb7@mail2.sysu.edu.cn}
\affiliation{%
  \institution{Sun Yat-sen University}
  \city{Guangzhou}
  \country{China}
}

\author{Ziyue Xu}
\email{1245570597@qq.com}
\affiliation{%
  \institution{Xi'an Jiaotong University}
  \city{Xian}
  \country{China}
}

\author{Zibin Zheng}
\email{zhzibin@mail.sysu.edu.cn}
\affiliation{%
  \institution{Sun Yat-sen University}
  \city{Guangzhou}
  \country{China}
}

\renewcommand{\shortauthors}{Chen and Hu, et al.}

\begin{abstract}
Graph data are ubiquitous in the real world. Graph learning (GL) tries to mine and analyze graph data so that valuable information can be discovered. Existing GL methods are designed for centralized scenarios. However, in practical scenarios, graph data are usually distributed in different organizations, i.e., the curse of isolated data islands. To address this problem, we incorporate federated learning into GL and propose a general \underline{Fed}erated \underline{G}raph \underline{L}earning framework FedGL, which is capable of obtaining a high-quality global graph model while protecting data privacy by discovering the global self-supervision information during the federated training. Concretely, we propose to upload the prediction results and node embeddings to the server for discovering the global pseudo label and global pseudo graph, which are distributed to each client to enrich the training labels and complement the graph structure respectively, thereby improving the quality of each local model. Moreover, the global self-supervision enables the information of each client to flow and share in a privacy-preserving manner, thus alleviating the \textit{heterogeneity} and utilizing the \textit{complementarity} of graph data among different clients. Finally, experimental results show that FedGL significantly outperforms baselines on four widely used graph datasets.
\end{abstract}

\maketitle

\section{Introduction}\label{Intro}
In the real world, graph data are ubiquitous, such as social networks, financial transaction networks, and biological networks. Graph learning (GL) aims to dig out the valuable information from the graph data by using various graph model, including graph regularization \cite{smola2003kernels}, graph embedding \cite{cai2018comprehensive}, graph neural networks \cite{wu2020comprehensive}, etc. GL has boosted various applications, such as community detection \cite{fortunato2010community}, personalized recommendation \cite{fouss2007random}, and fraud detection \cite{noble2003graph}. Existing GL methods are designed for a centralized learning scenario, that is, centralized graph data storage and centralized model training. However, in most industries, graph data exists in the form of isolated islands \cite{yang2019federated}, i.e., distributed in different organizations or institutions. Considering a practical problem in the financial industry, each bank owns the customer information, transaction network, and default history. Banks have some common customers. There is a crucial demand that banks hope to collaborate to conduct a comprehensive credit assessment on their customers and identify a common industry blacklist. An intuitive idea is to collect the graph data together and merge them into a large graph, and then feed it to existing GL methods. However, it is almost impossible to collect the graph data from institutions scattered around the country due to privacy security \footnote{On May 25, 2018, the European Union promulgated the EU General Data Protection Regulation (GDPR) to protect users' personal privacy and data security: \url{https://gdpr-info.eu}} and industry competition. Therefore, how to collaborate the graph data distributed on different organizations to train a high-quality graph model without compromising data privacy is an open and crucial problem.

\begin{figure}[t]
  \centering
  \includegraphics[width=.7\textwidth]{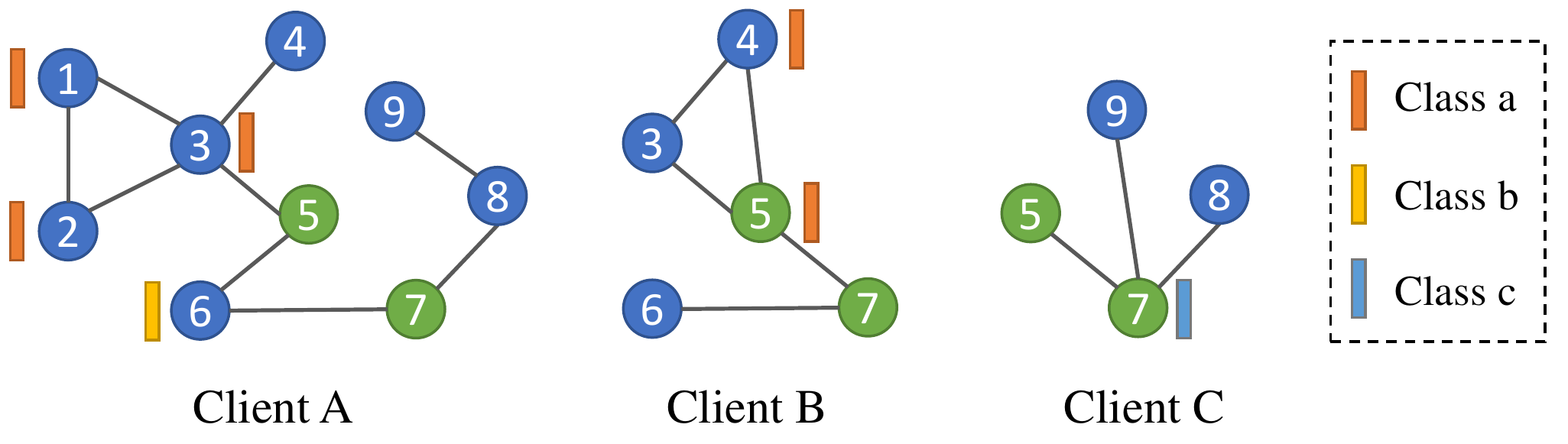}
  \caption{An example of \textit{heterogeneity} and \textit{complementarity} of graph data between different clients in federated scenarios. Client A/B/C has 9/5/4 nodes, 9/5/3 edges, and 4/2/1 labels. The three clients vary in the number of nodes, edges, and labels, and this difference becomes more obvious on large-scale graphs in the real world. This is so-called the \textit{heterogeneity}. On the other hand, there are overlapping nodes among different clients. In client A, nodes 1, 2, and 3 are closely connected and have the same label a. Nodes 4 and 5 are not connected, and node 5 has no label. While in client B, nodes 4 and 5 are connected with the same label a. Therefore, client A can infer that nodes 4 and 5 should also be closely connected and predicted to have label a by combining the information of client B. This is so-called the \textit{complementarity}.}
  \label{fig_example}
\end{figure}

Federated learning \cite{mcmahan2017communication} is an emerging technique that trains machine learning models based on datasets distributed across multiple devices while preventing data leakage. The key idea is to leave the data on the devices (or clients) and train a shared global model by uploading and aggregating the local updates (e.g., gradients or model parameters) yielded by clients to a central server. Commencing with the first and most famous federated learning algorithm FedAvg \cite{mcmahan2017communication}, many improved works have been proposed to address various problems of federated learning, including reducing the communication cost \cite{konevcny2016federated,tao2018esgd,liu2019edge}, overcoming the systems heterogeneity \cite{sprague2018asynchronous,nishio2019client,yoshida2020hybrid}, overcoming the statistical heterogeneity \cite{smith2017federated,chen2018federated,khodak2019adaptive}, further protecting data privacy \cite{hao2019towards,bagdasaryan2020backdoor,li2020blockchain}. Intuitively, incorporating the framework of federated learning into GL is a promising solution for the above demand. However, existing federated learning related research and applications are mainly focused on processing structured data, such as image and text data \cite{yang2019federated,lim2020federated,zhang2021survey}, and very little work is focused on graph data. There are serval unpublished works \cite{zhou2020privacy,wang2020graphfl} that attempt to develop a federated framework for graph data. \cite{zhou2020privacy} assumes that clients have the same nodes, different features and edges, and only one client has labels. \cite{wang2020graphfl} assumes that clients have the same nodes, features and edges, different labels. Since they make different scenario assumptions, it is difficult to generalize them to address general federated GL problems. Besides, in the real world, it is commonly observed that clients have different nodes, features, edges, and labels, and have some overlapping nodes.

In general, there are two severe challenges for federated GL problems. (1) \textit{Heterogeneity}: Graph data distributed on different clients are essentially and potentially highly Non-Independent Identically Distributed (Non-IID). In this situation, the local model trained by each client using its graph data could also has large differences, leading to an unsatisfactory global model after aggregation. (2) \textit{Complementarity}: Graph data distributed on different clients usually contain complementary information due to the overlapping nodes. For these overlapping nodes, the graph structure on each client is not comprehensive due to the inability to share and aggregate data. Fig. \ref{fig_example} is an illustration of \textit{heterogeneity} and \textit{complementarity}.

The above two challenges lead to two motivations during the federated training. (1) How to alleviate the \textit{heterogeneity} of graph data between clients, so that server can aggregate and obtain a high-quality global graph model? (2) How to utilize the graph structure on each client to complement each other, so as to help each client learn a better local graph model?

\begin{figure}[t]
  \centering
  \includegraphics[width=.6\textwidth]{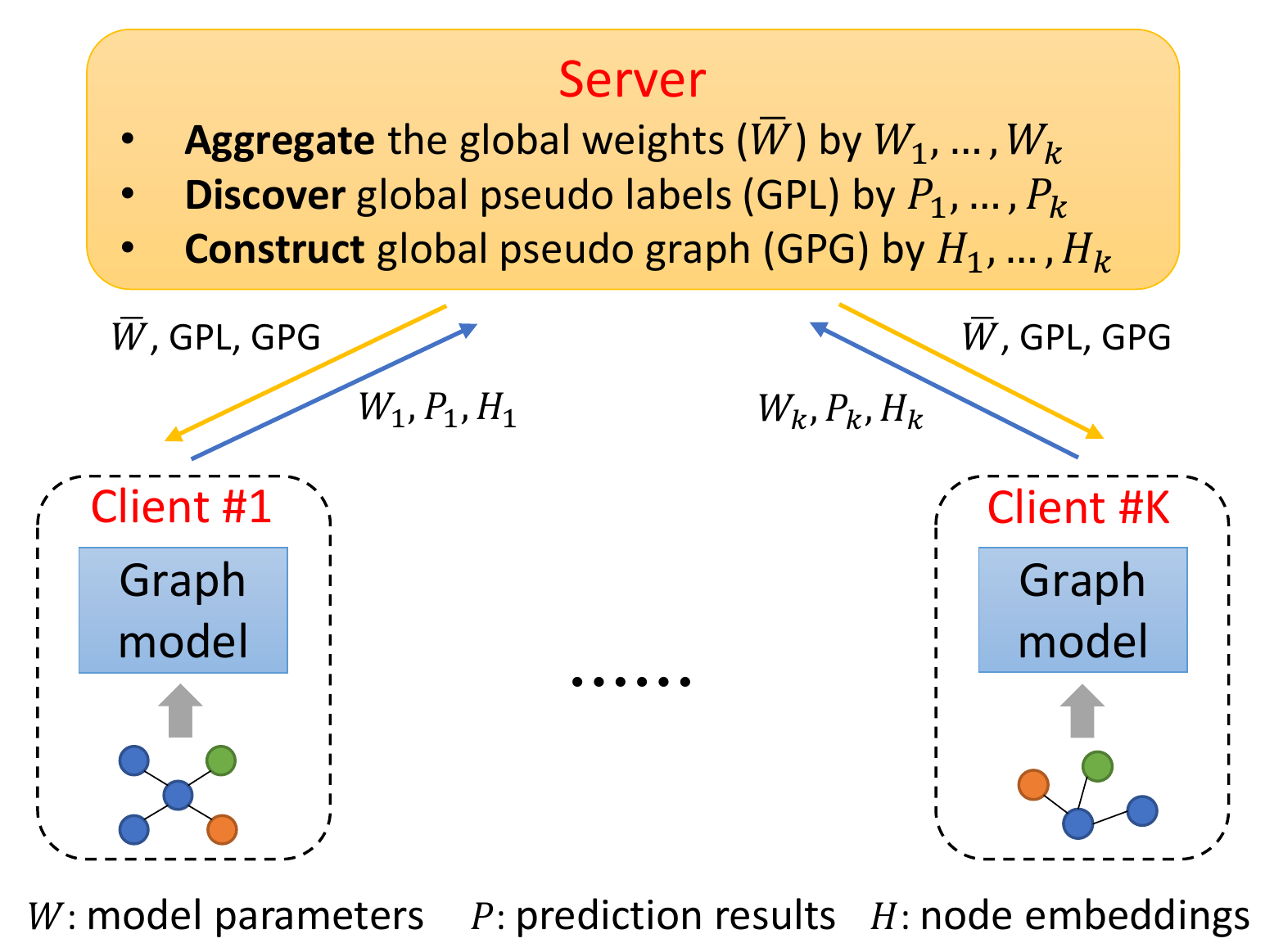}
  \caption{The general framework of the proposed FedGL.}
  \label{fig_framework}
\end{figure}

In this paper, we propose a general \underline{Fed}erated \underline{G}raph \underline{L}earning framework FedGL, which is capable of learning a high-quality graph model by discovering and exploiting the global self-supervision information to effectively deal with the \textit{heterogeneity} and \textit{complementarity}. The general framework is shown in Fig. \ref{fig_framework}. There are $K$ clients and one server. Each client locally trains a graph model (local model) by using its graph data. Existing federated learning methods upload the gradients or model parameters to the server which aggregates them to obtain the global model and distributes it for the next iteration. For the proposed FedGL, we additionally upload the prediction results and node embeddings of each client to the server for discovering the global self-supervision information, including global pseudo label and global pseudo graph, thereby alleviating \textit{heterogeneity} and utilizing \textit{complementarity}. Concretely, we propose to discover the global pseudo label by firstly fusing the prediction results and then selecting the results of unlabeled nodes with high confidence. Server distributes the discovered pseudo label to each client to enrich the training labels, thereby improving the quality of each local model. The process of global pseudo label discovery enables the information of each client to flow and be integrated in a privacy-preserving manner, thus mitigating the \textit{heterogeneity}. Besides, we propose to construct a global pseudo graph by firstly fusing the node embeddings from each client and then reconstructing the whole adjacent matrix. Server distributes the constructed global pseudo graph to each client to complement the graph structure, thus further improving each local model and leading to a high-quality global model. The process of global pseudo graph discovery enables the graph structure of each client to be collected and shared in a privacy-preserving manner, fully utilizing the \textit{complementarity}.

As a general federated framework for the distributed graph data, FedGL is not restricted to any specific graph model. In this work, we adopt graph neural networks \cite{wu2020comprehensive} as the graph model, which have presented state-of-the-art performance on graph-based tasks. Finally, we conduct extensive experiments on four widely used graph datasets. Experimental results show that FedGL significantly outperforms the centralized method, simple federated method, and local method, which fully verifies the effectiveness of FedGL.

Our main contributions are summarized as follows:

\begin{itemize}
  \item We propose a general federated graph learning framework FedGL that can collaborate the graph data distributed on different clients to train a high-quality graph model while protecting data privacy. FedGL provides a feasible solution for cooperative graph learning between organizations in the real world.
  \item We propose to additionally upload the prediction results and node embeddings to the server for discovering the global self-supervision information, including global pseudo label and global pseudo graph, which are distributed to each client to enrich the training labels and complement the graph structure respectively, thereby improving the quality of each local model and obtaining a high-quality global model.
  \item The proposed global pseudo label dexterously enables the information of each client to flow and be integrated in a privacy-preserving way. Especially high-quality clients can give aid to low-quality clients through global pseudo labels, thereby alleviating the \textit{heterogeneity}. The global pseudo graph subtly enables the graph structure of each client to be collected and shared in a privacy-preserving way, making full use of the \textit{complementarity}.
  \item We choose the graph neural network as the graph model and conduct extensive experiments on four widely used graph datasets. Experimental results show that FedGL significantly outperforms the centralized method, simple federated method, and local method, which fully verifies the effectiveness of FedGL. Besides, plentiful parameter and ablation experiments verify the stability and robustness of FedGL.
\end{itemize}

The rest of this paper is organized as the followings. In Section \ref{RW}, we review the related work on federated learning, graph learning, and self-supervised learning. Section \ref{Pre} introduces the theoretical knowledge of graph neural networks and federated learning. In Section \ref{Method}, we detail the proposed framework FedGL. Extensive experimental results and analyses are presented in Section \ref{Exp} followed by the conclusion and future work in Section \ref{Con}.

\section{Related Work}\label{RW}
\subsection{Federated Learning}
Federated learning is an emerging decentralized learning technique that can collaboratively train multiple models with the training data distributed on different devices (also called clients) and maintain a shared global model on a server by aggregating locally computed updates \cite{konevcny2016federated,mcmahan2017communication}. It well solves the data isolated island problem and protects data privacy. Specifically, each client trains a local model and computes the local update based on its data. The local updates are uploaded to a server that aggregates them to update the global model. The updated model is further distributed to each client to continue the next round of training. This process is iteratively executed until the global model converges. For example, FedAvg \cite{mcmahan2017communication}, the most representative federated learning method, uploads the model parameters and averages them to obtain the global model. Up to now, many improvement efforts have been devoted to address various problems of federated learning, including reducing the communication cost \cite{konevcny2016federated,tao2018esgd,liu2019edge}, overcoming the systems heterogeneity \cite{sprague2018asynchronous,nishio2019client,yoshida2020hybrid}, overcoming the statistical heterogeneity \cite{smith2017federated,chen2018federated,khodak2019adaptive}, further protecting data privacy \cite{hao2019towards,bagdasaryan2020backdoor,li2020blockchain}, etc. Federated learning has a promising application in finance \cite{long2020federated}, healthcare \cite{xu2021federated}, mobile edge networks \cite{lim2020federated}, and many other industries \cite{lim2020federated}, in which data cannot be directly aggregated for training machine learning models due to factors such as intellectual property rights, privacy protection, and data security. Besides, the framework of federated learning also has been effectively integrated into various techniques, such as multi-task learning \cite{smith2017federated}, transfer learning \cite{liu2020secure}, and reinforcement learning \cite{zhuo2019federated}.

\subsection{Graph Learning}
Graph learning (GL) aims to mine and analyze the graph data, thus obtaining lots of valuable information. Due to the complexity of graph data, it is often necessary to firstly transform it into structured data, so graph embedding \cite{cai2018comprehensive,huang2020nonuniform} that embeds each node to a  low-dimensional dense vector (node embedding) remains the most important technique of GL. These node embeddings can be readily applied to the downstream tasks, such as node classification, node clustering, link prediction, etc. Up to now, extensive graph embedding methods are proposed \cite{zhang2018network}, such as the famous methods DeepWalk \cite{perozzi2014deepwalk} and node2vec \cite{grover2016node2vec}. Recently, graph neural networks (GNNs) \cite{wu2020comprehensive}, an emerging type of neural network model on graphs, have presented state-of-the-art performance on various graph-based tasks. It integrates the graph topology, node attributes, and neural network to jointly learn node embeddings. Meanwhile, the downstream tasks and node labels are added to the model for end-to-end training. As the most important branch of GNNs, graph convolutional networks (GCNs) borrow ideas from convolutional neural networks (CNNs) and redefine the convolution operation for graph data. The pioneering work on GCNs is proposed in \cite{bruna2013spectral}, which defines graph convolution by introducing spectral filters from the perspective of graph signal processing \cite{shuman2012emerging}. Since then, there have been increasing improvements, approximations, and extensions on spectral-based GCNs \cite{henaff2015deep,defferrard2016convolutional,kipf2016semi}. Among them, the most famous is the improved version proposed by \cite{kipf2016semi}, which simplifies the spectral graph convolution by only using the first-order neighbors. By stacking multiple convolutional layers, this GCN can encode both graph structure and node features to be useful for the node classification task. Since the spectral-based GCNs require the whole graph as the inputs and cannot scale to large graphs, spatial-based GCNs have been proposed, including GraphSAGE \cite{hamilton2017inductive}, GAT \cite{velivckovic2017graph}, and LGCN \cite{gao2018large}. These methods define graph convolution via directly aggregating information from neighbors. By combining with sampling and subgraph training strategies, the computation efficiency can be improved effectively. After that, further improvements mainly focus on convolution function and mechanism \cite{velivckovic2017graph,thekumparampil2018attention,li2015gated}, expressive power and depth of network \cite{xu2018representation,li2019deepgcns,liu2020towards}, large-scale and training efficiency \cite{chen2018fastgcn,chiang2019cluster,jia2020redundancy}, robustness \cite{zugner2020certifiable,hu2021robust}, etc.

\subsection{Self-Supervised Learning}
Self-supervised learning (SSL) originates from the field of computer vision and aims to learn visual features from a large number of unlabeled images or videos without using any manually annotated information \cite{jing2020self}. In recent years, SSL has gradually been used in the field of graph data learning \cite{you2020does,sun2020multi}. SSL can be roughly divided into two categories. One is to use pretext tasks, which usually do not require labels, thus the model can be trained in an unsupervised manner. For image data, common pretext tasks include image rotation, image clustering, image restoration, etc \cite{jing2020self}. For graph data, common pretext tasks include node clustering, link prediction, graph partitioning, etc \cite{you2020does}. Another is to discover pseudo labels \cite{lee2013pseudo,hu2021learning} and treat them as real labels to train models. Pseudo labels can be constructed based on source data, or discovered from prediction results. SSL can be used alone or used as a pre-training step \cite{jing2020self}. SSL also can be used as a regularization item to help the main task achieve better results \cite{wu2017semi,sun2020multi}.

\section{Preliminaries}\label{Pre}
\subsection{Notations}
A graph with $N$ nodes is denoted as $G = (V, E)$, where $V=\{v_1,...,v_N\}$ is the node set and $E \subseteq V \times V$ is the edge set. Let $A \in \mathbb{R}^{n \times n}$ represent the adjacency matrix. Let $D \in \mathbb{R}^{N \times N}$ denote the degree matrix, which is a diagonal matrix with $D_{ii}=\sum_{j}A_{ij}$. For an attributed graph, $X=\{x_1,...,x_N\}$ is the associated node feature matrix and $x_i \in \mathbb{R}^d$ denotes the feature vector of node $v_i$, where $d$ is the feature dimensionality. The node labels are represented as one-hot matrix $Y \in \mathbb{R}^{N \times C}$, where $C$ is the number of classes of the node, and $Y_{ij}=1$ if node $i $ belongs to class $j$, otherwise $Y_{ij}=0$.

\subsection{Graph Neural Networks}\label{GCN}
Although there are numerous variants of GNNs, in this paper, we mainly focus on the most general and representative one proposed in \cite{kipf2016semi}. For this GCN, the convolutional layer and layer-wise propagation rule are defined as
\begin{equation}
H^{(l+1)}=\sigma\big(\tilde{D}^{-\frac{1}{2}} \tilde{A} \tilde{D}^{-\frac{1}{2}} H^{(l)} W^{(l)}\big),
\end{equation}
where $\tilde{A}=A+I_N$ is the adjacency matrix with added self-connections. $I_N$ is the identity matrix, $\tilde{D}_{ii}=\sum_{j}\tilde{A}_{ij}$, and $W^{(l)}$ is the trainable weight matrix of layer $l$. $\sigma(\dots)$ is an activation function such as ReLU. $H^{(l)}$ is the latent representation matrix of layer $l$ and $H^{(0)}=X$, i.e., using the node feature matrix as input. It is worth noting that the information of $H$ is continually propagated through the immediate neighbors. Following \cite{kipf2016semi}, we consider a two layer GCN model to obtain the final node embeddings:
\begin{equation}\label{emb}
H = \hat{A} \, \sigma\big(\hat{A} X W^{(0)}\big) W^{(1)},
\end{equation}
where $\hat{A}=\tilde{D}^{-\frac{1}{2}} \tilde{A} \tilde{D}^{-\frac{1}{2}}$. Then, by inputting $H$ to the Softmax function, we can obtain the predicted class probability matrix:
\begin{equation}\label{pred}
P_{ij} = \frac{\text{e} ^{H_{ij}}}{\sum_{j=1}^{C}\text{e}^{H_{ij}}},
\end{equation}
where $P_{ij}$ indicates the probability of node $i$ belonging to class $j$. For semi-supervised node classification task, we compute the cross-entropy loss over the labeled samples:
\begin{equation}\label{L_gcn}
L_{\text{GCN}} = -\sum_{i \in V_L} \sum_{j=1}^{C} Y_{ij} \log P_{ij},
\end{equation}
where $V_L$ is the set of labeled nodes, $C$ is the number of classes, and $Y$ is the one-hot label matrix. Up to now, the neural network weights $W=\{W^{(0)}, W^{(1)}\}$ can be updated by back-propagation with the goal of minimizing Eq. \eqref{L_gcn}.

\subsection{Federated Learning Framework}
In the federated learning framework, there are mainly two types of entities, i.e., clients and server. Clients refer to the party that owns the data, which can be mobile edge devices or organizations. The model trained and stored on each client is called \emph{local model}. Server aggregates the local models uploaded by each client to obtain a \emph{global model}. Specifically, suppose there are $K$ clients, $D_k$ denotes the data owned by the client $k$, and $W_k$ denotes the local model trained on the client $k$. $W_G$ denotes the global model on the server. The training process of federated learning can be summarized as the following steps:
\begin{enumerate}
  \item \emph{Initialization}. Server determines the training tasks, hyper-parameters, initial model parameters $W_G^0$, etc., and distributes them to each client.
  \item \emph{Local model training}. Based on the global model $W_G^t$ in $t$-th iteration, that is, each client utilizes its local data for training and updates the local model parameters. For the client $k$, the update formula of the $t$-th iteration is
      \begin{equation}\label{local_model}
        W_k^t = W_k^t - \alpha \nabla L_{D_k}(W_k^t),
      \end{equation}
      where $\alpha$ is the learning rate, and the updated local model parameters are uploaded to the server.
  \item \emph{Global model update}. Server aggregates the model parameters uploaded by the clients to obtain the updated global model parameters $W_G^{t+1}$, and then sends them to each client. The commonly used weighted average aggregation formula is as follows:
      \begin{equation}\label{agg}
        W_G^{t+1} = \sum_{k=1}^{K} \frac{|D_k|}{\sum_{k=1}^{K} |D_k|} W_k^t,
      \end{equation}
      where $|D_k|$ represents the number of samples of the client $k$.
\end{enumerate}
The whole progress will repeat steps 2) and 3) until the global model converges or reaches the maximum number of iterations. In practical applications, the number of clients may be very large. Hence, in step 2), we can randomly select or designate some clients to participate in the training, thereby reducing the training time. In addition, we can also upload the model gradients instead of model parameters in step 2). If so, server aggregates the gradients and performs gradient descent to update the global model in step 3). After the federated training is completed, the final global model can be used for prediction. It is expected that its performance on the testing set is equivalent to the model that collects data from clients for centralized training.

\section{Federated Graph Learning Framework (FedGL)}\label{Method}
\subsection{Problem Definition}\label{problem_def}
In this work, we aim to propose a federated graph learning framework, which can collaborate the graph data distributed on different clients to train a high-quality graph model while protecting data privacy. Suppose there are $K$ clients, the graph data owned by the client $k$ is denoted as $G_k=(V_k,E_k)$, where the number of nodes is $N_k=|V_k|$. The adjacency matrix, node feature matrix, and label matrix are denoted as $A_k \in \mathbb{R}^{N_k \times N_k}$, $X_k \in \mathbb{R}^{N_k \times d}$, and $Y_k \in \mathbb{R}^ {N_k \times C}$, respectively. The total number of nodes of $K$ clients is denoted as $M=\sum_{i=k}^{K}N_k$. Note that there are certain differences in the number of nodes, graph structure, and label distribution of each client, but there are also some overlapping nodes between the clients. That is, for any graph $G_k$, there exists $k \neq t$, so that $V_k \bigcap V_t \neq \emptyset$. This setting is derived from the distribution of graph data in the real world, which is rational and practical. In such a scenario, we have clarified the following two goals:
\begin{itemize}
  \item \emph{Global goal}. It is expected that the global model achieves a promising performance on the global testing set. The global testing set is stored on the server and can be jointly determined by each client. The global goal is also the original intention of federated learning. It is expected that the data distributed on different clients could be collaborated to train a shared global model, whose performance can be close to the model that collects data from clients for centralized training.
  \item \emph{Local goal}. It is expected that the global model also achieves a promising performance on the local testing set. The local testing set refers to the testing set of each client. Under the local goal, it is expected that the performance of the global model is better than the model that trains independently by only using the data of each client.
\end{itemize}



\begin{figure}[t]
  \centering
  \includegraphics[width=.9\linewidth]{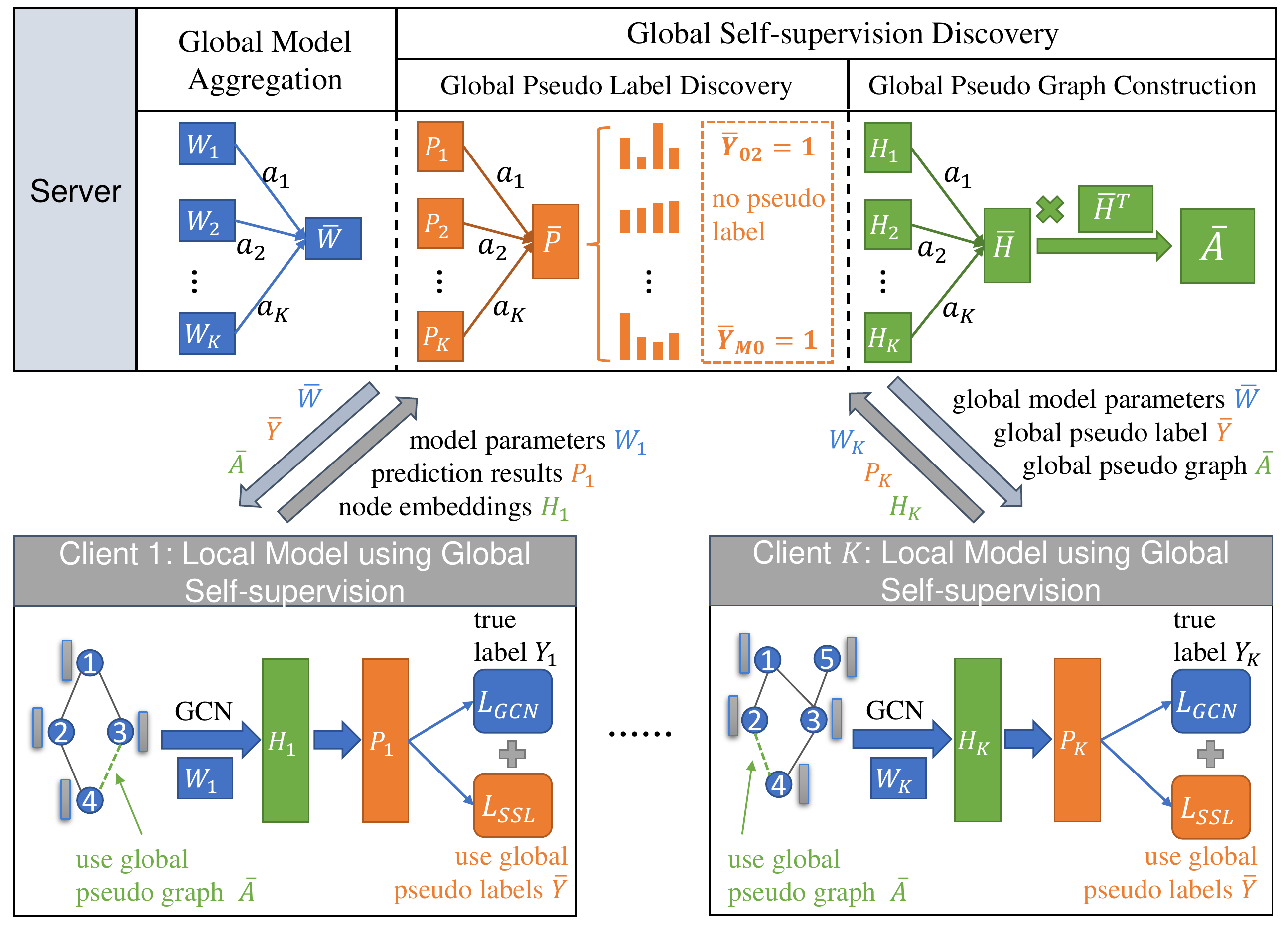}
  \caption{The detailed framework of FedGL. $L_{GCN}$ is the loss function of graph convolutional network, and $L_{SSL}$ is the loss function of self-supervised learning. $a_1,...,a_K$ is the aggregation weights. $\bar{Y}_{02}=1$ indicates that the discovered global pseudo label of node 1 is class 3.}
  \label{fig:framework}
\end{figure}

\subsection{The Framework of FedGL}
Based on the aforementioned motivations, we propose a general \underline{Fed}erated \underline{G}raph \underline{L}earning framework FedGL. The detailed framework is shown in Fig. \ref{fig:framework}. Overall, FedGL consists of two parts: 1) clients: local model using global self-supervision, 2) server: global model aggregation and global self-supervision discovery. The main ideas and workflow of FedGL are summarized as follows:
\begin{itemize}
  \item \textbf{Clients: local model training.} Each client uses its local graph data to train several rounds of GCN model, obtaining model parameters $W_k$, node embeddings $H_k$, and prediction results $P_k$, then upload them to the server. Note that $K$ clients train their local models in parallel.
  \item \textbf{Server: global model aggregation.} Server performs weighted average aggregation on the model parameters $W_1,...,W_K$ to obtain the global model $\bar{W}$, and then distributes $\bar{W}$ to each client.
  \item \textbf{Server: global self-supervision discovery.} Except aggregating local model parameters to obtain a global model, we propose to discover the global self-supervision information on the server, including global pseudo label and global pseudo graph, to deal with the \textit{heterogeneity} and \textit{complementarity}. Specifically, server firstly performs a weighted average fusion on the prediction results $P_1,...,P_K$ to obtain the global prediction result $\bar{P}$. Then, server selects the result with higher probability from the predicted probability vector of each row in $\bar{P}$ as the pseudo label of each node, which constitutes the one-hot matrix $\bar{Y}$ of the global pseudo label. Similarly, server performs weighted average fusion on the node embeddings $H_1,...,H_K$ to obtain the global node embedding $\bar{H}$. By multiplying $\bar{H}$ and its transpose, server can reconstruct the whole adjacency matrix, obtaining the weighted adjacency matrix $\bar{A}$ of the global pseudo graph. Server distributes the discovered global pseudo label $\bar{Y}$ and global pseudo graph $\bar{A}$ to each client to start the next round of training.
  \item \textbf{Clients: global self-supervision utilization.} The global pseudo label is regarded as the "real" label to enrich the relatively rare real training labels by constructing a self-supervised learning loss $L_{SSL}$ and adding it to the main task loss $L_{GCN}$ for joint optimization. The global pseudo graph is directly used to complement the incomplete graph structure. For example in Fig. \ref{fig:framework}, edge (3, 4) in client $1$ and edge (2, 4) in client $K$ have been well complemented. By exploiting the global pseudo label and global pseudo graph, the quality of each local model can be effectively improved, thereby leading to a high-quality global model.
\end{itemize}

The above first two steps are the standard federated learning processes, while the last two steps are the proposed global self-supervision discovery and utilization process, which are the core of FedGL. In summary, global pseudo label dexterously enables the information of each client to flow and be integrated in a privacy-preserving way, especially high-quality clients can give aid to low-quality clients through the global pseudo label, thereby alleviating the \textit{heterogeneity}. The global pseudo graph subtly enables the graph structure of each client to be collected and shared in a privacy-preserving way, making full use of the \textit{complementarity}. Meanwhile, global pseudo label and global pseudo graph are complementary to each other. Global pseudo label contributes to learning better node embeddings, thereby conducing to construct a more accurate pseudo graph, and global pseudo graph contributes to obtaining better prediction results, thus conducing to discover more reliable pseudo label.

\subsection{Local Model using Global Self-supervision}
For the local model of each client, we adopt the GCN model introduced in Section \ref{GCN}. At the same time, we use the discovered global pseudo label and global pseudo graph to improve the local model.

\subsubsection{Global Pseudo Label Utilization}
\begin{figure}[h]
  \centering
  \includegraphics[width=.6\linewidth]{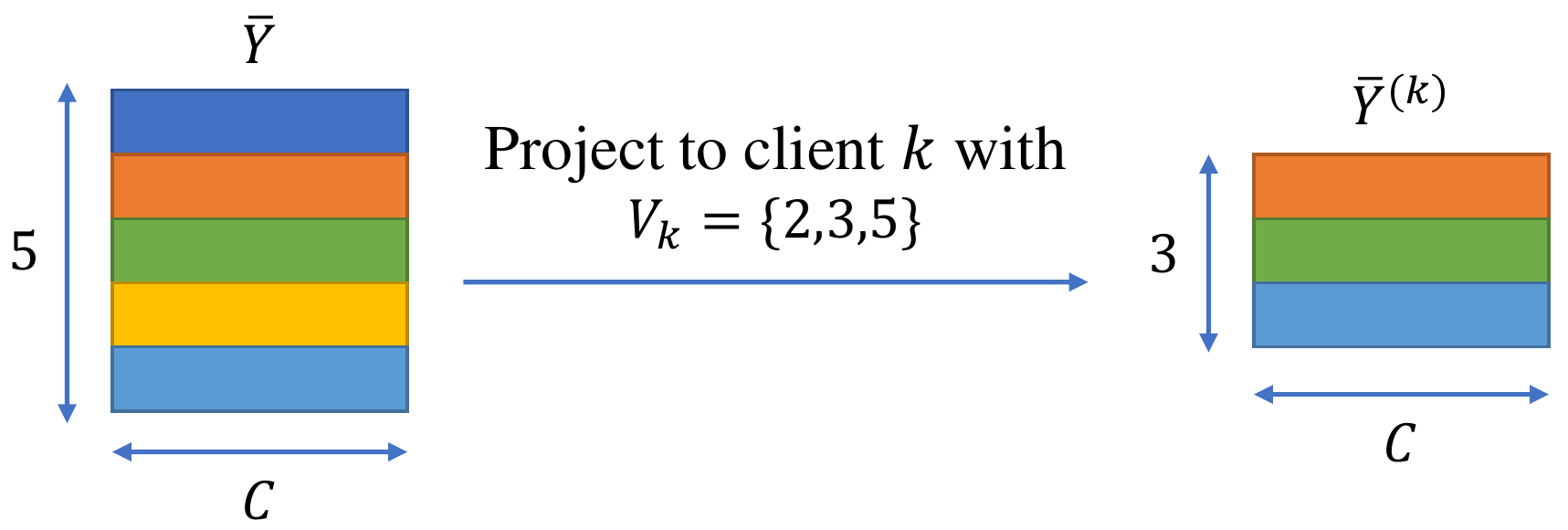}
  \caption{Example of global pseudo label $\bar{Y}$ projection to client $k$.}
  \label{fig:projection}
\end{figure}

Taking the client $k$ as an example, after receiving the global pseudo label $\bar{Y}$ that is represented as a one-hot matrix, it needs to project $\bar{Y}$ into its nodes to get $\bar{Y}^{(k)}$, since $\bar{Y}$ contains the pseudo label information of nodes on all clients. The projection process is shown in Fig. \ref{fig:projection}.

We then set all the rows of $\bar{Y}^{(k)}$ corresponding to the labeled nodes in the training set to 0. That is, we only reserve the rows corresponding to the unlabeled nodes, in which 1 represents the discovered pseudo label. We regard these pseudo labels as the "real" labels and add them into the training set to participate in training. Concretely, we also calculate a cross-entropy loss between the prediction result $P_k$ and $\bar{Y}^{(k)}$ on the client $k$, which is called self-supervised learning (SSL) loss:
\begin{equation}\label{L_ssl}
L_{SSL} = - \sum_{i \in V_{GPL}} \sum_{j=1}^{C} \bar{Y}_{ij}^{(k)} \log P_{k[ij]},
\end{equation}
where $V_{GPL}$ denotes a set of unlabeled nodes with global pseudo labels, and $P_{k[ij]}$ denotes the $i$-th row and $j$-th column of the prediction result of the client $k$. By adding Eq. \eqref{L_ssl} to the GCN loss in Eq. \eqref{L_gcn}, the final loss function of the local model on the client $k$ is formulated as
\begin{equation}\label{L_final}
\begin{aligned}
L & = L_{\text{GCN}} + \alpha L_{\text{SSL}} \\
  & = -\sum_{i \in V_L} \sum_{j=1}^{C} Y_{k[ij]} \log P_{k[ij]} - \alpha \sum_{i \in V_{PL}} \sum_{j=1}^{C} \bar{Y}_{ij}^{(k)} \log P_{k[ij]} ,
\end{aligned}
\end{equation}
where $\alpha$ is a coefficient to control the strength of self-supervised learning. By introducing the SSL loss, the performance of main task can be effectively enhanced, which has been proved in many studies \cite{lee2013pseudo,wu2017semi,sun2020multi,you2020does}. Moreover, in Eq. \eqref{L_final}, the pseudo label used to compute the SSL loss is global pseudo label, which is discovered by combining the prediction results of the local model on each client. Therefore, it is believed to be more reliable than the one discovered by the prediction results of only a single client.

\subsubsection{Global Pseudo Graph Utilization}
Taking the client $k$ as an example, after receiving the global pseudo graph $\bar{A}$ that is represented as a weighted adjacent matrix, it also need to project $\bar{A}$ into its nodes to get $\bar{A}^{(k)}$, since $\bar{A}$ contains the pseudo graph structure of nodes on all clients. The projection process is similar as Fig. \ref{fig:projection}. After obtaining the projection $\bar{A}^{(k)}$, it is easy to complement the original graph structure. The connection relationship that does not appear in the graph of client $k$ can be complemented by the global pseudo graph. The connection relationship that exists in the graph of client $k$ can be further strengthened by the global pseudo graph. The specific implementation is also very convenient. We only need to fuse $\bar{A}^{((k)}$ with the original normalized graph structure. Corresponding to GCN model, we can directly update $\hat{A}$ in Eq. \eqref{emb} as follows:
\begin{equation}\label{update_adj}
\hat{A}_k = \hat{A}_k + \beta \bar{D}^{-\frac{1}{2}} \bar{A}^{(k)} \bar{D}^ {-\frac{1}{2}},
\end{equation}
where $\beta$ is a coefficient to control the strength of the global pseudo graph to complement the graph structure, and $\bar{D}$ is the degree matrix of $\bar{A}^{(k)}$.

\subsection{Global Model}
In Eq. \eqref{emb}, the trainable parameters of GCN model are $W=\{W^{(0)},W^{(1)}\}$, where $W^ {(0)}$ is related to the initial feature dimensionality of the nodes, and $W^ {(1)}$ is related to the dimensionality of the hidden layer. Both of them are uncorrelated to the number of nodes. Therefore, as long as the initial feature dimensionality and hidden layer dimensionality of each client are consistent with each other, server can directly aggregate the model parameters uploaded by different clients. Following FedAvg \cite{mcmahan2017communication}, we employ the weighted average aggregation method to aggregate the model parameters of $K$ clients to obtain the global model:
\begin{equation}\label{global_model}
  \bar{W} = \sum_{k=1}^{K} \frac{N_k}{M} W_k,
\end{equation}
where $N_k$ is the number of nodes in the graph on the client $k$, and $M$ is the sum of the number of nodes in the graph of the $K$ clients, and $W_k$ is the model parameters of the client $k$. $\frac{N_k}{M}$ denotes the proportion of the data volume of each client, which is used to measure the importance of its model parameters in aggregation. Intuitively, the larger the amount of data the client has, the better the model it trains, and then it should dominate during the aggregation, i.e., assigning a larger weight. Furthermore, the weighting way can reduce the impact of the imbalance of the data volume of each client to some extent.

\subsection{Global Self-supervision Discovery}
Eq. \eqref{global_model} introduces the proportion of data volume of each client as the aggregation weight to treat each client differentially, which can alleviate the impact of data imbalance between clients to some extent. However, due to the \textit{heterogeneity} of graph data between clients, including graph structure and label distribution, the local models trained by different clients usually have uneven quality. It indicates that the global model obtained by weighted aggregation may still be unsatisfactory. Therefore, to obtain a high-quality global model, essentially, the quality of the local model on each client needs to be improved by alleviating the \textit{heterogeneity}. On the other hand, the graph structure of each client is complementary, due to the overlapping nodes. Making full use of the \textit{complementarity} is expected to further improve the quality of the local model.

In summary, the intrinsic reason for the \textit{heterogeneity} and \textit{complementarity} is that the graph data of each client cannot be collected to train a centralized model. Is there a privacy-preserving way to enable the information to flow and share between each client? In the framework of federated learning, server is naturally capable of accomplishing this task. Since the source data cannot be uploaded, in addition to the model parameters, other useful information can also be uploaded to the server for integration, and then distributed to each client, thus making the information flow. Therefore, we propose that clients additionally upload the prediction results and node embeddings to the server for discovering global self-supervision information, including global pseudo label and global pseudo graph. Server distributes them to each client to enrich the training labels and complement the graph structure, thereby alleviating the \textit{heterogeneity} and utilizing the \textit{complementarity}.

\subsubsection{Global Pseudo Label Discovery}
After receiving the prediction results uploaded by each client, similar as \eqref{global_model}, server performs a weighted average fusion on the results to obtain the global prediction results:
\begin{equation}\label{global_pred}
  \bar{P} = \sum_{k=1}^{K} \frac{N_k}{M} P_k,
\end{equation}
where $\frac{N_k}{M}$ is used as the fusion weight to measure the importance of the prediction results of the client $k$. Distinctly, a client with a larger amount of data has more abundant graph structure, more training labels, and more accurate prediction results. Therefore, assigning a larger weight can guarantee the fused global prediction results more accurate. At the same time, the nodes that have inferior predictions on some clients with a small amount of data can also become better by integrating the prediction results of other clients.

Based on $\bar{P}$, we try to discover pseudo labels for self-supervised learning, which has been proven to be effective in the learning of image and graph data \cite{lee2013pseudo,wu2017semi,sun2020multi,hu2021learning}. Concretely, we unearth these high-confidence prediction results from $\bar{P}$ and take out the predicted labels, thus obtaining the global pseudo labels. For the prediction result vector $\bar{P}_i$ of the $i$-th node in $\bar{P}$, if its predicted probability of a certain class is higher than a certain threshold, then it is selected as a pseudo label:
\begin{equation}\label{pseudo_label}
\begin{aligned}
\bar{Y}_{ij} =
& \begin{cases}
    1, \quad &\mbox{if} \; (\bar{P}_{ij}> \lambda) \; \text{and} \; (j = \arg\max \bar{P}_i), \\
    0, \quad &\text{otherwise},
\end{cases}\\
\end{aligned}
\end{equation}
where $\bar{Y}$ is the one-hot matrix of the global pseudo label, and $\lambda \in [0, 1)$ is the confidence threshold for determining the pseudo label. A small value of $\lambda$ means that a little more reliable prediction results could be selected as pseudo labels, so the number of pseudo labels is relatively large. That is, $\bar{Y}$ has more rows containing 1. A large value of $\lambda$ means that only enough reliable prediction results could be selected as pseudo labels, so the number of pseudo labels is relatively small. That is, $\bar{Y}$ has more rows with all 0s.

\subsubsection{Global Pseudo Graph Construction}
After receiving the node embeddings uploaded by each client, similar as \eqref{global_model}, server performs a weighted fusion on the results to obtain the global node embeddings:
\begin{equation}\label{global_emb}
  \bar{H} = \sum_{k=1}^{K} \frac{N_k}{M} H_k,
\end{equation}
where $\frac{N_k}{M}$ is also used as the fusion weight to measure the importance of the  node embeddings of the client $k$. Recall that the idea of graph embedding, each node is mapped into a low-dimensional dense vector by preserving the topological structure information of the graph as much as possible. The original tightly connected nodes still keep close in the vector space. Hence, by computing the distance or similarity between node vectors, it can approximately reconstruct the original graph structure. This idea is also commonly used in graph auto-encoder \cite{kipf2016variational,pan2018adversarially} or feature-based graph construction \cite{liu2010large}. Based on this insight, we employ the global node embeddings to construct the global pseudo graph:
\begin{equation}\label{pseudo_graph}
  \bar{A} = \phi (\bar{H}\bar{H}^T),
\end{equation}
where $\phi(x)=\max(x, 0)$. Since graphs in the real world are generally sparse, we limits the number of neighbors of each node in the constructed pseudo graph no more than $s$. i.e., each row of $\bar{A}$ only reserves the largest $s$ elements, and other elements are set to 0. Besides, the rows of $\bar{A}$ are normalized with $\sum_{j}^{M} \bar{A}_{ij} = 1$. By the way, if the pseudo graph to be constructed is too large, any large-scale graph construction methods can also be considered \cite{liu2010large}.

\subsection{Model Training}
Algorithm \ref{algorithm} is the training process of FedGL. It mainly contains clients and server two parts. Clients are responsible for the training of local models in parallel. They simultaneously exploit the global self-supervision information discovered by the server to improve the quality of local models. Server is responsible for aggregating the local models uploaded by clients to obtain the global model. More importantly, server discovers the global pseudo label and global pseudo graph from the prediction results and node embeddings uploaded by clients. Clients and server alternately iteratively perform, until the global model converges.

\begin{algorithm}[t]
    \caption{The algorithm of FedGL}
    \label{algorithm}
    \begin{algorithmic}[1]
    \REQUIRE Adjacent matrix $\{A_k\}_{k=1}^{K}$, node feature matrix $\{X_k\}_{k=1}^{K}$, label matrix $\{Y_k\}_{i=1}^{K}$, self-supervised learning coefficient $\alpha$, global pseudo graph coefficient $\beta$, confidence threshold $\lambda$, neighbor number $s$.
    \ENSURE Global model $\bar{W}$
    \STATE Server randomly initialize global model parameters $\bar{W}$, and initialize global pseudo label $\bar{Y}$ and global pseudo graph $\bar{A}$ to be zero matrices, distributing them to each client.
    \STATE \textbf{while} not converge \textbf{do}
    \STATE \quad // \textit{Client: local model training using global self-supervision}
    \STATE \quad \textbf{for} $k = 1$ to $K$ \textbf{do in parallel}
    \STATE \quad \quad Use $\bar{A}$ to complement $A_k$ by Eq. \eqref{update_adj}.
    \STATE \quad \quad Obtain the node embedding $H_k$ and prediction result $P_k$ by Eq. \eqref{emb} and Eq. \eqref{pred}.
    \STATE \quad \quad Minimize the GCN loss and SSL loss in Eq. \eqref{L_final} to update the local model parameter $W_k$ by back-propagation.
    \STATE \quad \quad Upload $W_k$, $P_k$, and $H_k$ to the server.
    \STATE \quad \textbf{end}
    \STATE \quad // \textit{Server: global model aggregation and global self-supervision discovery}
    \STATE \quad Update global model $\bar{W}$ by Eq. \eqref{global_model}.
    \STATE \quad Obtain global prediction results by Eq. \eqref{global_pred}.
    \STATE \quad Discover the global pseudo label $\bar{Y}$ by Eq. \eqref{pseudo_label}.
    \STATE \quad Obtain global node embeddings by Eq. \eqref{global_emb}.
    \STATE \quad Construct global pseudo graph $\bar{A}$ by Eq. \eqref{pseudo_graph}.
    \STATE \quad Distribute $\bar{W}$, $\bar{Y}$, and $\bar{A}$ to each client.
    \STATE \textbf{end}
    \end{algorithmic}
\end{algorithm}

\section{Experiments}\label{Exp}
In this section, we conduct extensive experiments on the node classification task to empirically evaluate the effectiveness of the proposed framework FedGL. Besides, experiments are also conducted under different settings of federated learning. Finally, parameter study experiments are also conducted to comprehensively analyze the developed FedGL framework.

In short, we conduct extensive experiments to answer the following questions:
\begin{itemize}
   \item \textbf{Q1}: Whether FedGL can learn a high-quality global graph model, and its performance is close to or even better than the centralized method under the \textit{global goal}?
   \item \textbf{Q2}: Whether the proposed global self-supervision can mitigate the \textit{heterogeneity} and utilize the \textit{complementarity} of graph data between clients, so as to learn more superior node embeddings and achieve better performance than the simple federated method under the \textit{global goal}?
   \item \textbf{Q3}: Whether the learned global model can gain some performance improvements compared to the local method under the \textit{local goal}?
   \item \textbf{Q5}: Whether FedGL consistently performs well under different settings of federated learning?
   \item \textbf{Q5}: How do the parameters of global self-supervision affect the performance of FedGL?
\end{itemize}

\begin{table}[t]
\caption{Statistics of datasets.}
\centering
\begin{tabular}{lcccc}
\toprule
Dataset & \#Nodes & \#Edges & \#Features & \#Classes  \\
\midrule
Cora         & 2708         & 5429         & 1433          & 7            \\
Citeseer     & 3327         & 4732         & 3703          & 6            \\
ACM          & 3025         & 13128        & 1870          & 3            \\
Wiki         & 2405         & 17981        & 4973          & 17           \\
\bottomrule
\end{tabular}
\label{tab_dataset}
\end{table}

\subsection{Datasets}
We conduct experiments on four widely used graph datasets \cite{yang2016revisiting,kipf2016semi,velivckovic2017graph,gao2018large,liu2020towards}, Cora, Citeseer, ACM, and Wiki. The statistics of these datasets are presented in Table \ref{tab_dataset}. The details of each dataset are as follows:
\begin{itemize}
  \item \textbf{Cora}. It is an academic citation network, each node represents a paper, and the edge represents the citation relationship between the papers. The field of the paper is used as the node label. The content of papers is transformed into bag-of-words representations as the initial node features.
  \item \textbf{Citeseer}. It is also an academic citation network. Like Cora, it uses bag-of-words representations as the initial node features.
  \item \textbf{ACM}. It is an academic network. Each node represents a paper, and the edge represents a co-author between the papers. The papers are collected from three fields as node labels. The keywords of papers are transformed into bag-of-words representations as the initial node features.
  \item \textbf{Wiki}. It is a web page link network, derived from the English Wikipedia website. Each node represents a web page containing an explanation of the term, and the edge represents the hyperlink references between web pages. The category of the web page entry is used as the node label. The content of web pages is transformed into bag-of-words representations as the initial node features.
\end{itemize}

In our experiments, the graph data are split into the training set, validation set, and testing set in two different ways to comprehensively evaluate the effectiveness of FedGL.
\begin{itemize}
  \item \textit{Fixed split}. Originating from \cite{yang2016revisiting}, which uses all node features with 20 labels per class as the training set, 500 labels as the validation set for early-stopping, and 1000 labels as the testing set. This fixed split has been widely followed by the GCN related papers \cite{kipf2016semi,velivckovic2017graph,gao2018large,liu2020towards}, since the split data is publicly available \footnote{\url{https://github.com/tkipf/gcn}} and facilitates performance comparison between papers. If there are no special instructions in subsequent experiments, this split method will be adopted by default.
  \item \textit{Random split}. It has more severely limited labels and greater randomness. For Cora, Citeseer, and Wiki, we randomly choose 5, 10, 15 labels per class as the training set, 500 labels for validation, and 1000 labels for testing. Since ACM only has 3 classes, we randomly choose 15, 25, 35 labels per class as the training set to ensure that the training labels are not too few so that the model can be learned normally.
\end{itemize}

\subsection{Comparison Methods}
Note that there are rare few studies focusing on graph data learning in federated scenarios. Several unpublished works that can be found are also under different scenario assumptions and cannot be directly compared. To demonstrate the rationality and effectiveness of FedGL, we compare with the following methods:
\begin{itemize}
  \item \textbf{Centralized method (Centralized)}. For \textit{global goal} comparison, the graph data (including training set, validation set, and testing set) of each client are collected and merged. The merged graph data are fed into the same GCN model for training. Finally, the trained model is evaluated on the global testing set. Note that this method is an ideal method that is not feasible in real scenarios, because it is often unrealistic to gather data together due to privacy security and industry competition.
  \item \textbf{Local method (Local)}. For \textit{local goal} comparison, each client trains the same GCN model by feeding its graph data independently. Finally, the trained model is evaluated on the local testing set.
  \item \textbf{Simple federated method (Federated)}. For \textit{global goal} and \textit{local goal} comparison, this method uses the weighted average method to aggregate the local models to obtain the global model, which can be regarded as FedGL without global self-supervision. It is used to intuitively verify the effectiveness of the proposed global self-supervision module.
  \item \textbf{FedGL w/o GPG}. For \textit{global goal} and \textit{local goal} comparison, this method is an ablation version of FedGL by removing the global pseudo graph (GPG), which is used to verify the effectiveness of global pseudo graph.
  \item \textbf{FedGL w/o GPL}. For \textit{global goal} and \textit{local goal} comparison, this method is an ablation version of FedGL by removing the global pseudo label (GPL), which is used to verify the effectiveness of global pseudo label.
\end{itemize}

\subsection{Experimental Settings}\label{exp_set}
\subsubsection{Data Settings for Federated Learning} In order to simulate the graph data distribution in the real world, the graph data of each client comes from the random sampling results of the experimental datasets under different proportions to ensure the number of nodes, graph structure, and label distribution between clients to be diverse. That is, the graph data between clients are Non-IID. Meanwhile, there are some overlapping nodes between clients due to random sampling. In Section \ref{problem_def}, we defined two goals in federated scenario, namely \textit{global goal} and \textit{local goal}, which are evaluated based on global testing set and local testing set. The following is the specific implementation:
\begin{itemize}
   \item \textit{Global goal with global testing set}. The graph structure, feature matrix, and testing labels on each client are merged as the global testing set. This is the most intuitive implementation to evaluate the global model. In practical scenarios, the global testing set can be specially customized. The final global model is evaluated on the global testing set.
   \item \textit{Local goal with local testing set}. The local testing set is exactly the testing set of each client. The final global model is distributed to each client and evaluated on each local testing set.
\end{itemize}

\subsubsection{Parameter Settings} FedGL consists of three modules, i.e., federated learning, GCN model, and global self-supervision. The parameters of the GCN model directly follow the settings of its original paper \cite{kipf2016semi}, i.e., two convolutional layers, 16 hidden units, 0.5 dropout rate, 0.01 learning rate, and $5\times 10^{-4}$ $L_2$ regularization. For federated learning, we use 6 clients with sampling proportion of each client [30\%,40\%,50\%,50\%,60\%,70\%], client participation ratio per round 100\%, local training epochs of client per round 10, maximum iteration round 300, and early-stopping round 30. For global self-supervision, the confidence threshold $\lambda$ is set to 0.1 for Wiki, 0.5 for other datasets. The self-supervised learning coefficient $\alpha$ is set to 0.1 for Wiki, 0.2 for other datasets. The global pseudo graph coefficient $\beta$ is set to 1. The neighbor number $s$ is set to 100. All the experiments are repeated 5 times and the average results are reported.

\begin{table*}[t]
    \caption{Node classification accuracy under fixed split.}
    \centering
    \begin{tabular}{l|c|c|ccc}
    \toprule
    Dataset      & Centralized & Federated         & FedGL w/o GPG     & FedGL w/o GPL & FedGL  \\
    \midrule
    Cora     & 0.811     & 0.810   & 0.828          & 0.812     & \textbf{0.830}  \\
    Citeseer & 0.705     & 0.676   & 0.732          & 0.676     & \textbf{0.734} \\
    ACM      & 0.848     & 0.855   & \textbf{0.892} & 0.858     & 0.891          \\
    Wiki     & 0.619     & 0.678   & 0.689          & 0.673     & \textbf{0.691} \\
    \bottomrule
    \end{tabular}
    \label{tab_global_fixed_acc}
\end{table*}

\subsection{Q1Q2: Experimental Results under Global Goal}
\subsubsection{Fixed Split} Table \ref{tab_global_fixed_acc} shows the node classification accuracy under fixed split. The best results are highlighted in bold fonts. As can be seen, FedGL remarkably outperforms Centralized and Federated on all datasets. Further analysis, we have the following observations.
\begin{itemize}
  \item Compared with ideal method Centralized, FedGL gains about 2\%-7\% absolute performance improvement under various datasets, which indicates that FedGL is not only not affected by the inability to collect data, but also fully integrates the data of each client for training, and learns a high-quality global model. There are two reasons why FedGL outperforms Centralized. (1) The proposed global self-supervision module improves each local model from the training labels and graph structure respectively, leading to a high-quality global model. (2) Due to the particularity of graph data, there are overlapping nodes between clients, so the graph data of each client can be regarded as a sampling from the large graph data. Each client uses the sampled graph data to train a local model, which is equivalent to the process of Bagging ensemble learning, or understood as the process of data augmentation, so it is better than Centralized using merged single graph data.
  \item Compared with Federated, FedGL performs better under all datasets. Especially on Citeseer, the absolute performance improvement is up to 5.8\%. Note that the only difference between Federated and FedGL is that FedGL discovers and exploits the global self-supervision information to tackle the \textit{heterogeneity} and \textit{complementarity} of graph data between clients. This result directly verifies the effectiveness of global self-supervision.
  \item FedGL w/o GPG achieves better performance than FedGL w/o GPL and is closer to or even slightly surpassing FedGL, which indicates that the global pseudo label is more helpful to learn a high-quality global model than the global pseudo graph. Meanwhile, the global pseudo label and the global pseudo graph are essentially complementary to each other, so using both simultaneously performs best in most cases.
\end{itemize}

\begin{table*}[t]
    \caption{Node classification accuracy under random split.}
    \centering
    \begin{tabular}{l|c|c|c|ccc}
    \toprule
    Dataset                                             & Label ratio & Centralized & Federated       & FedGL w/o GPG   & FedGL w/o GPL   & FedGL  \\
    \midrule
    \multirow{3}{*}{Cora}     & 5      & 0.549          & 0.587          & 0.629          & 0.580          & \textbf{0.640}  \\ 
                              & 10     & 0.698          & 0.689          & 0.733          & 0.692          & \textbf{0.737} \\ 
                              & 15     & 0.740          & 0.738          & 0.799          & 0.739          & \textbf{0.806} \\ \hline
    \multirow{3}{*}{Citeseer} & 5      & 0.555          & 0.577          & \textbf{0.610} & 0.579          & 0.605          \\ 
                              & 10     & 0.636          & 0.644          & 0.635          & \textbf{0.646} & 0.620           \\ 
                              & 15     & 0.648          & 0.662          & 0.708          & 0.662          & \textbf{0.710}  \\ \hline
    \multirow{3}{*}{ACM}      & 15     & 0.723          & 0.754          & 0.811          & 0.760          & \textbf{0.852} \\ 
                              & 25     & 0.845          & 0.855          & 0.892          & 0.855          & \textbf{0.892} \\ 
                              & 35     & 0.902          & 0.902          & \textbf{0.908} & 0.901          & 0.903          \\ \hline
    \multirow{3}{*}{Wiki}     & 5      & 0.345          & 0.464          & \textbf{0.484} & 0.463          & 0.482          \\ 
                              & 10     & 0.435          & 0.516          & \textbf{0.551} & 0.510          & 0.550          \\ 
                              & 15     & 0.524          & 0.646          & 0.624          & 0.638          & \textbf{0.651} \\
    \bottomrule
    \end{tabular}
    \label{tab_global_random_acc}
\end{table*}

\subsubsection{Random Split} Considering that the model may exist data preferences for specific data split, we introduce greater randomness and simulate the strictly limited label scenario to repeat the above node classification experiment. The experimental results are reported in Table \ref{tab_global_random_acc}. FedGL still dramatically outperforms Centralized and Federated under most datasets and label ratios, which further verifies the effectiveness of the proposed framework. Further analysis, we have the following observations.
\begin{itemize}
  \item Compared with fixed split (20 labels per class), FedGL shows more conspicuous superiority under random split especially when there are rare few training labels. When there are only 5 (ACM is 15) labels per class, FedGL obtains more than 10\% absolute performance improvement compared to Centralized on Cora, ACM, and Wiki . Such a characteristic is especially suitable for practical applications since it is common to observe graphs with a small number of labeled nodes.
  \item Under various label ratios of random split, FedGL consistently outperforms Federated, gaining more than 5\% absolute performance improvement in most cases, which fully verifies the stability and robustness of the proposed global self-supervision module.
\end{itemize}

\begin{figure*}[t]
    \centering
    \subfloat[Centralized]{
        \includegraphics[width=.19\textwidth]{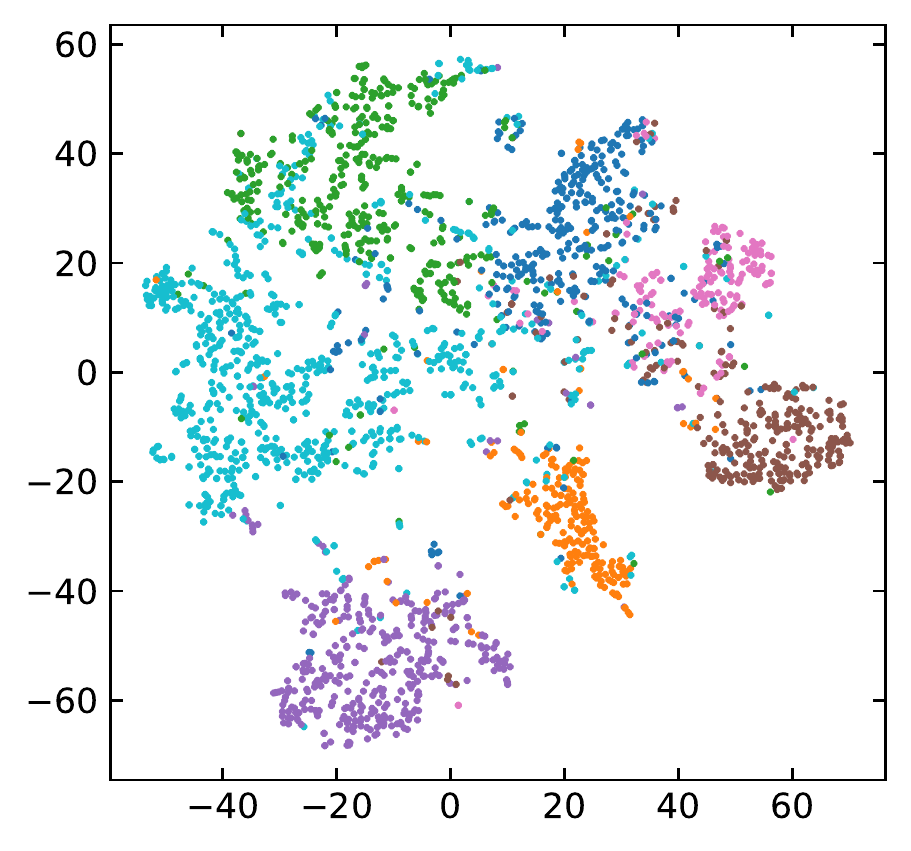}
    }
    \subfloat[Federated]{
        \includegraphics[width=.19\textwidth]{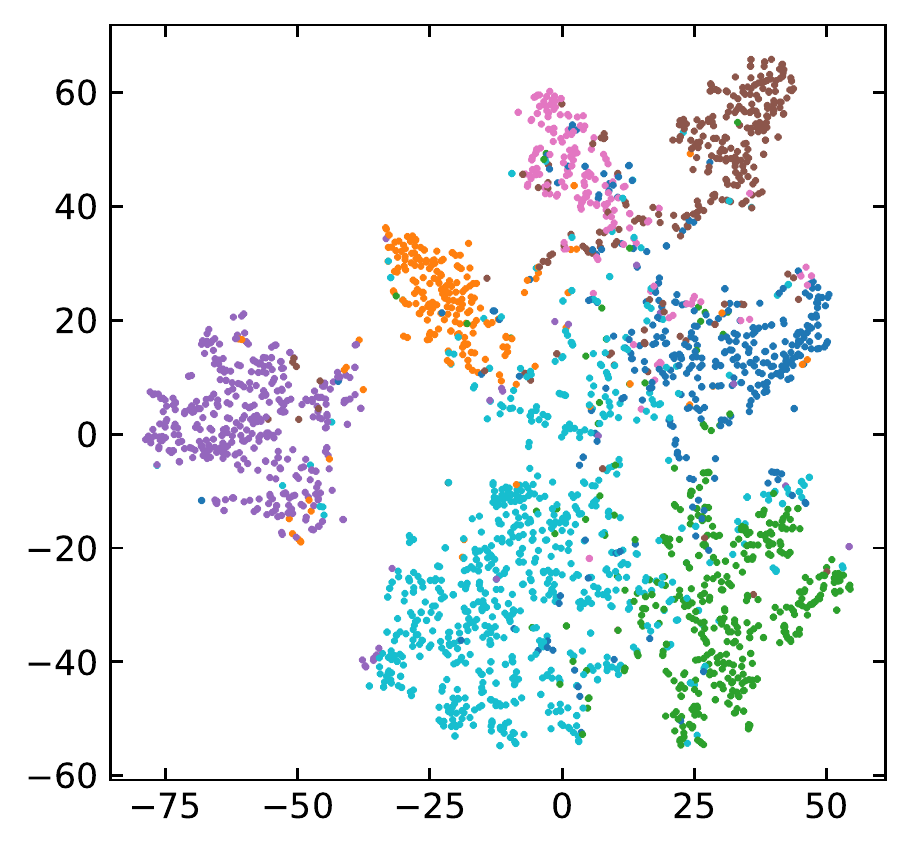}
    }
    \subfloat[FedGL w/o GPG]{
        \includegraphics[width=.19\textwidth]{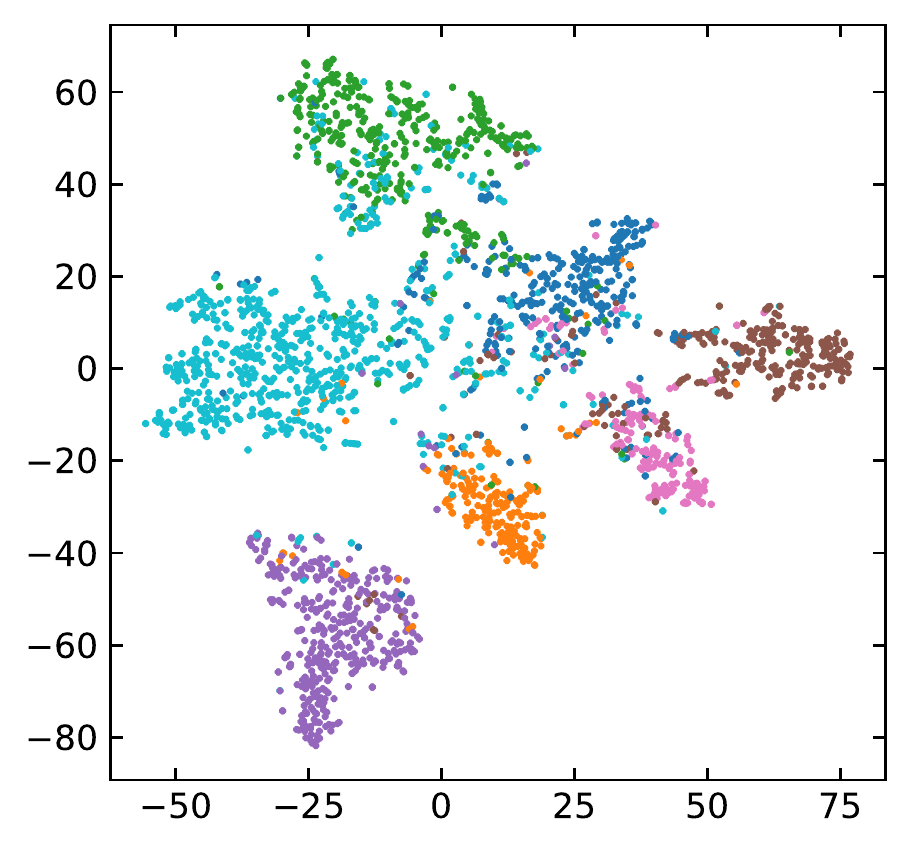}
    }
    \subfloat[FedGL w/o GPL]{
        \includegraphics[width=.19\textwidth]{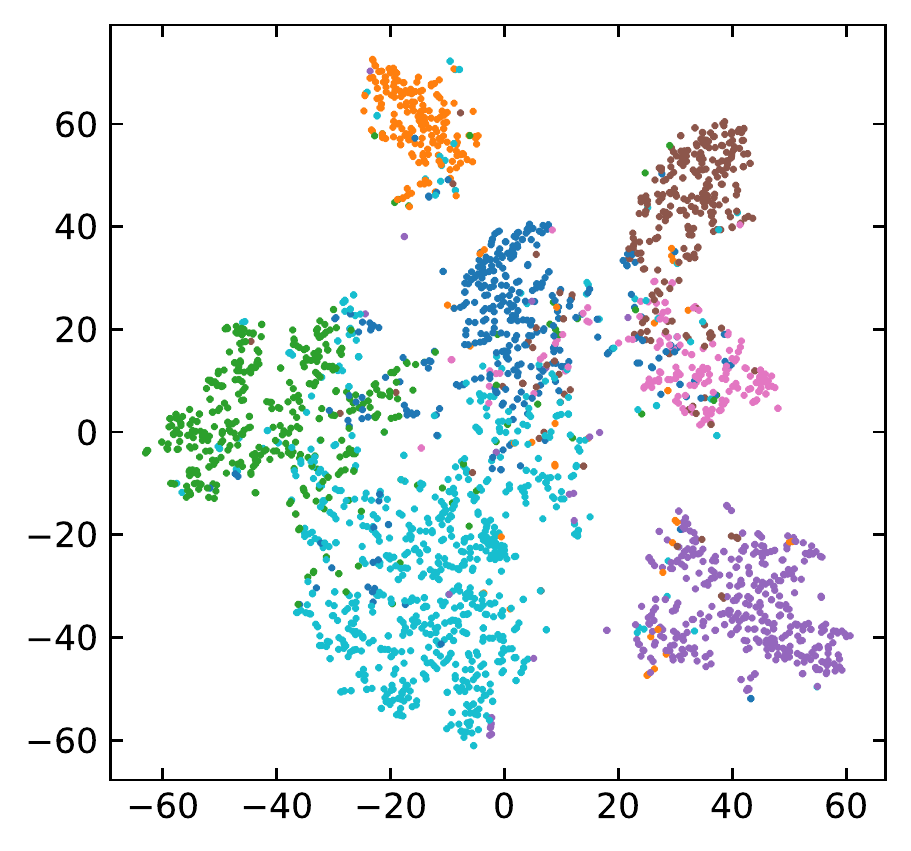}
    }
    \subfloat[FedGL]{
        \includegraphics[width=.19\textwidth]{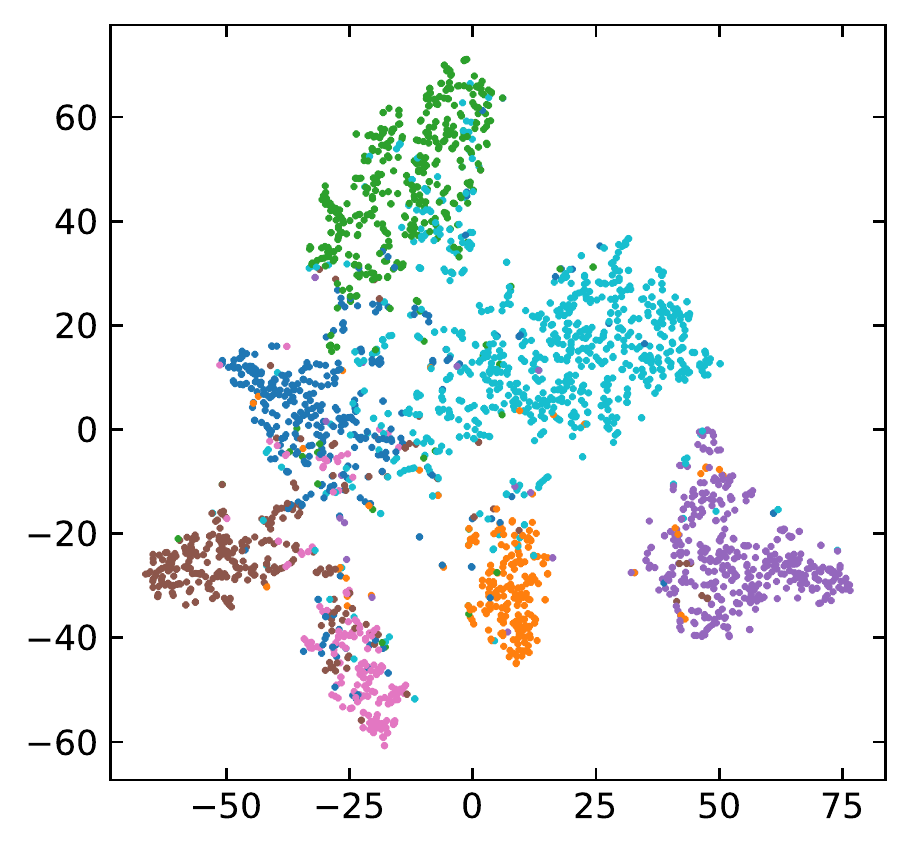}
    }
    \caption{Visualization of node embedding learned by different methods on Cora.}
    \label{fig:plot_cora}
\end{figure*}

\begin{figure*}[t]
    \centering
    \subfloat[Centralized]{
        \includegraphics[width=.19\textwidth]{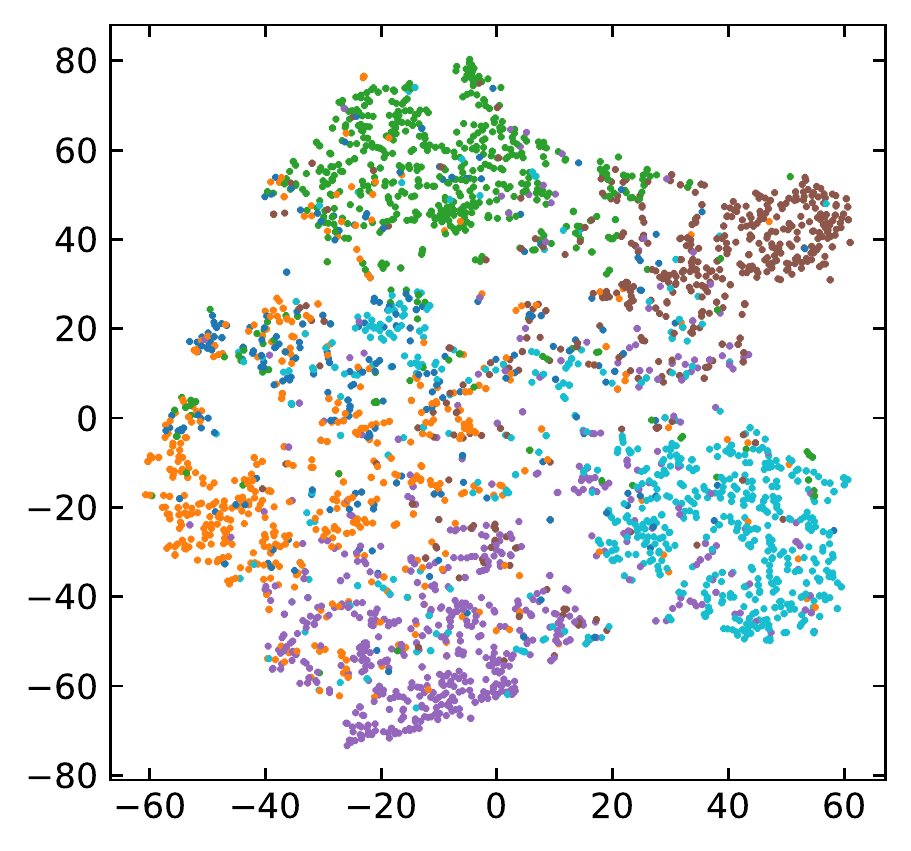}
    }
    \subfloat[Federated]{
        \includegraphics[width=.19\textwidth]{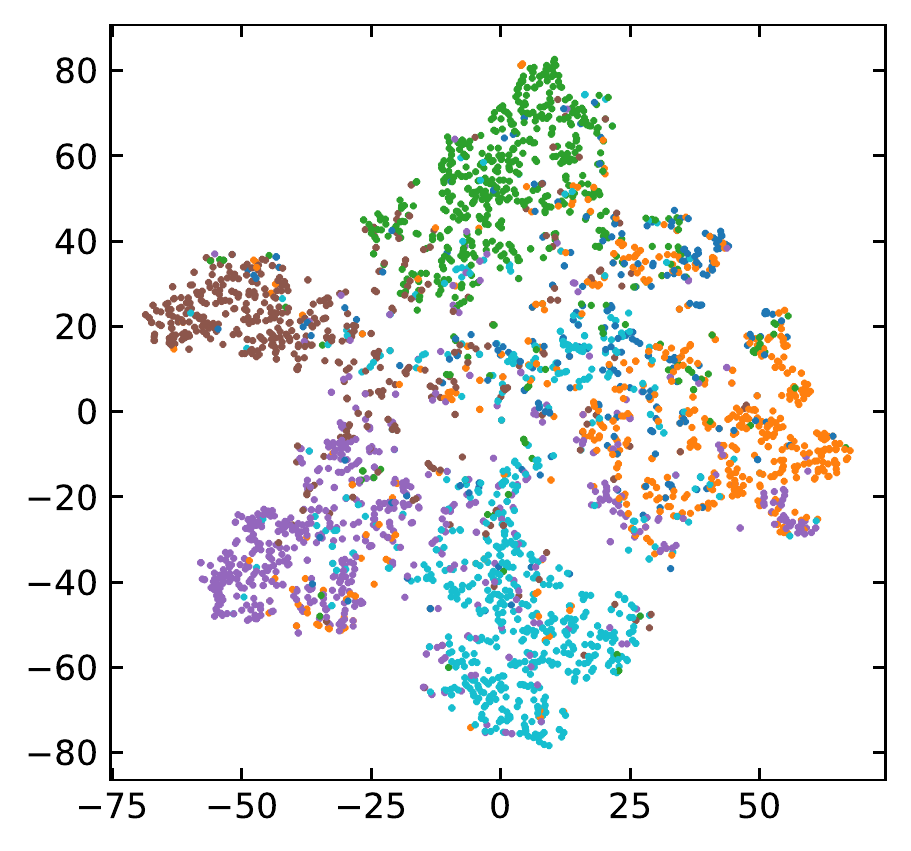}
    }
    \subfloat[FedGL w/o GPG]{
        \includegraphics[width=.19\textwidth]{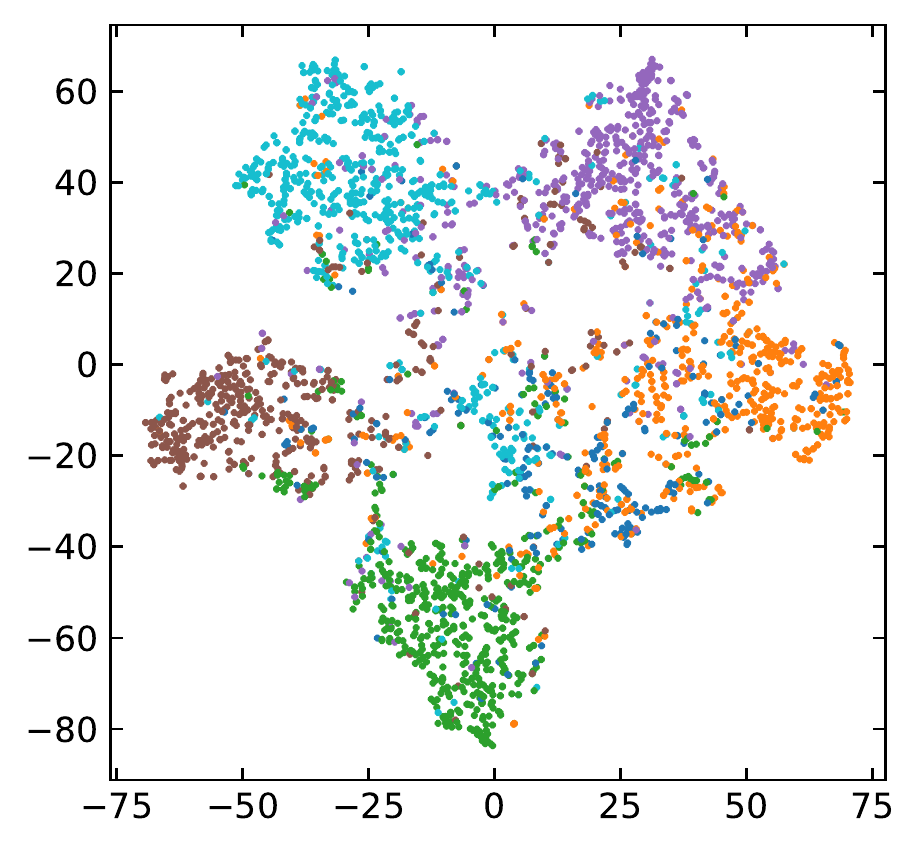}
    }
    \subfloat[FedGL w/o GPL]{
        \includegraphics[width=.19\textwidth]{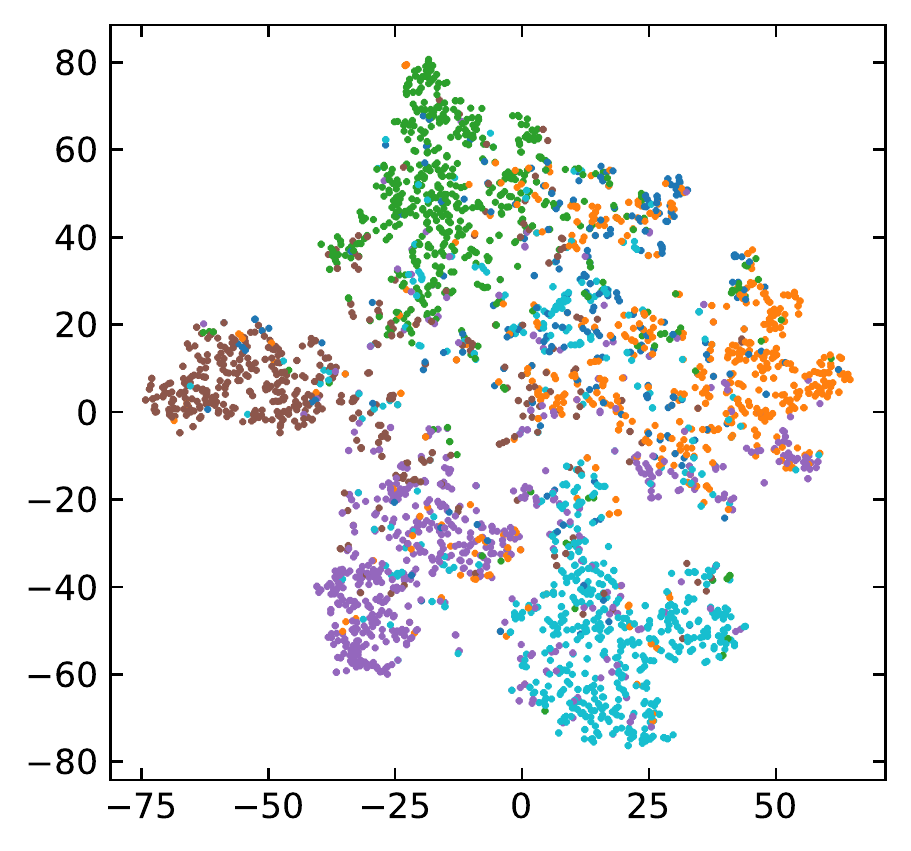}
    }
    \subfloat[FedGL]{
        \includegraphics[width=.19\textwidth]{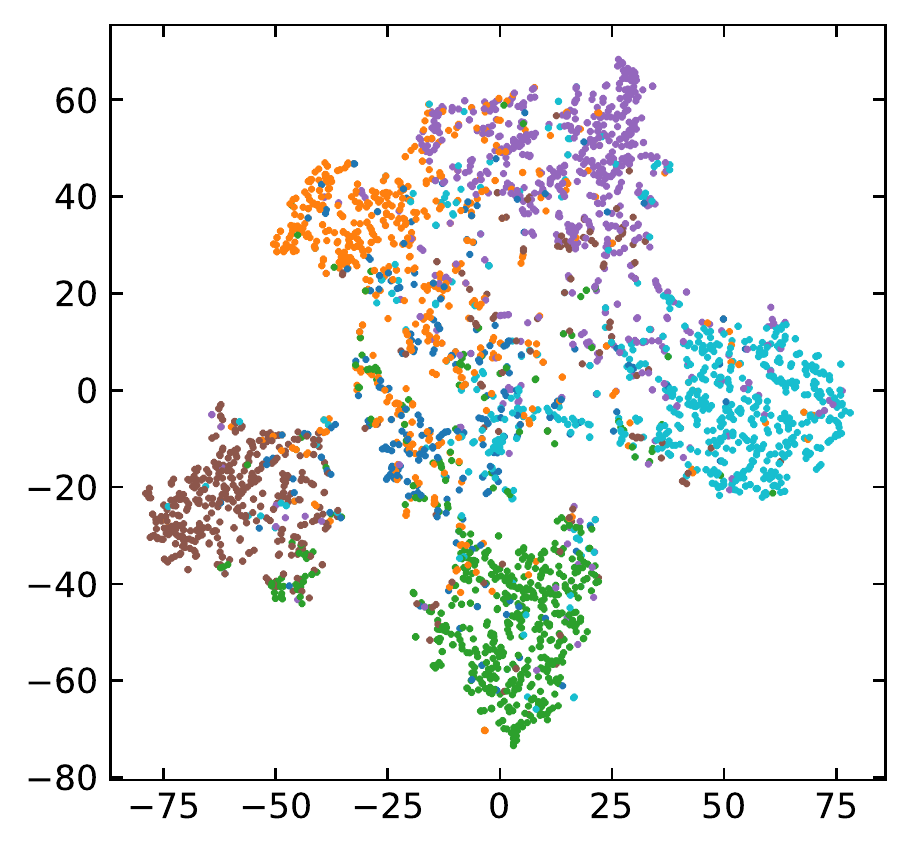}
    }
    \caption{Visualization of node embedding learned by different methods on Citeseer.}
    \label{fig:plot_citeseer}
\end{figure*}

\begin{figure*}[t]
    \centering
    \subfloat[Centralized]{
        \includegraphics[width=.19\textwidth]{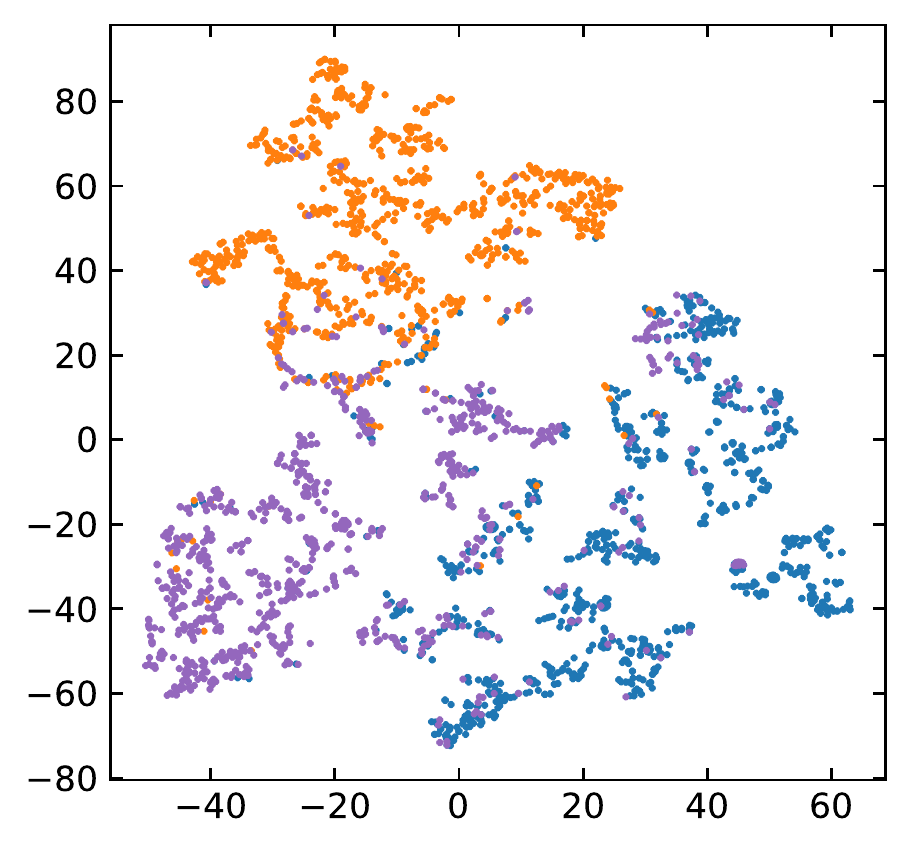}
    }
    \subfloat[Federated]{
        \includegraphics[width=.19\textwidth]{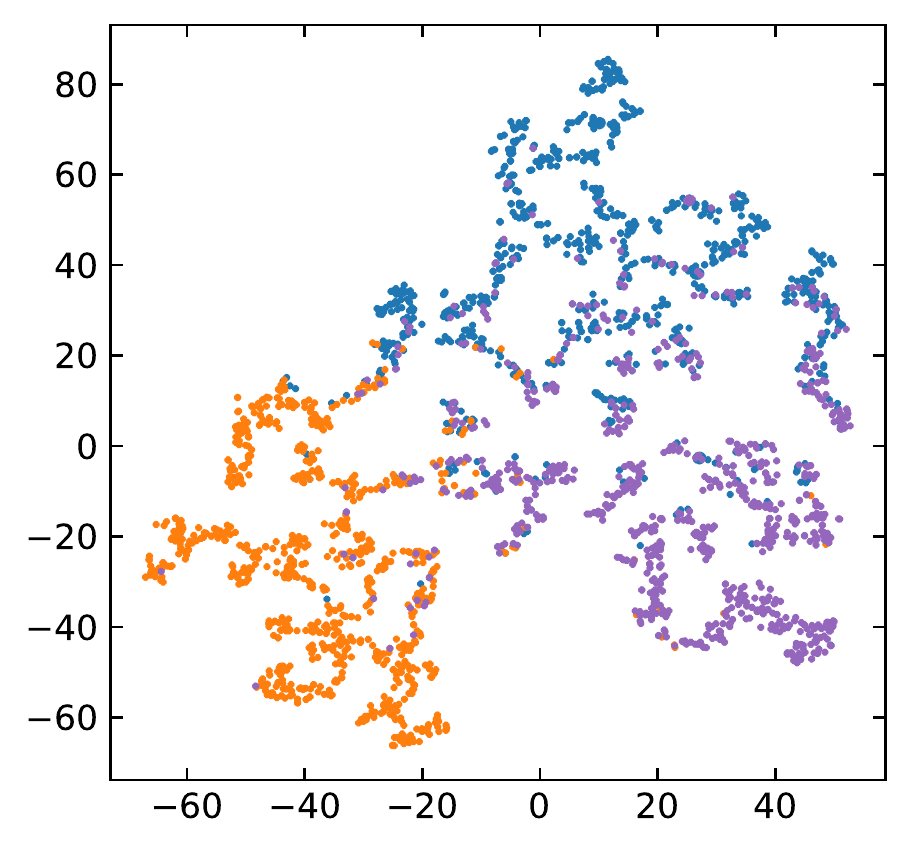}
    }
    \subfloat[FedGL w/o GPG]{
        \includegraphics[width=.19\textwidth]{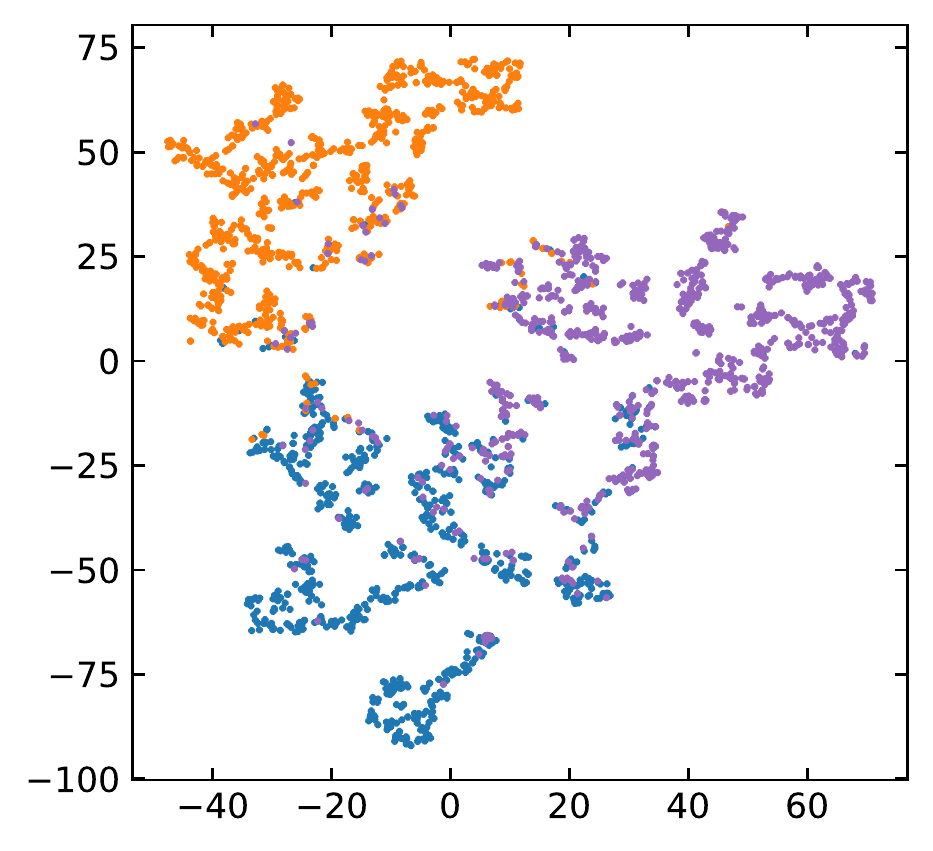}
    }
    \subfloat[FedGL w/o GPL]{
        \includegraphics[width=.19\textwidth]{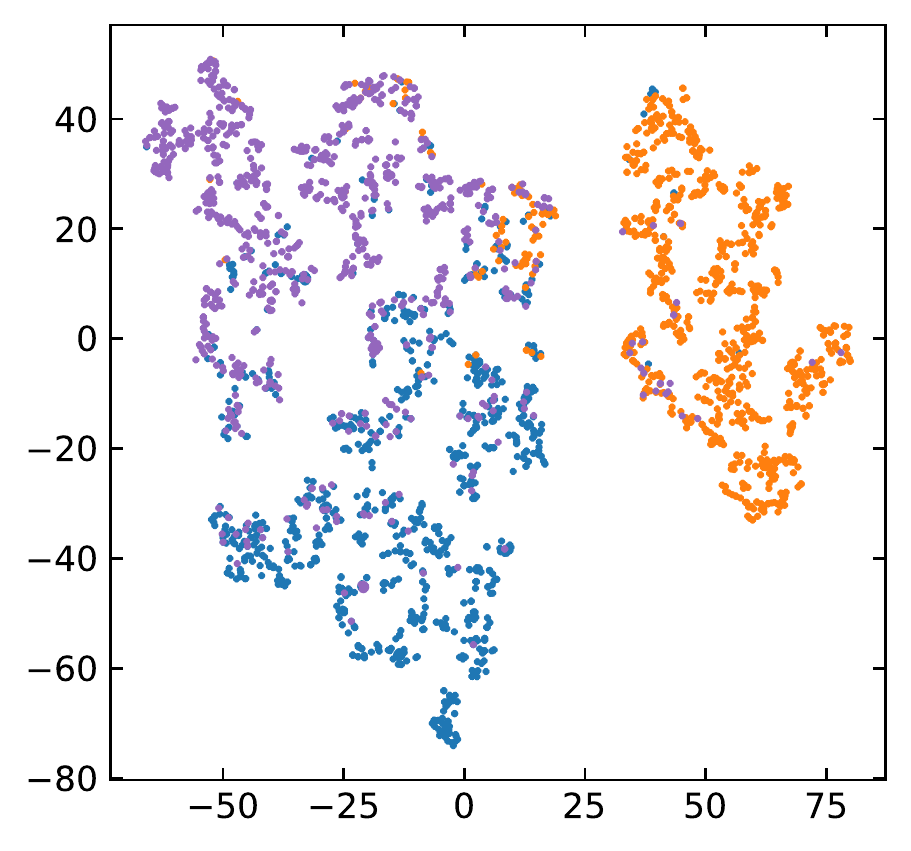}
    }
    \subfloat[FedGL]{
        \includegraphics[width=.19\textwidth]{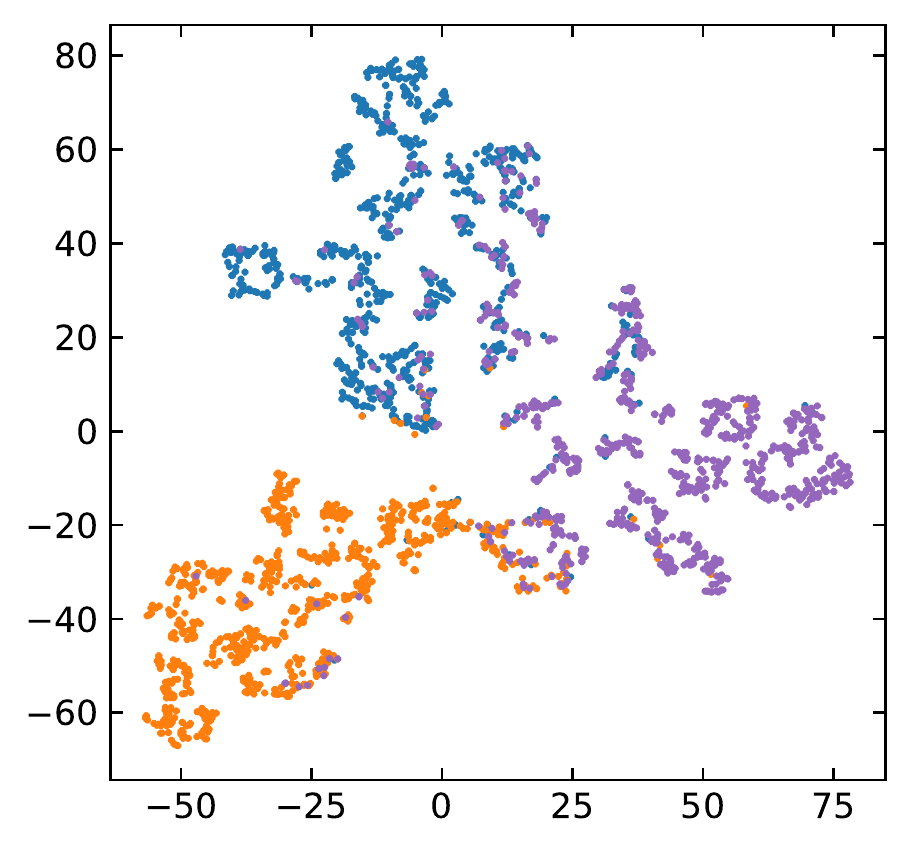}
    }
    \caption{Visualization of node embedding learned by different methods on ACM.}
    \label{fig:plot_acm}
\end{figure*}

\subsubsection{Visualization of Node Embedding} Except for the performance comparison, we intuitively compare the quality of node embeddings by visualization. Concretely, we firstly feed the global testing set into the model learned by Centralized, Federated, FedGL w/o GPG, FedGL w/o GPL, and FedGL to obtain the node embeddings. Then, we map the embeddings into a 2-dimensional space with t-SNE algorithm \cite{van2008visualizing} and draw a scatter plot. Fig. \ref{fig:plot_cora}-\ref{fig:plot_acm} is the visualization results of Cora, Citeseer, and ACM. Since Wiki has 17 classes, it is not easy to display them in color, so Wiki is not reported. In the figures, each scattered point represents a node, and the node with the same color belongs to the same category. As can be seen, the nodes in Centralized are scattered, and there are many overlapping nodes between classes. Federated has fewer overlapping nodes between classes, but the nodes are still scattered and the boundaries between classes are not clear. The nodes in FedGL are quite compact, and the boundaries between classes are clear, which verifies FedGL is capable of learning more discriminative node embeddings, thus performing better than Centralized and Federated in the downstream tasks.

\begin{figure*}[t]
    \centering
    \subfloat[Cora]{
        \includegraphics[width=.24\textwidth]{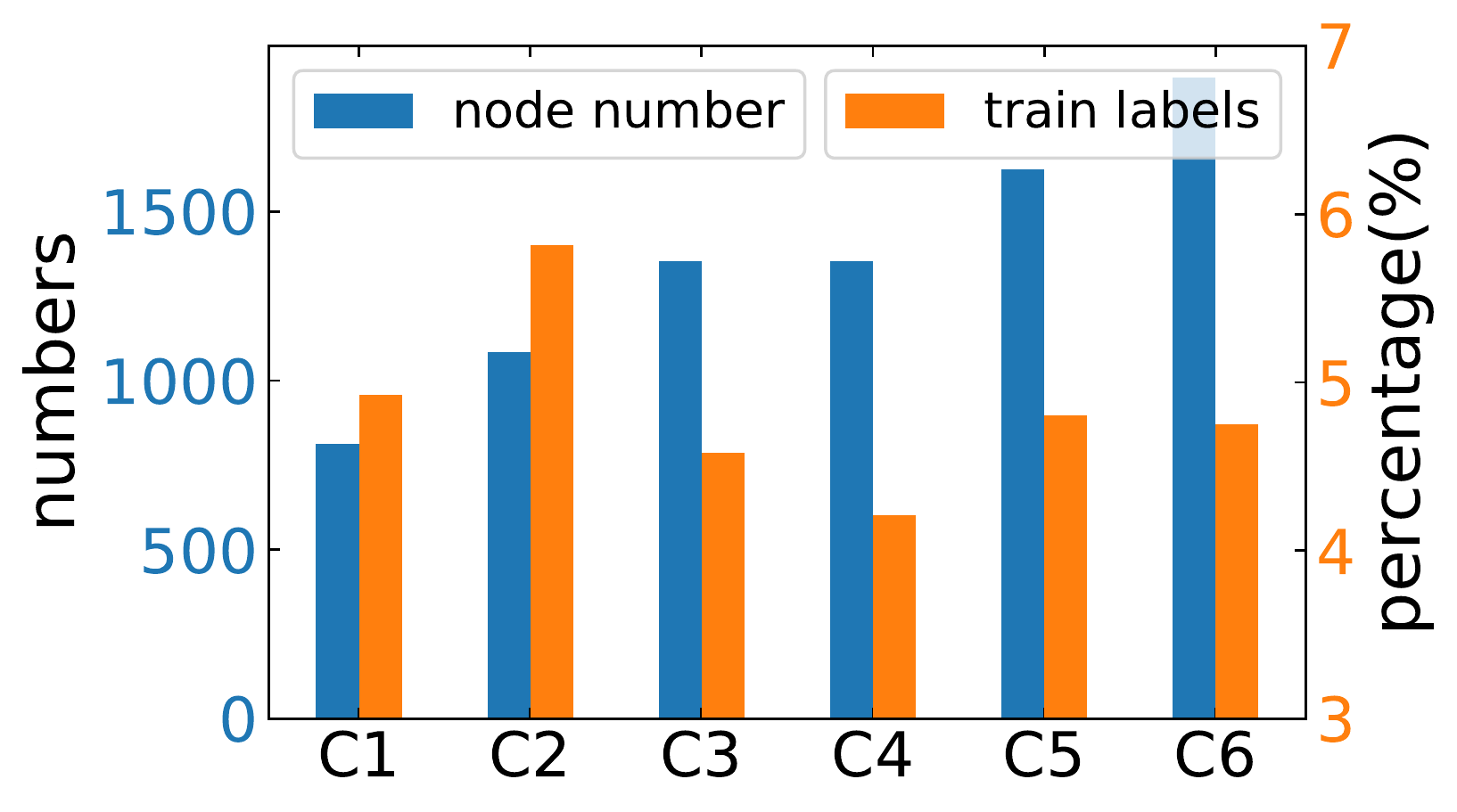}
    }
    \subfloat[Citeseer]{
        \includegraphics[width=.24\textwidth]{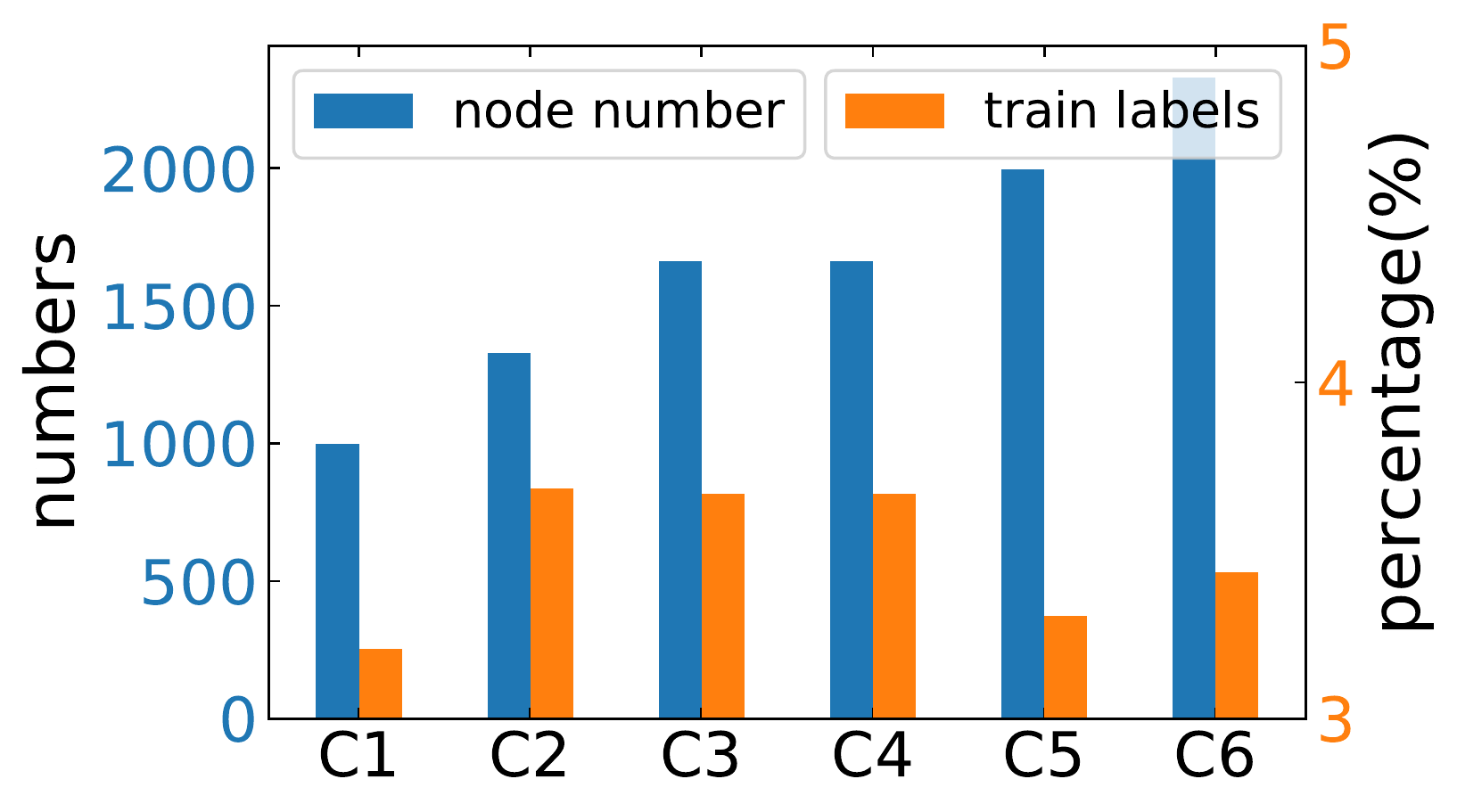}
    }
    \subfloat[ACM]{
        \includegraphics[width=.24\textwidth]{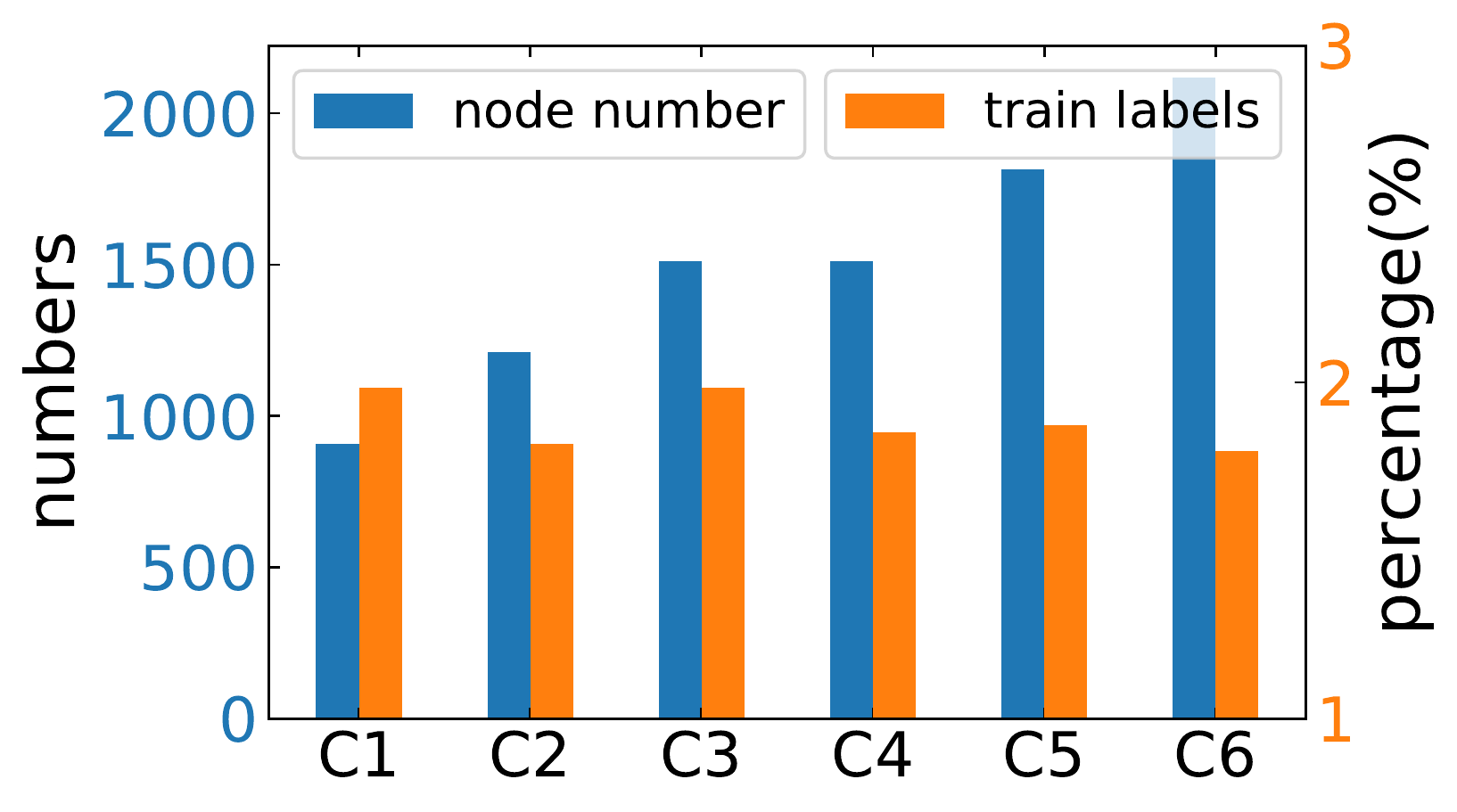}
    }
    \subfloat[Wiki]{
        \includegraphics[width=.24\textwidth]{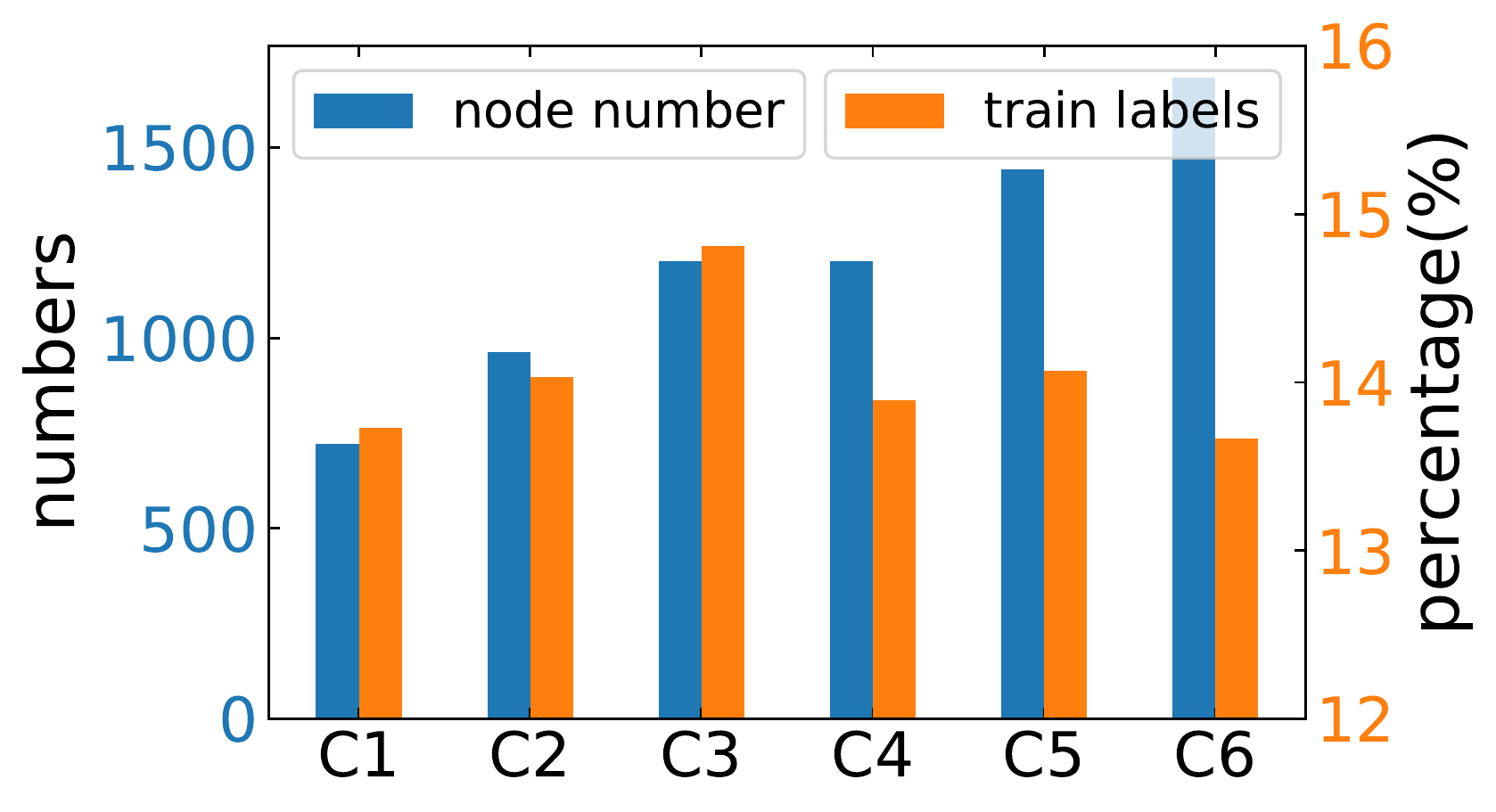}
    }
    \caption{The number of nodes and the proportion of training labels of each client (C1-C6) on different datasets.}
    \label{fig:client_data}
\end{figure*}

\begin{figure*}[t]
    \centering
    \subfloat[Cora]{
        \includegraphics[width=.24\textwidth]{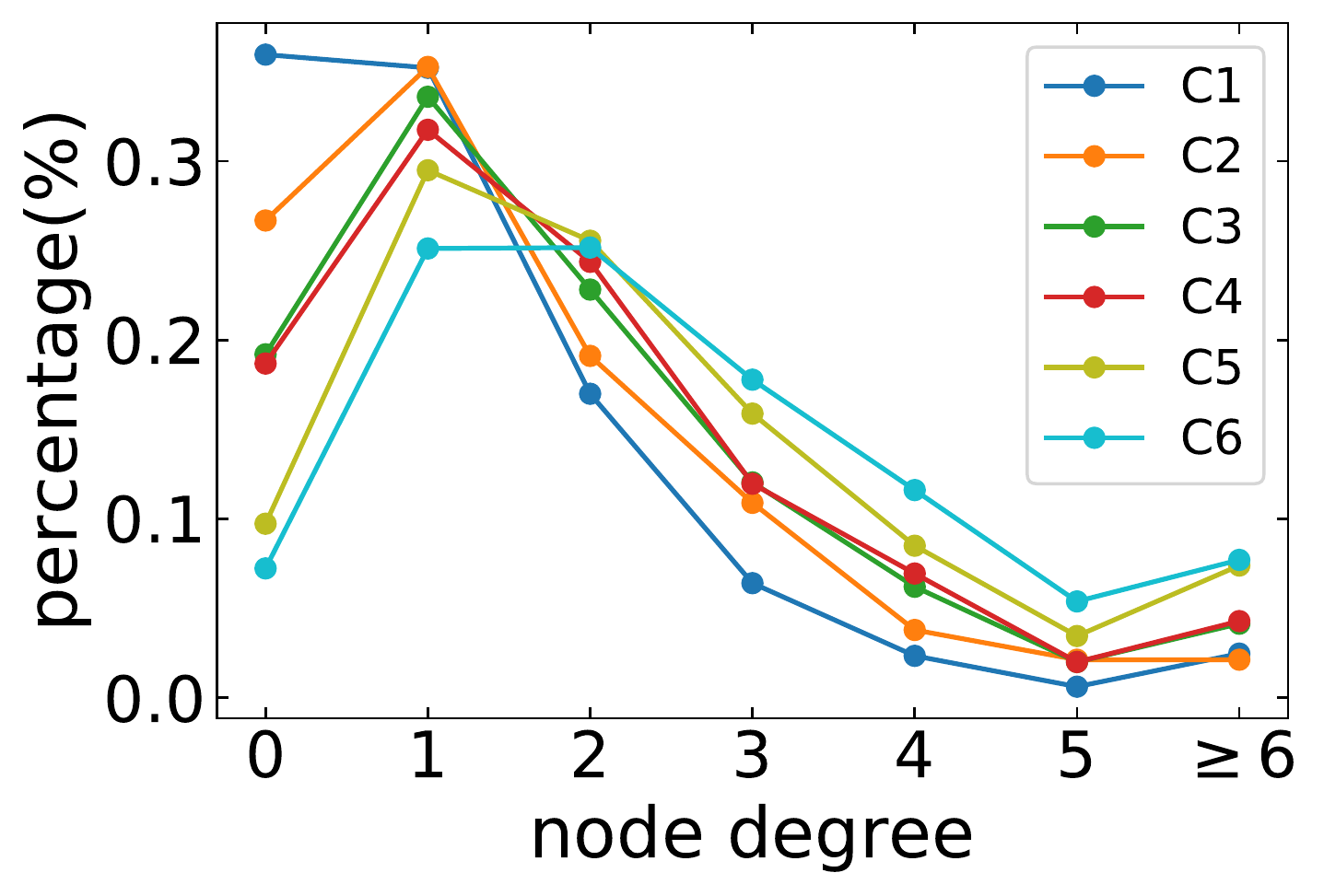}
    }
    \subfloat[Citeseer]{
        \includegraphics[width=.24\textwidth]{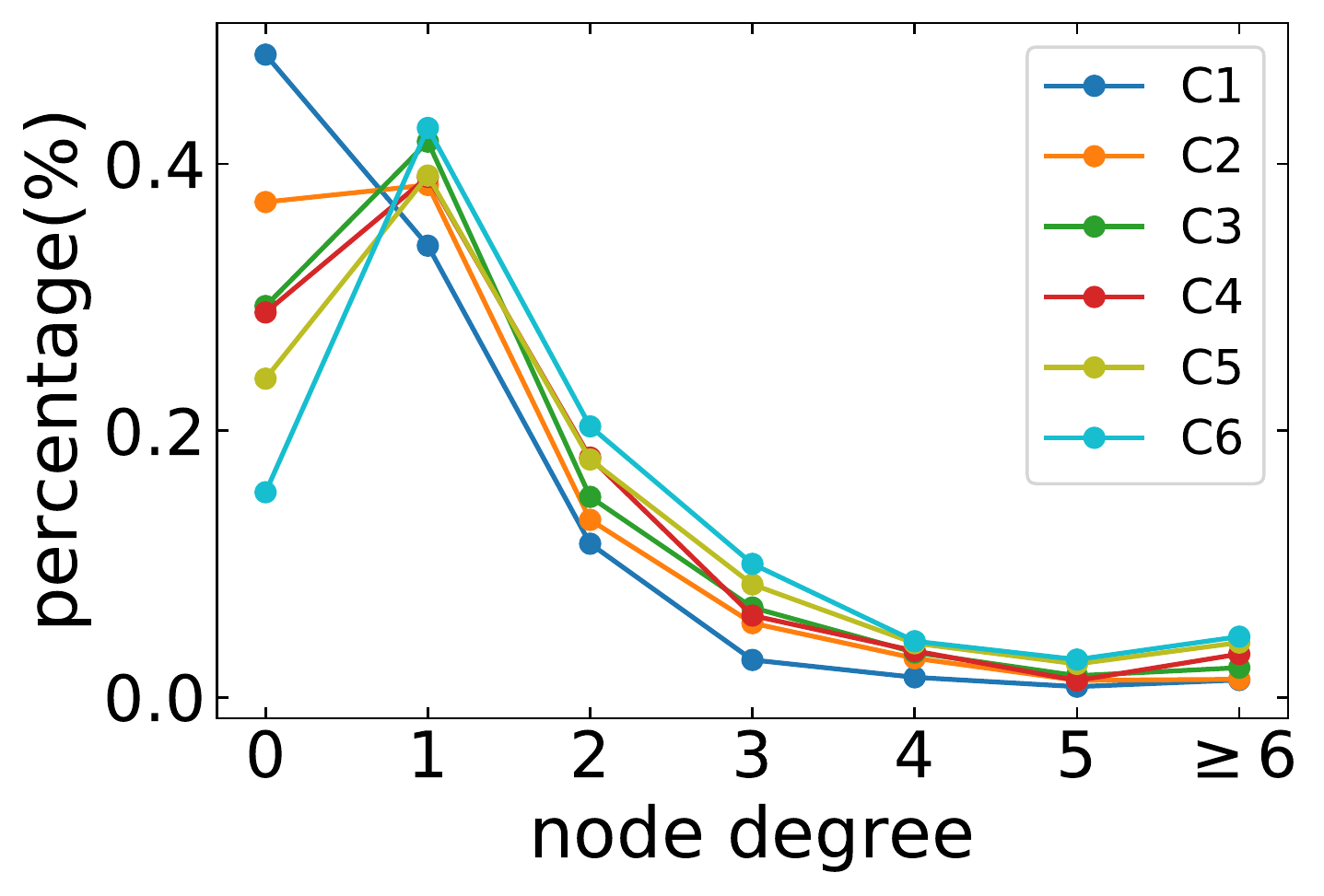}
    }
    \subfloat[ACM]{
        \includegraphics[width=.24\textwidth]{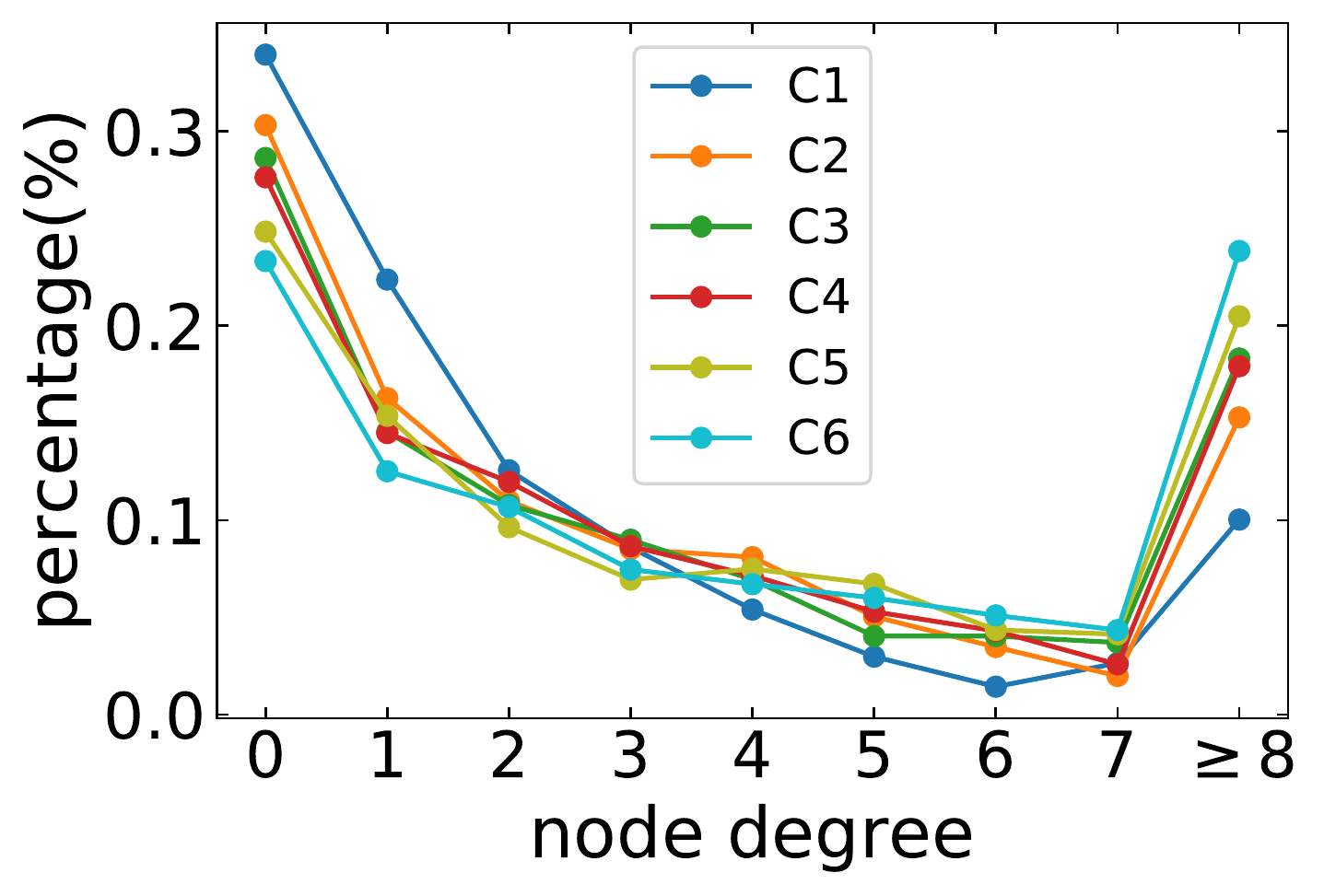}
    }
    \subfloat[Wiki]{
        \includegraphics[width=.24\textwidth]{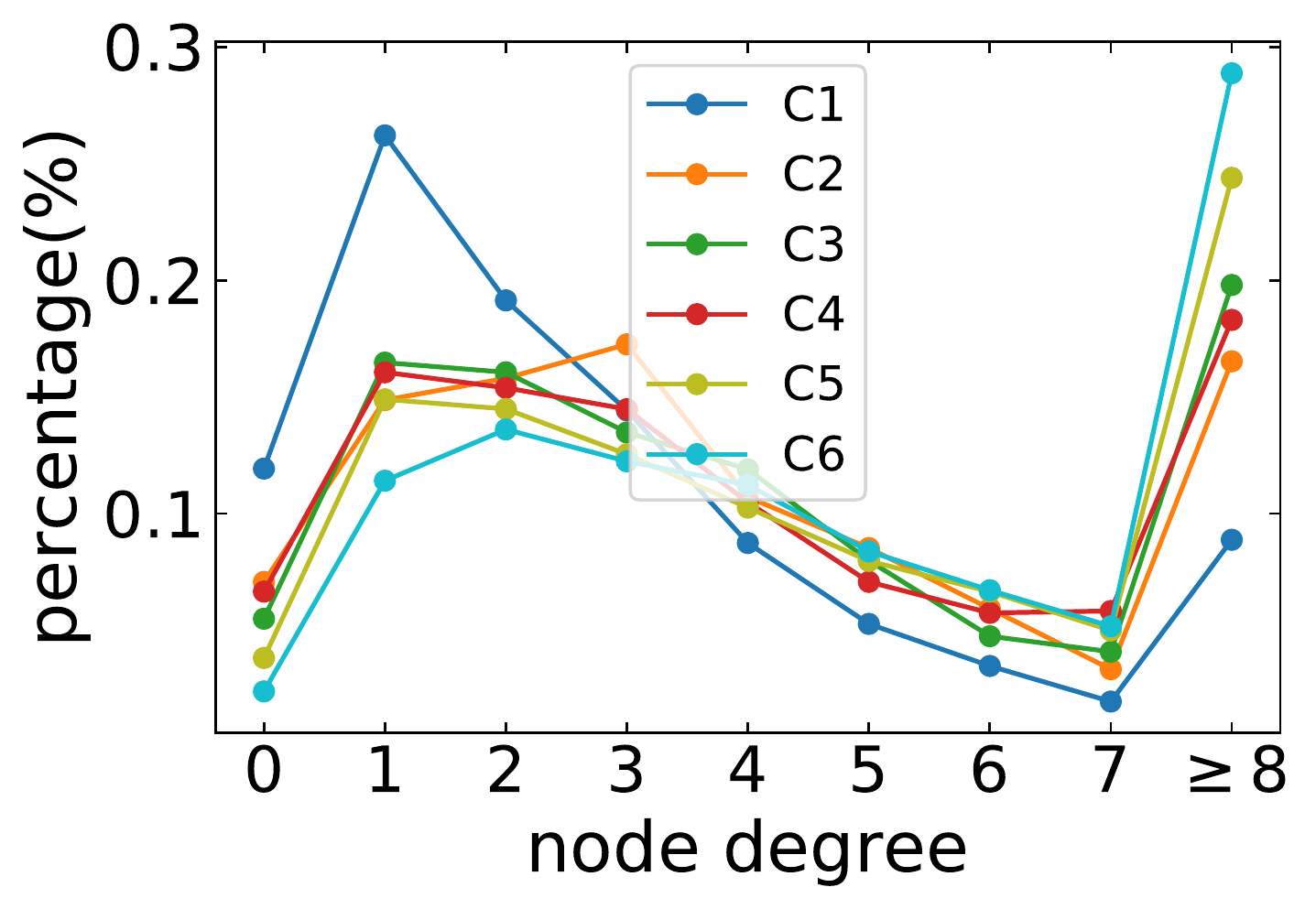}
    }
    \caption{The node degree distribution of each client (C1-C6) on different datasets.}
    \label{fig:client_degree}
\end{figure*}

\begin{figure*}[t]
    \centering
    \subfloat[Cora]{
        \includegraphics[width=.24\textwidth]{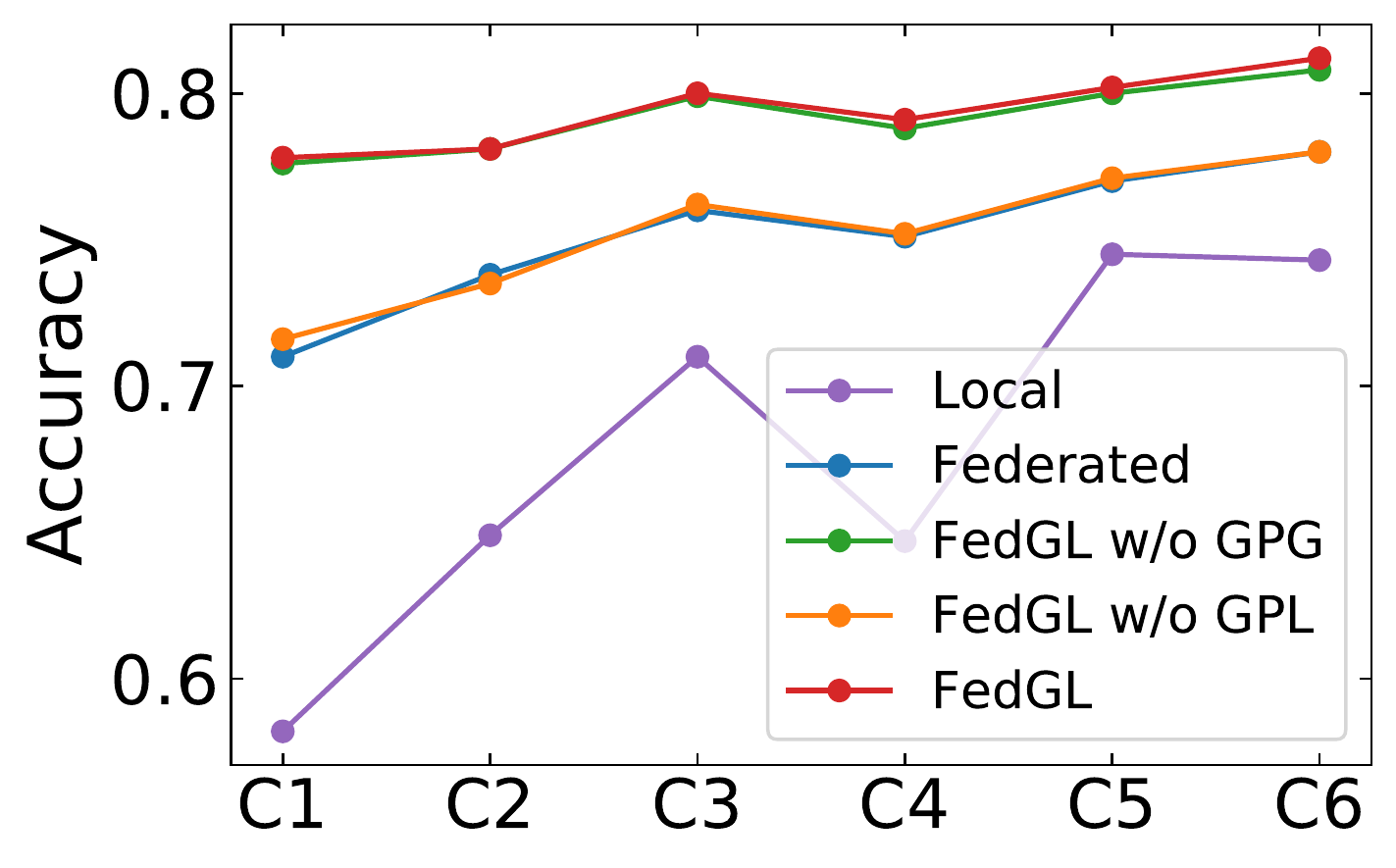}
    }
    \subfloat[Citeseer]{
        \includegraphics[width=.24\textwidth]{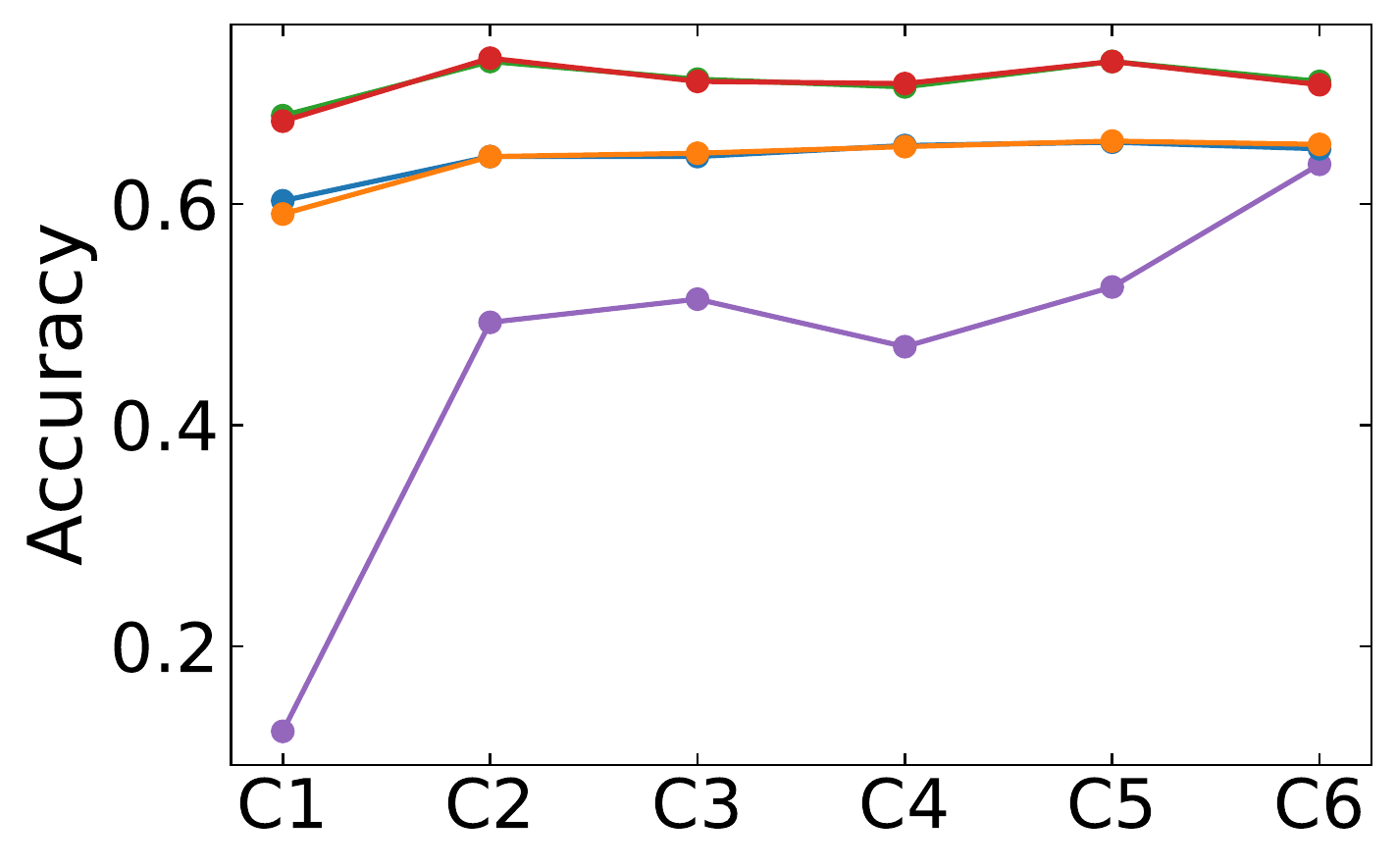}
    }
    \subfloat[ACM]{
        \includegraphics[width=.24\textwidth]{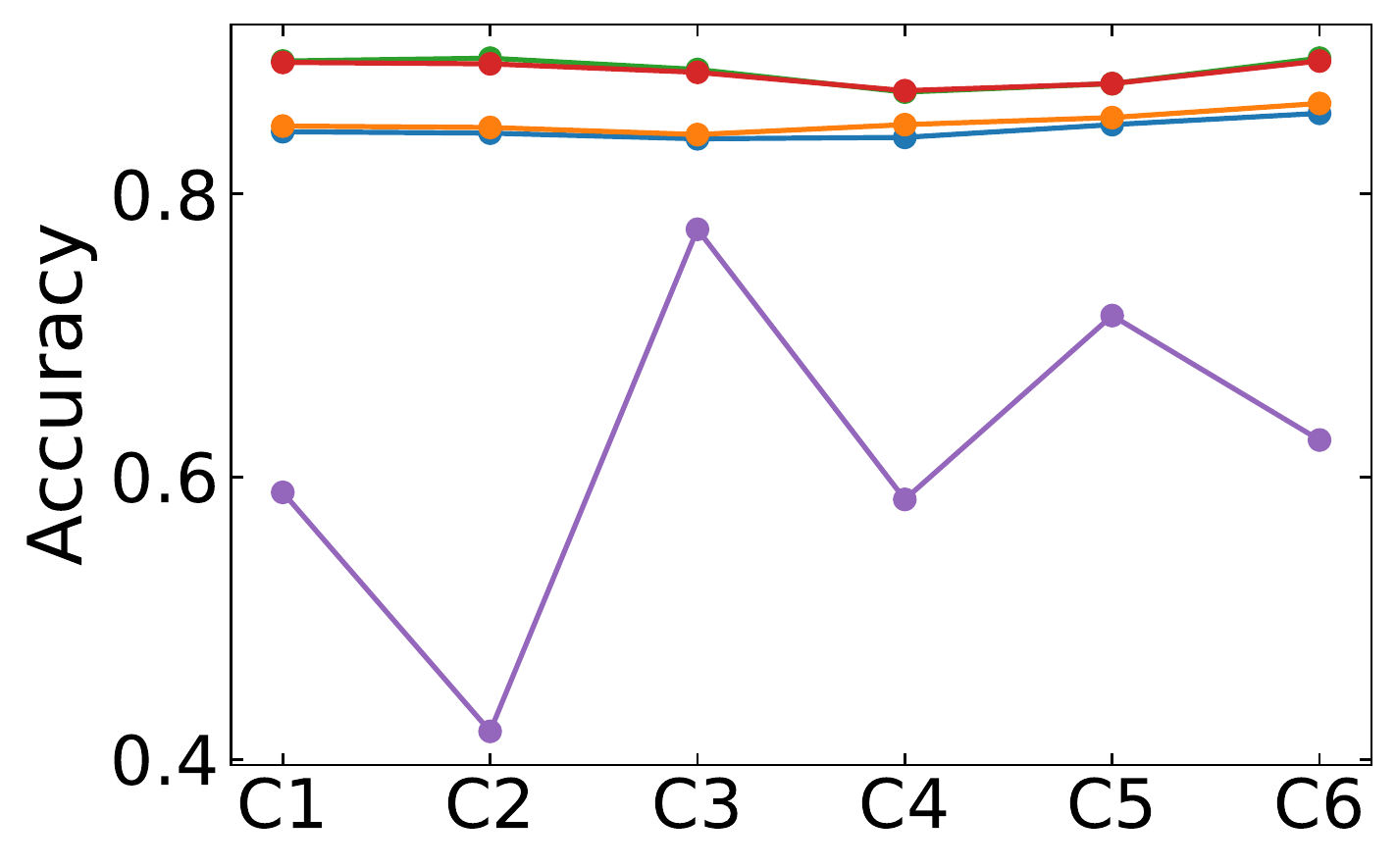}
    }
    \subfloat[Wiki]{
        \includegraphics[width=.24\textwidth]{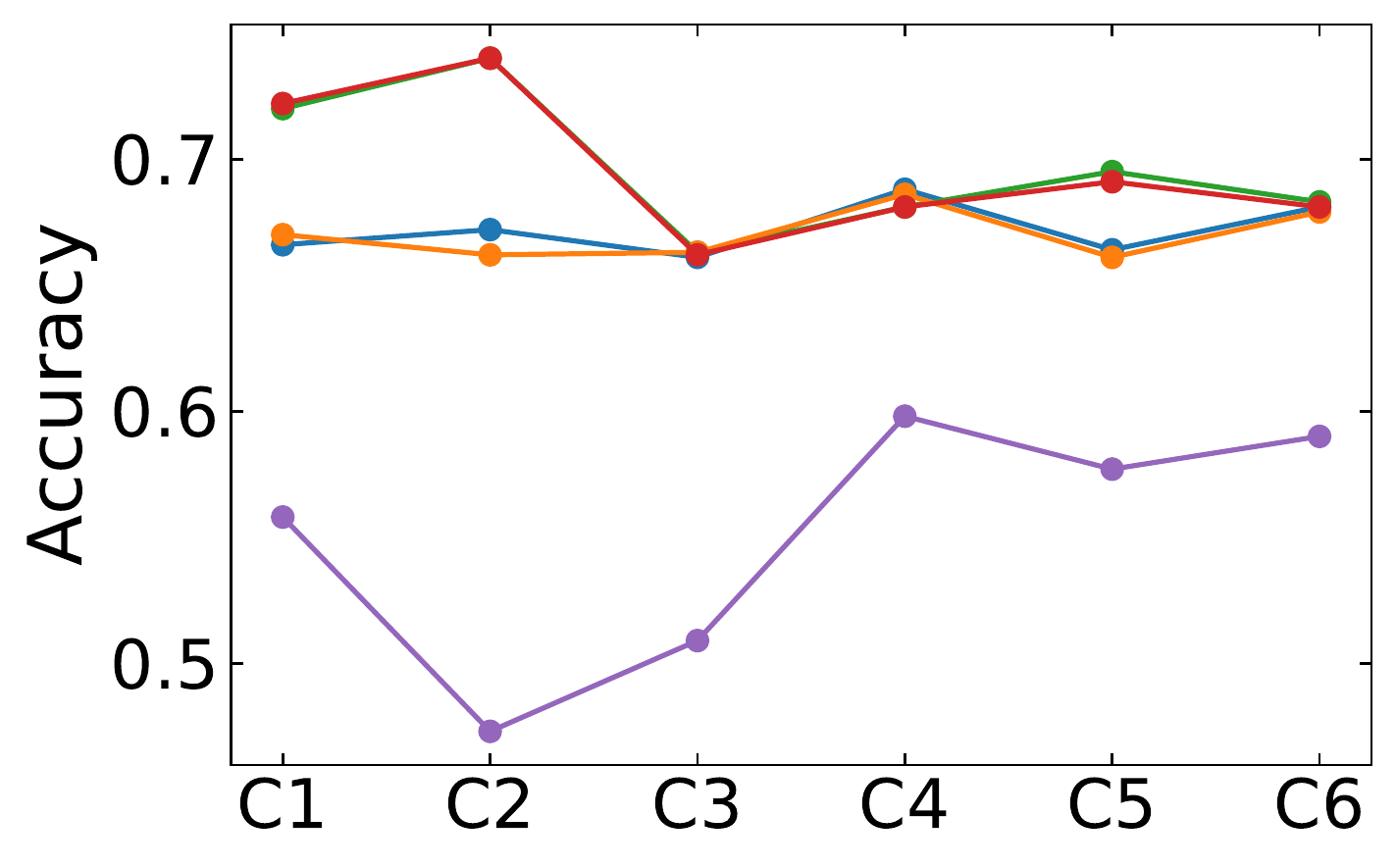}
    }
    \caption{Node classification accuracy of each client (C1-C6) on different datasets.}
    \label{fig:local_fixed_acc}
\end{figure*}

\subsection{Q3: Experimental Results under Local Goal}
\subsubsection{Graph Data Distribution of Clients}
In order to more intuitively understand the experimental results under the local goal, we visualize the number of nodes and the proportion of training labels for 6 clients on 4 datasets in Fig. \ref{fig:client_data}. It can be seen that the proportion of training labels of each client is quite different. For example, in Cora, client 2 has the highest proportion of training labels, while client 3 and client 4 with the same number of nodes have different proportions of training labels. Fig. \ref{fig:client_degree} shows the node degree distribution for 6 clients on 4 datasets. As can be seen, the node degree distribution of each client is roughly similar, because they are all randomly sampled from the original graph data, but there are also certain differences. For example, in all datasets, the node degree distribution of client 1 is obviously different from other clients. It has more 0-degree nodes and 1-degree nodes. In other words, it has more isolated nodes and nodes with only one edge, which indicates the graph structure of client 1 is very poor. In short, the graph data between clients in the experiment are Non-IID, and there exists highly \textit{heterogeneity} and \textit{complementarity}.

\subsubsection{Performance Comparison}
Fig. \ref{fig:local_fixed_acc} shows the node classification accuracy under local goal. FedGL significantly outperforms Local, and is also distinctly superior to Federated. Further analysis, we have the following observations.
\begin{itemize}
  \item Compared with Local, FedGL outperforms it on each client and each dataset, and the performance on each client is not much different and relatively stable, even though clients have Non-IID graph data. There are two main reasons. (1) Federated learning effectively cooperates with the data of each client for training, so that the learned global model performs better and more stable than the model that only uses the data of each client for training. (2) The proposed global self-supervision module discovers the useful information between clients and transmits it to each client through the server, thereby further improving the performance of the global model.
  \item Compared with Federated, FedGL introduces a global self-supervision module, which uses the global pseudo label to enrich the training labels and the global pseudo graph to complement the graph structure, directly improving the quality of each local model and leading to a high-quality global model. Therefore, FedGL can also perform better under the local goal.
\end{itemize}

\begin{table}[t]
    \centering
    \caption{Node classification accuracy under different client participation ratio per round.}
    \begin{tabular}{c|c|cccc}
    \toprule
    Dataset          & Method       & 30\%           & 50\%           & 70\%           & 90\%           \\
    \midrule
    \multirow{2}{*}{Cora}     & Federated & 0.798          & 0.798          & 0.815          & 0.814          \\ 
                              & FedGL      & \textbf{0.822} & \textbf{0.826} & \textbf{0.826} & \textbf{0.828} \\ \hline
    \multirow{2}{*}{Citeseer} & Federated & 0.651          & 0.674          & 0.686          & 0.684          \\ 
                              & FedGL      & \textbf{0.736} & \textbf{0.736} & \textbf{0.738} & \textbf{0.738} \\ \hline
    \multirow{2}{*}{ACM}      & Federated & 0.777          & 0.841          & 0.840          & 0.844          \\ 
                              & FedGL      & \textbf{0.862} & \textbf{0.886} & \textbf{0.868} & \textbf{0.888} \\ \hline
    \multirow{2}{*}{Wiki}     & Federated & 0.668          & 0.684          & 0.686          & 0.686          \\ 
                              & FedGL      & \textbf{0.696} & \textbf{0.695} & \textbf{0.698 }& \textbf{0.696} \\
    \bottomrule
    \end{tabular}
    \label{tab_client_ratio_per_round_acc}
\end{table}

\subsection{Q4: Different Settings for Federated Learning}
\subsubsection{Client Participation Ratio per Round}
In the above experiment, we use the settings in Section \ref{exp_set} by default. The client participation ratio per round is set to 100\%, i.e., all clients participate in federated training in each round. In real scenarios, due to the large number of clients, or the differences in computing power and network bandwidth of each client, it is time-consuming if all clients are required to participate in each round. Therefore, in this experiment, we randomly select 30\%, 50\%, 70\%, and 90\% of the clients from 6 clients (i.e., 2, 3, 4, 5 clients) to participate in federated training in each round. The experimental results are reported in Table \ref{tab_client_ratio_per_round_acc}. There are two observations. (1) Compared to Table \ref{tab_global_fixed_acc}, FedGL has almost no performance decline under different client participation ratios, and the accuracy even has a slight improvement on Citeseer. It indicates that FedGL can maintain training accuracy while ensuring training speed, and thus can be readily applied to real scenarios. (2) FedGL is still better than Federated, which manifests that the useful global self-supervision information can still be discovered to improve the quality of the global model, although the prediction results and node embeddings uploaded in each round are reduced.

\subsubsection{Number of Clients and Data Size} In this experiment, we aim to explore the impact of the number of clients and data size for federated learning. We change the sampling proportion of each client from [30\%,40\%,50\%,50\%,60\%,70\%] to [30\%,40\%,50\%], [20\%,40\%,60\%], [50\%,60\%,70\%,80\%], [20\%,40\%,50\%,70\%,70\%,90\%] with other parameters unchanged. The experimental results are reported in Table \ref{tab_client_data_sampling_rate_acc}. There are three observations. (1) Under 4 groups of the different number of clients and data size, FedGL and Federated both achieve the best results in [50\%,60\%,70\%,80\%]. Such an observation shows that the data size of clients is more important than the number of clients. Especially when the data of each client are relatively large and there is no magnitude difference in the data size between clients, a higher-quality global model could be learned. (2) FedFL still outperforms Federated in most case, especially under [20\%, 40\%, 50\%, 70\%, 70\%, 90\%], which shows that the proposed global self-supervision module has effectively alleviated the \textit{heterogeneity} of graph data between clients and learned a high-quality global model under relatively severe Non-IID situation. (3)  Compared to Table \ref{tab_global_fixed_acc}, FedGL can still achieve promising results and maintains its superiority although clients have the different number of clients and data size.

\begin{table*}[t]
    \centering
    \caption{Node classification accuracy under different number of clients and data size.}
    \begin{tabular}{c|c|cccc}
    \toprule
    Dataset                   & Method    & [30\%,40\%,50\%]  & [20\%,40\%,60\%]  & [50\%,60\%,70\%,80\%]   & [20\%,40\%,50\%,70\%,70\%,90\%] \\
    \midrule
    \multirow{2}{*}{Cora}     & Federated & 0.778          & 0.794          & 0.808          & 0.810          \\ 
                              & FedGL      & \textbf{0.816} & \textbf{0.817} & \textbf{0.838} & \textbf{0.824} \\ \hline
    \multirow{2}{*}{Citeseer} & Federated & 0.690          & 0.692          & 0.706          & 0.702          \\ 
                              & FedGL      & \textbf{0.730} & \textbf{0.692} & \textbf{0.741} & \textbf{0.747} \\ \hline
    \multirow{2}{*}{ACM}      & Federated & \textbf{0.718} & \textbf{0.758} & 0.875          & 0.849          \\ 
                              & FedGL      & 0.609          & 0.625          & \textbf{0.888} & \textbf{0.885} \\ \hline
    \multirow{2}{*}{Wiki}     & Federated & 0.666          & 0.657          & 0.686          & 0.682            \\ 
                              & FedGL      & \textbf{0.671} & \textbf{0.684} & \textbf{0.701} & \textbf{0.694}  \\
    \bottomrule
    \end{tabular}
    \label{tab_client_data_sampling_rate_acc}
\end{table*}

\begin{table*}[t]
    \caption{Node classification accuracy under different overlapping node ratio between clients.}
    \centering
    \begin{tabular}{c|c|c|c|ccc}
    \toprule
    Dataset                   & Overlapping ratio  & Centralized & Federated & FedGL w/o GPG   & FedGL w/o GPL   & FedGL           \\
    \midrule
    \multirow{3}{*}{Cora}     & 5\%  & 0.797       & 0.800     & \textbf{0.829} & 0.795          & 0.828          \\ 
                              & 10\% & 0.800       & 0.802     & \textbf{0.828} & 0.805          & 0.817          \\ 
                              & 15\% & 0.794       & 0.790     & \textbf{0.820} & 0.794          & 0.818          \\ \hline
    \multirow{3}{*}{Citeseer} & 5\%  & 0.696       & 0.692     & 0.730          & 0.699          & \textbf{0.730} \\ 
                              & 10\% & 0.681       & 0.670     & 0.632          & 0.676          & \textbf{0.720} \\ 
                              & 15\% & 0.699       & 0.700     & 0.733          & 0.700          & \textbf{0.736} \\ \hline
    \multirow{3}{*}{ACM}      & 5\%  & 0.870       & 0.870     & 0.870          & 0.870          & \textbf{0.872} \\ 
                              & 10\% & 0.716       & 0.708     & \textbf{0.717} & 0.708          & 0.712          \\ 
                              & 15\% & 0.890       & 0.891     & 0.891          & 0.892          & \textbf{0.893} \\ \hline
    \multirow{3}{*}{Wiki}     & 5\%  & 0.691       & 0.690     & 0.694          & 0.690          & \textbf{0.695} \\ 
                              & 10\% & 0.664       & 0.659     & 0.666          & \textbf{0.668} & 0.664          \\ 
                              & 15\% & 0.680       & 0.675     & 0.677          & \textbf{0.683} & 0.674          \\
    \bottomrule
    \end{tabular}
    \label{tab_overlap_acc}
\end{table*}

\subsubsection{Overlapping Node Ratio between Clients} In this experiment, we aim to explore the impact of the overlapping node ratio of the graph data between clients for federated learning. We keep the sampling proportion of the 6 clients unchanged but control their overlapping node ratios to 5\%, 10\%, 15\%, and repeat the node classification experiment. The experimental results are reported in Table \ref{tab_overlap_acc}. There are two observations. (1) FedGL still outperforms Centralized and Federated, which proves the effectiveness and stability of the proposed global self-supervision. Besides, the two ablation versions of FedGL have achieved the best results in different datasets, which manifests that the global pseudo label and global pseudo graph also work well when used alone. (2) Compared to Table \ref{tab_global_fixed_acc}, the overlapping node ratio is about 1\%. This experiment is 5\%, 10\%, and 15\%. As can be seen, the performance is not positively correlated with the overlapping node ratio. Because the \textit{heterogeneity} and \textit{complementary} are opposite to each other to some extent. On the one hand, it is necessary to alleviate the \textit{heterogeneity}. On the other hand, it is necessary to utilize the \textit{complementarity}. This is exactly what FedGL focuses on and solves.

\begin{figure}[t]
    \centering
    \subfloat[Cora]{
        \includegraphics[width=.24\textwidth]{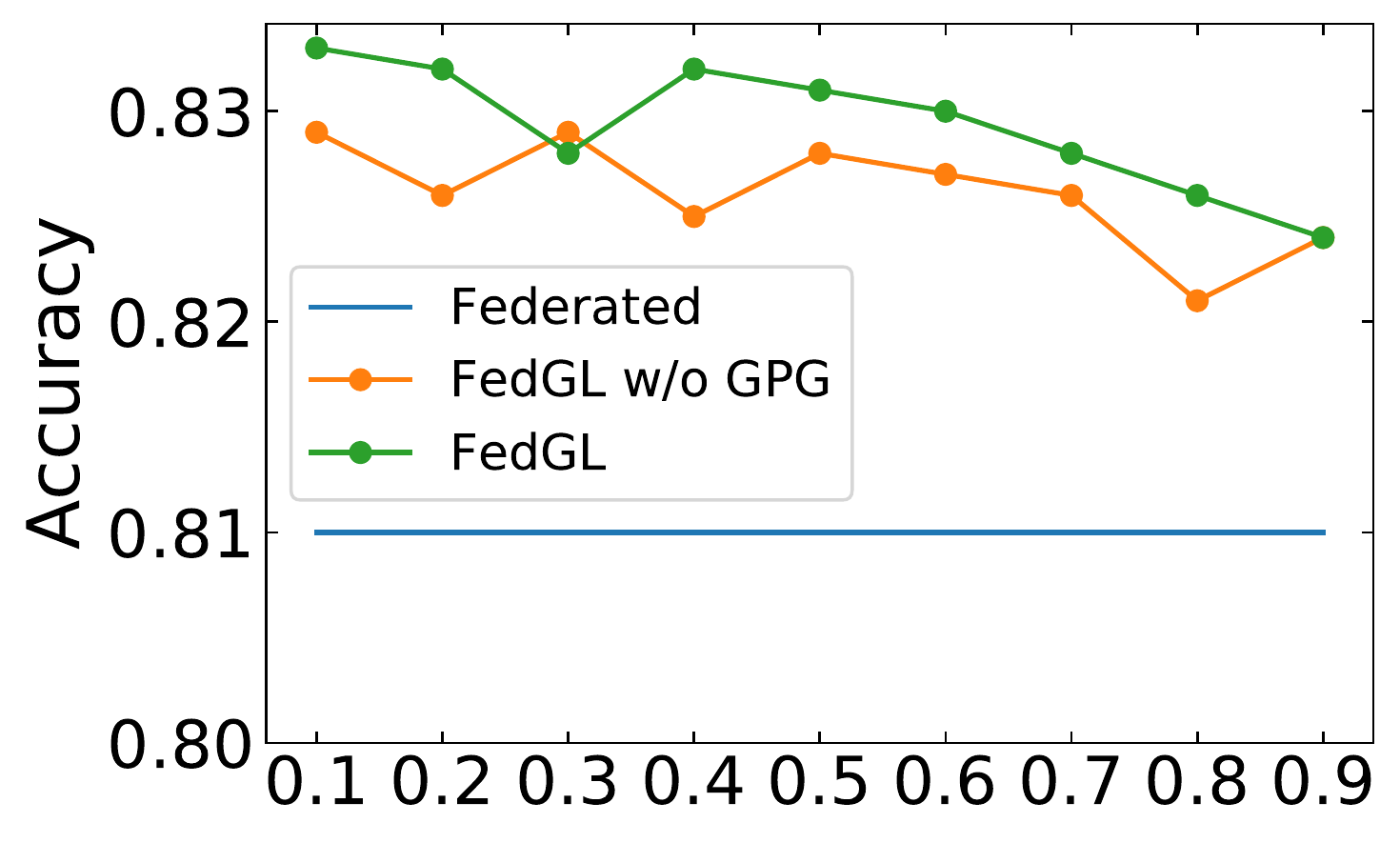}
    }
    \subfloat[Citeseer]{
        \includegraphics[width=.24\textwidth]{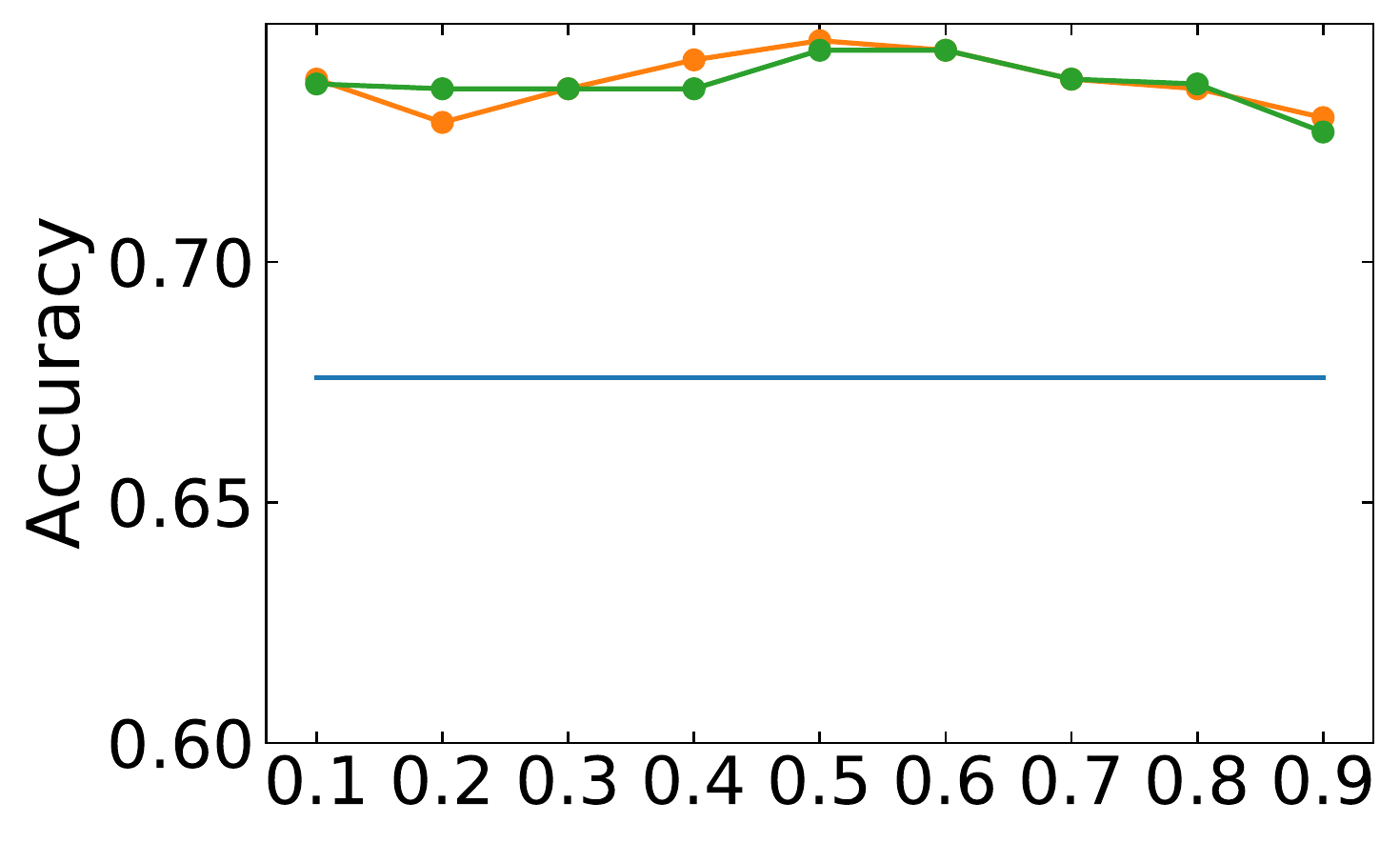}
    }
    \subfloat[ACM]{
        \includegraphics[width=.24\textwidth]{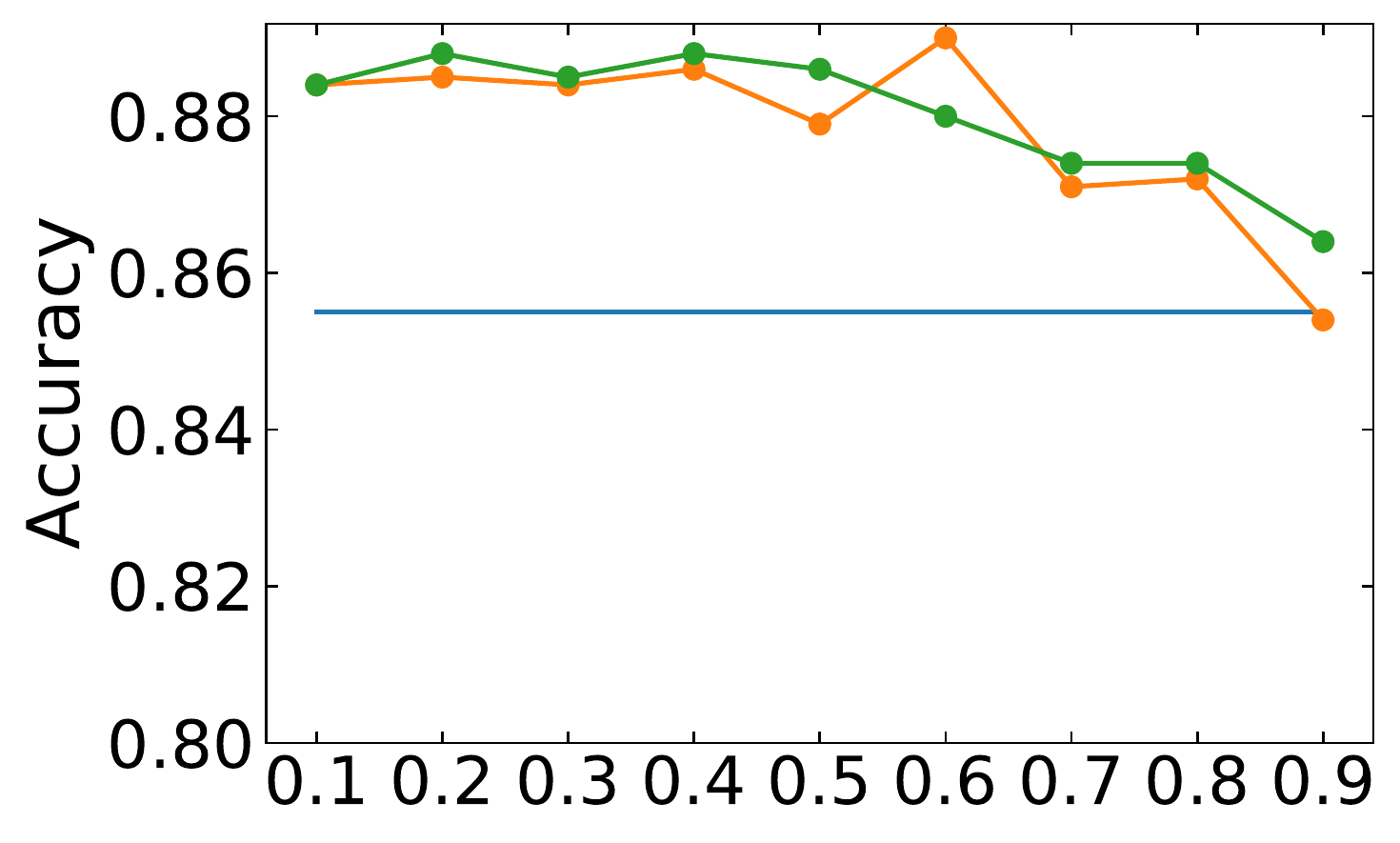}
    }
    \subfloat[Wiki]{
        \includegraphics[width=.24\textwidth]{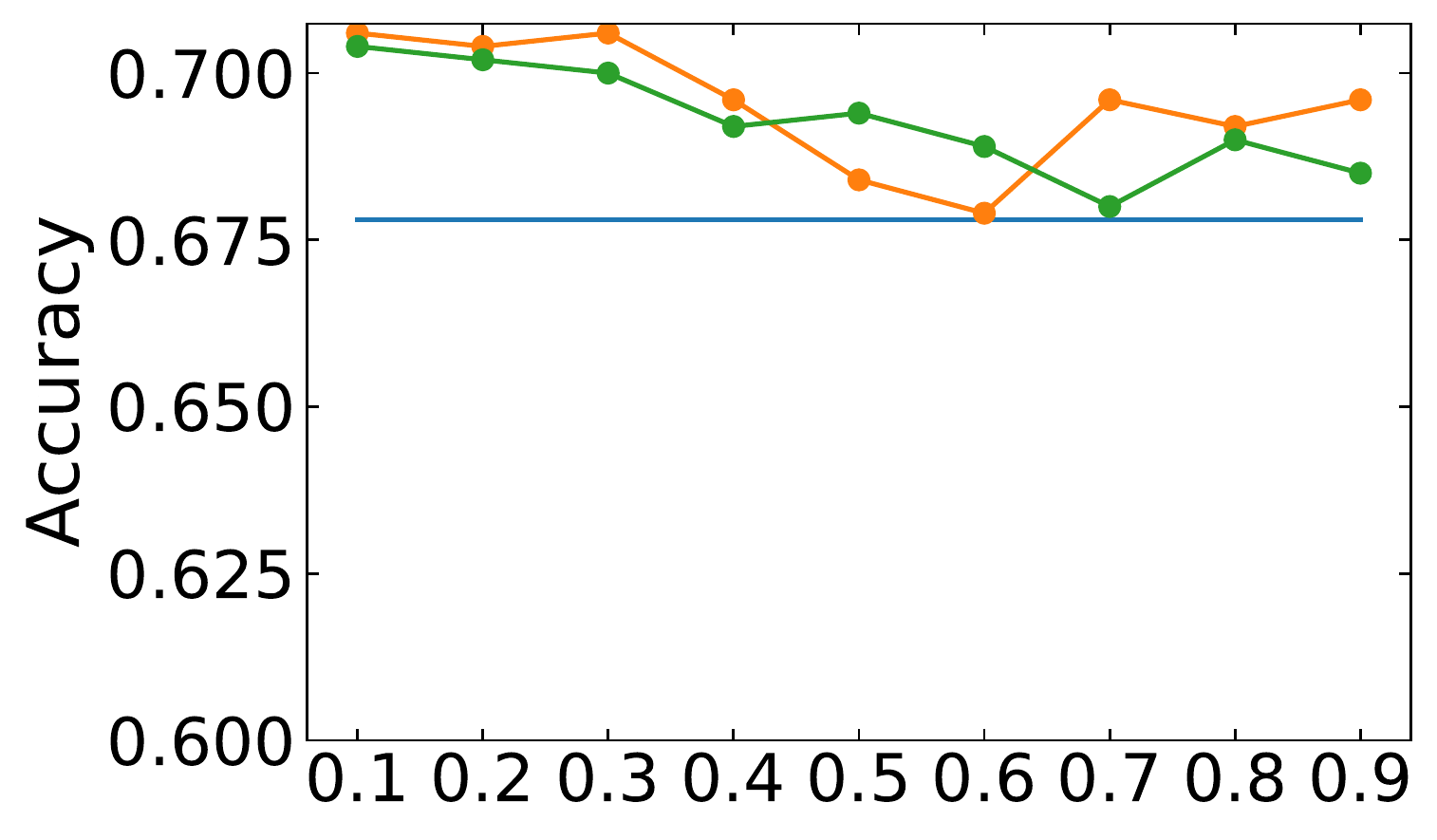}
    }
    \caption{Node classification accuracy under different confidence threshold $\lambda$.}
    \label{fig:probability_threshold_acc}
\end{figure}

\begin{figure}[t]
    \centering
    \subfloat[Cora]{
        \includegraphics[width=.24\textwidth]{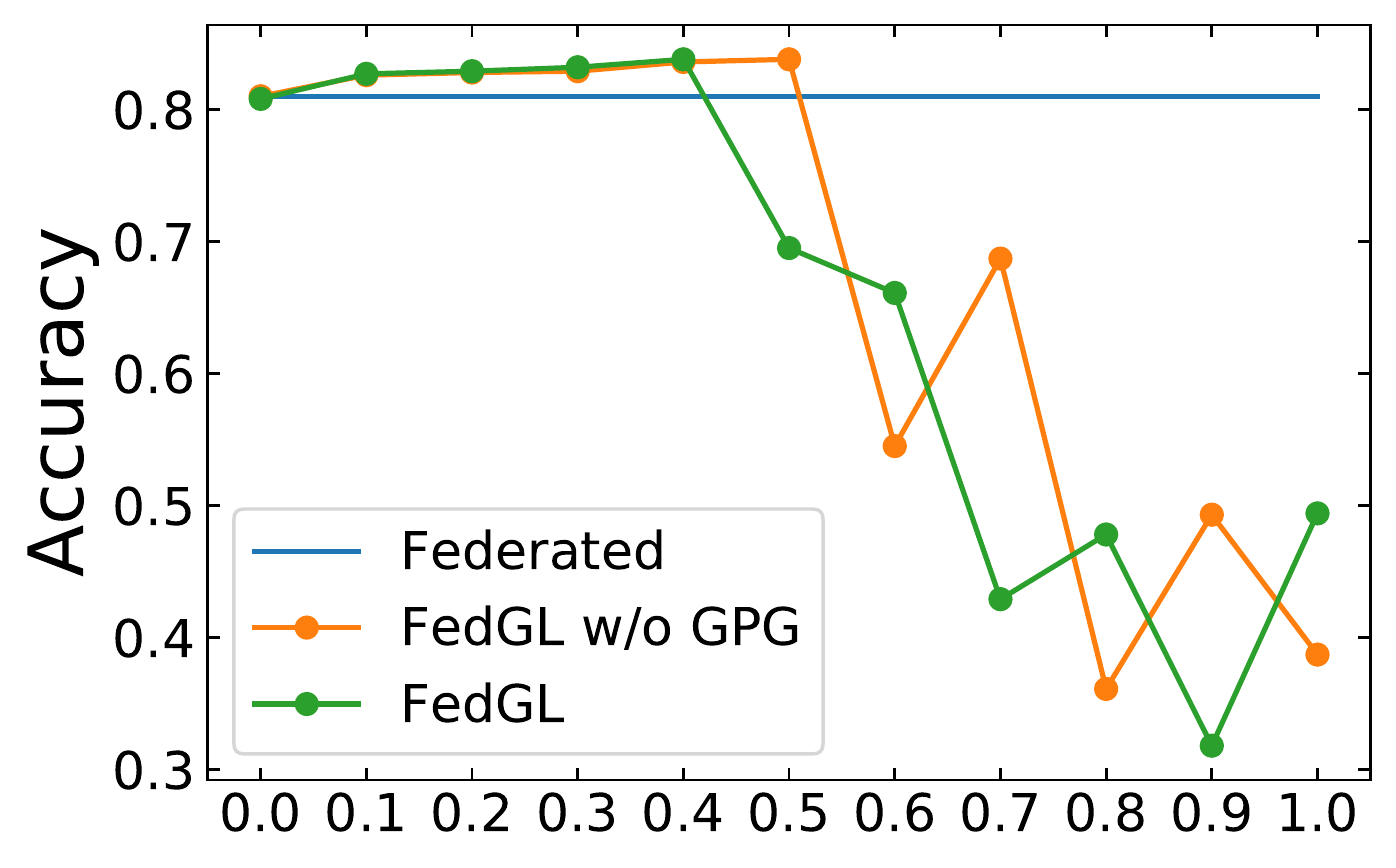}
    }
    \subfloat[Citeseer]{
        \includegraphics[width=.24\textwidth]{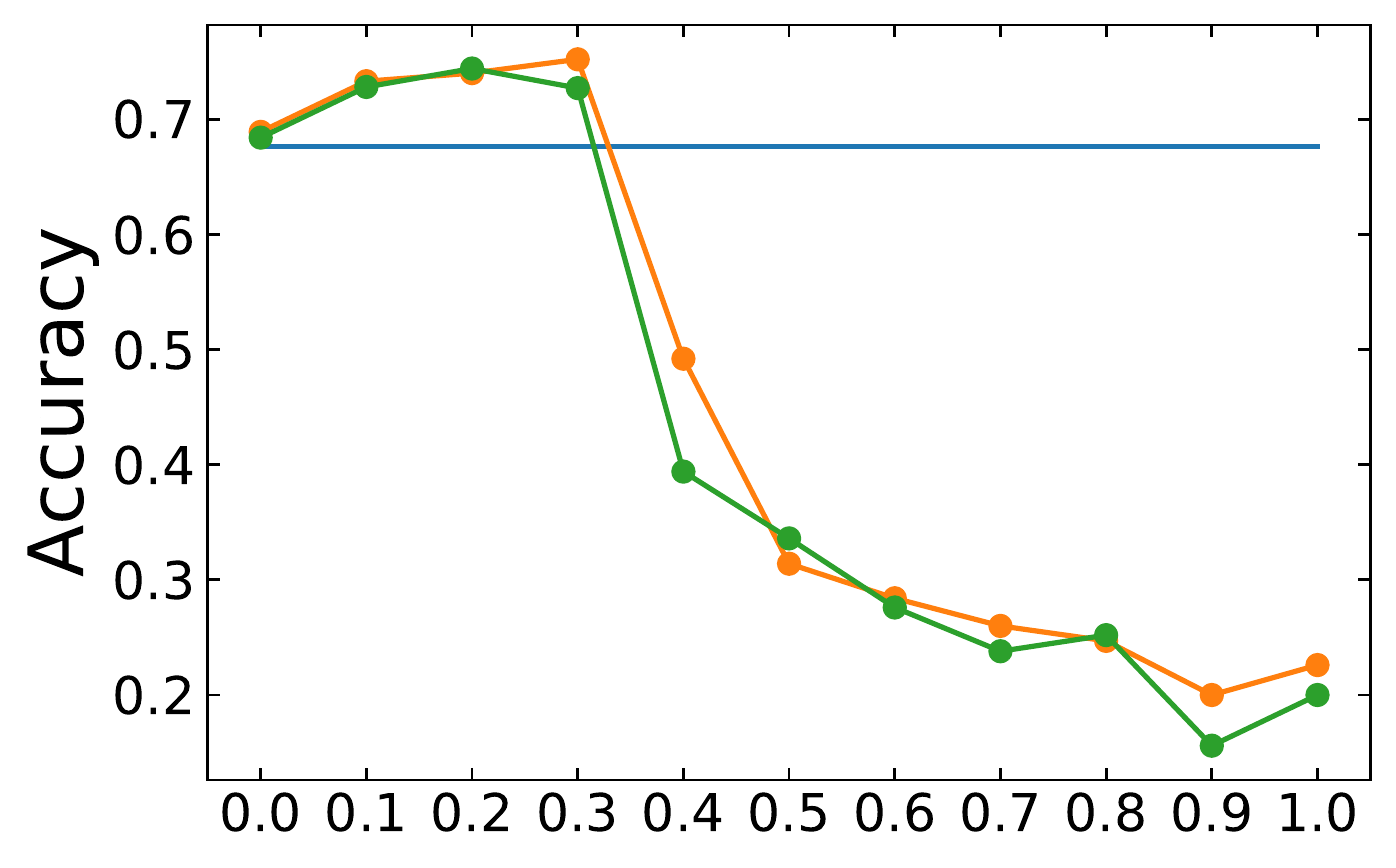}
    }
    \subfloat[ACM]{
        \includegraphics[width=.24\textwidth]{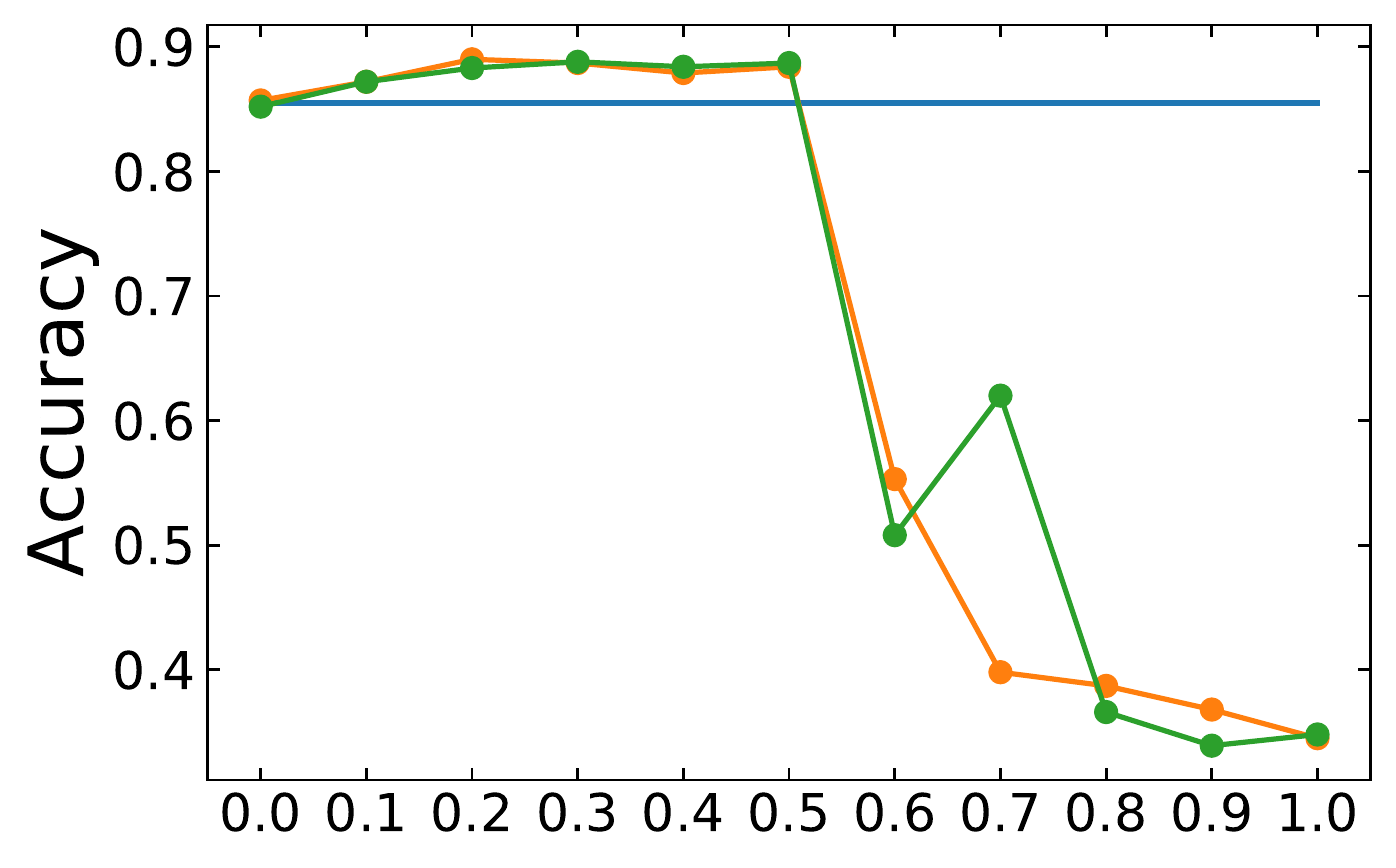}
    }
    \subfloat[Wiki]{
        \includegraphics[width=.24\textwidth]{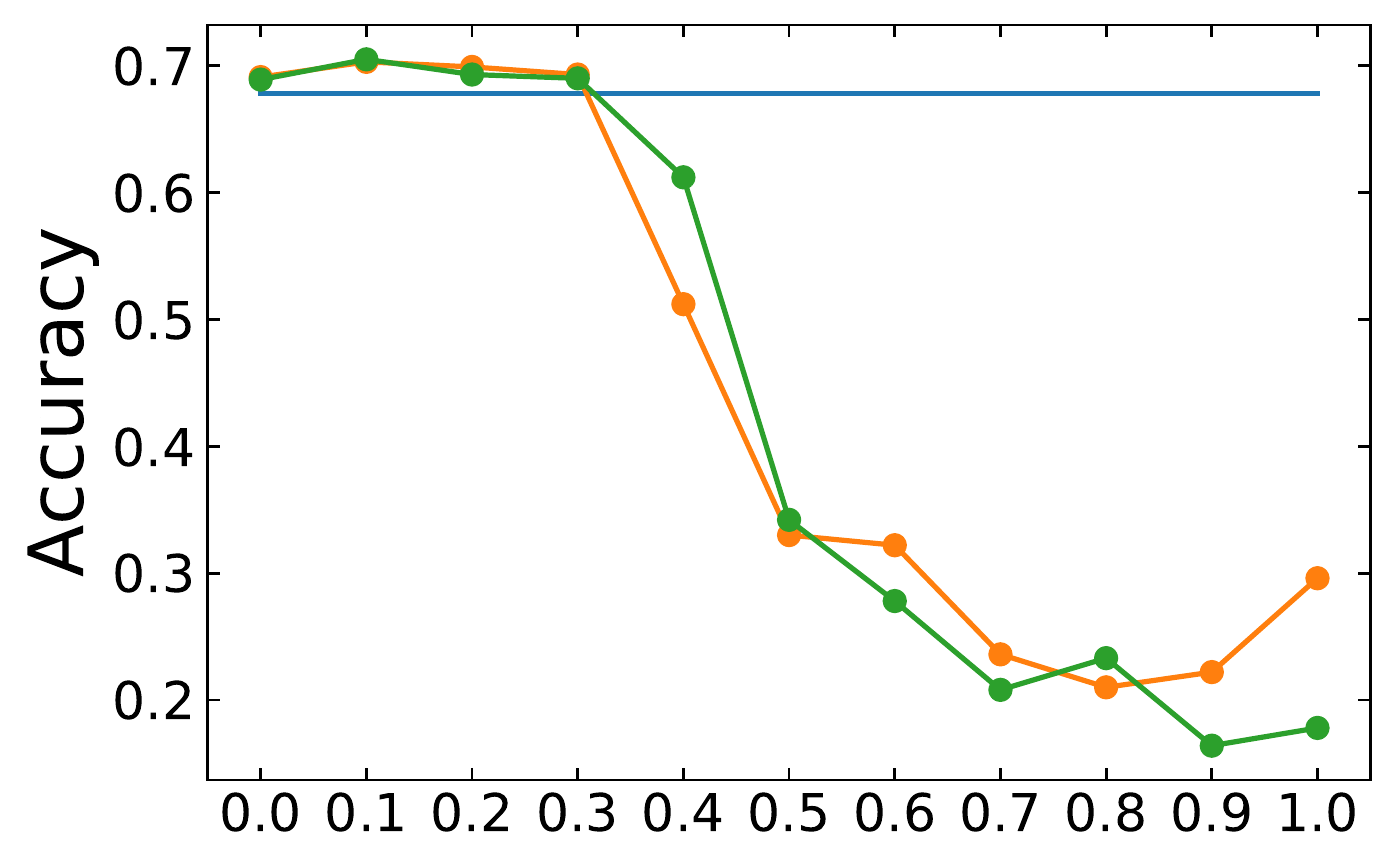}
    }
    \caption{Node classification accuracy under different self-supervision learning coefficient $\alpha$.}
    \label{fig:ssl_loss_weight_acc}
\end{figure}

\subsection{Q5: Parameter Sensitivity Analysis}
\subsubsection{Confidence Threshold} The confidence threshold $\lambda$ in Eq. \eqref{pseudo_label} is used to control the number of pseudo labels. To explore how $\lambda$ affects the performance of FedGL, we tune $\lambda$ in [0.1, 0.9] with step size 0.1. The results are shown in Fig. \ref{fig:probability_threshold_acc}. On all datasets, FedGL and FedGL w/o GPG are significantly better than Federated under various $\lambda$, which shows the effectiveness and stability of the proposed global pseudo label. Especially on Citeseer, the classification accuracy of the global model can be improved by at least 5\% as long as the global pseudo label is utilized. For different datasets, FedGL achieves the best performance with different $\lambda$, but $\lambda$ is overall small. From the results, [0.1, 0.3] is a desirable interval.

\subsubsection{Self-supervised Learning Coefficient} The self-supervised learning coefficient $\alpha$ in Eq. \eqref{L_final} is used to control the strength of self-supervised learning. To explore the impact of $\alpha$ on FedGL, we tune $\alpha$ in [0, 1] with step size 0.1. Note that $\alpha=0$ means without using the global pseudo label, and $\alpha=1$ means that the SSL loss is as important as the main task loss. The results are shown in Fig. \ref{fig:ssl_loss_weight_acc}. As the value of $\alpha$ increases, the performance of FedGL and FedGL w/o GPG first increase, and then decrease sharply after exceeding a certain threshold. It is consistent with our analysis that the SSL item plays a supporting role to assist the main task, so $\alpha$ should not be set too large. When $\alpha$ is relatively small, FedGL and FedGL w/o GPG both outperform Federated.

\begin{figure}[t]
    \centering
    \subfloat[Cora]{
        \includegraphics[width=.24\textwidth]{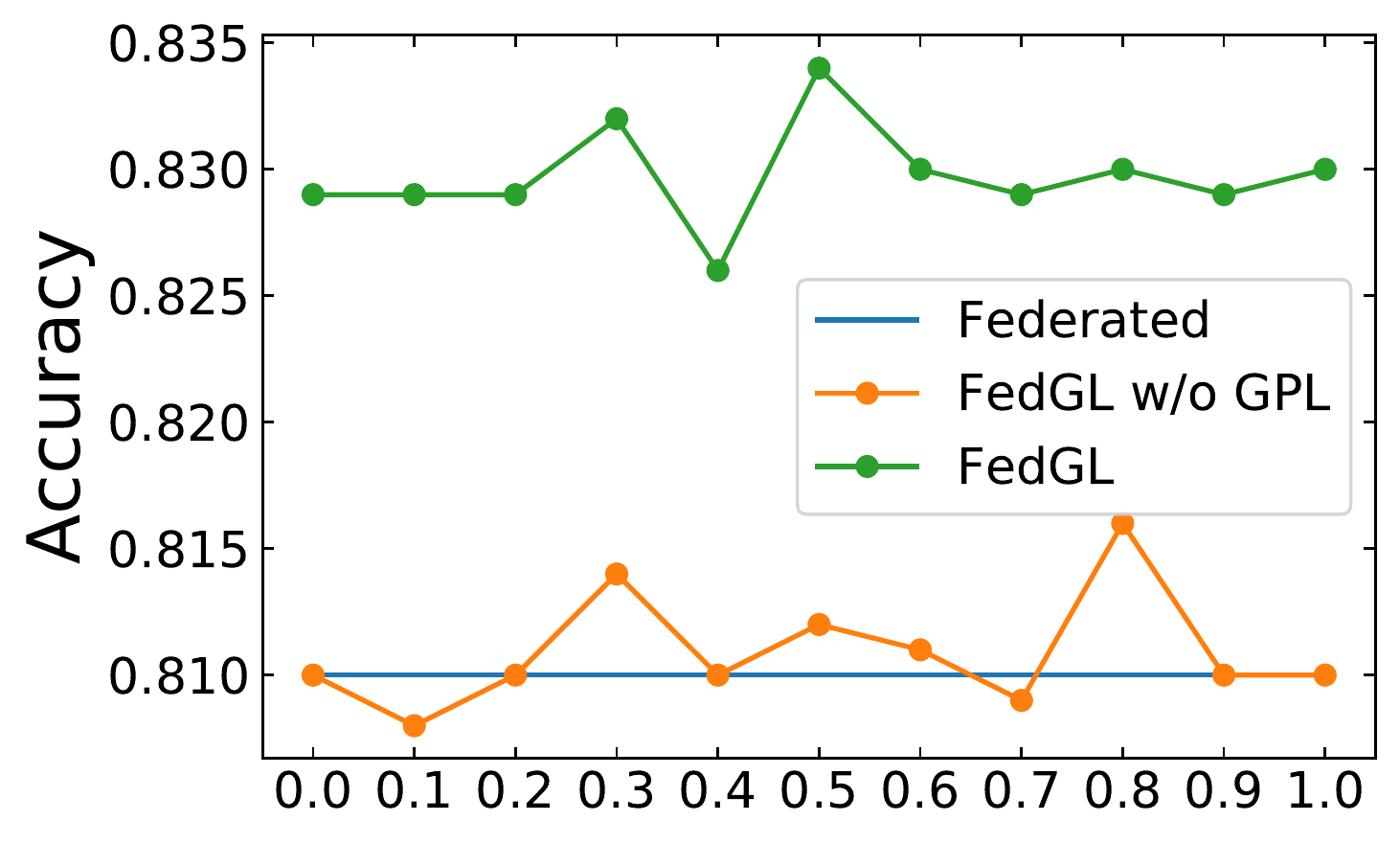}
    }
    \subfloat[Citeseer]{
        \includegraphics[width=.24\textwidth]{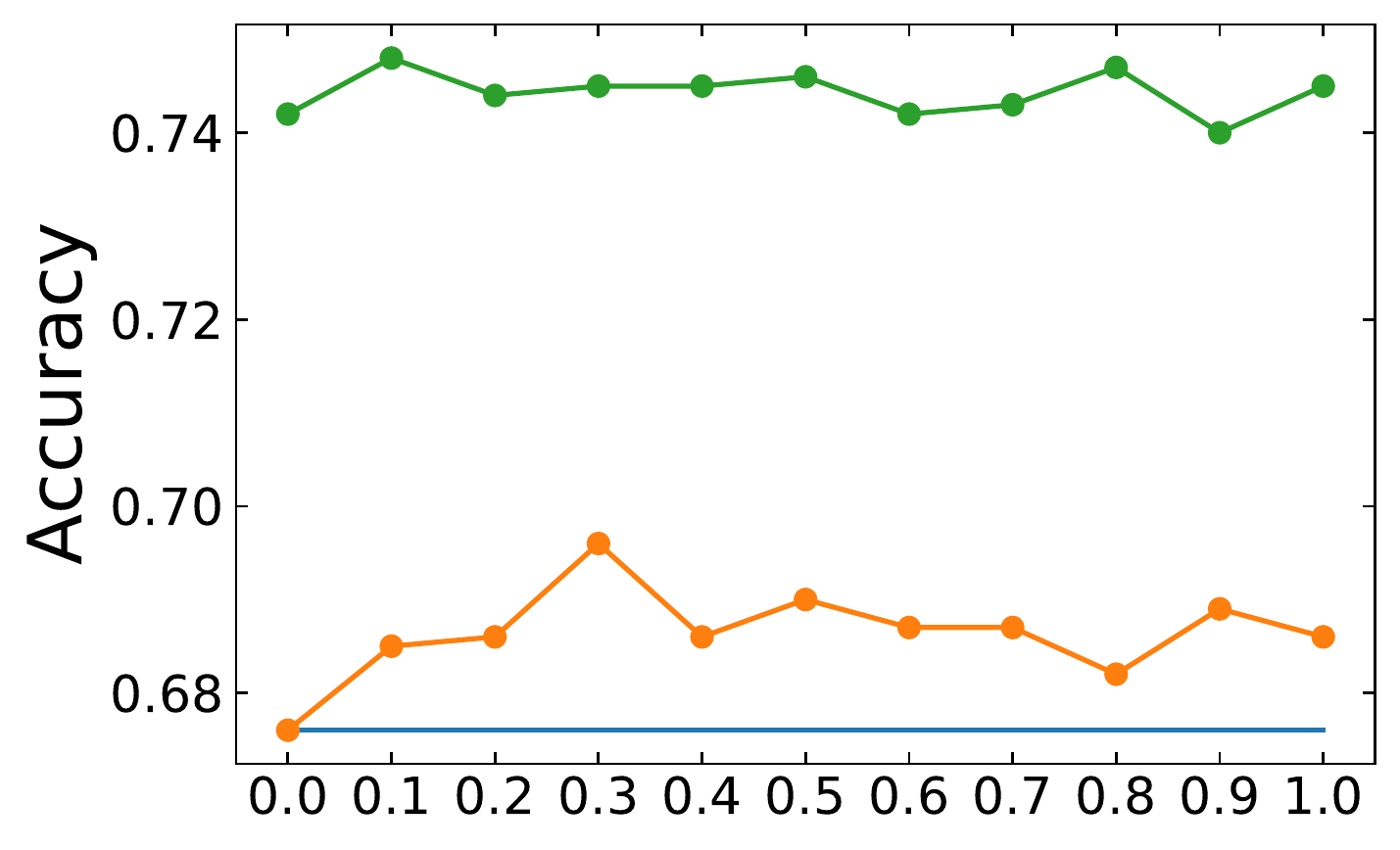}
    }
    \subfloat[ACM]{
        \includegraphics[width=.24\textwidth]{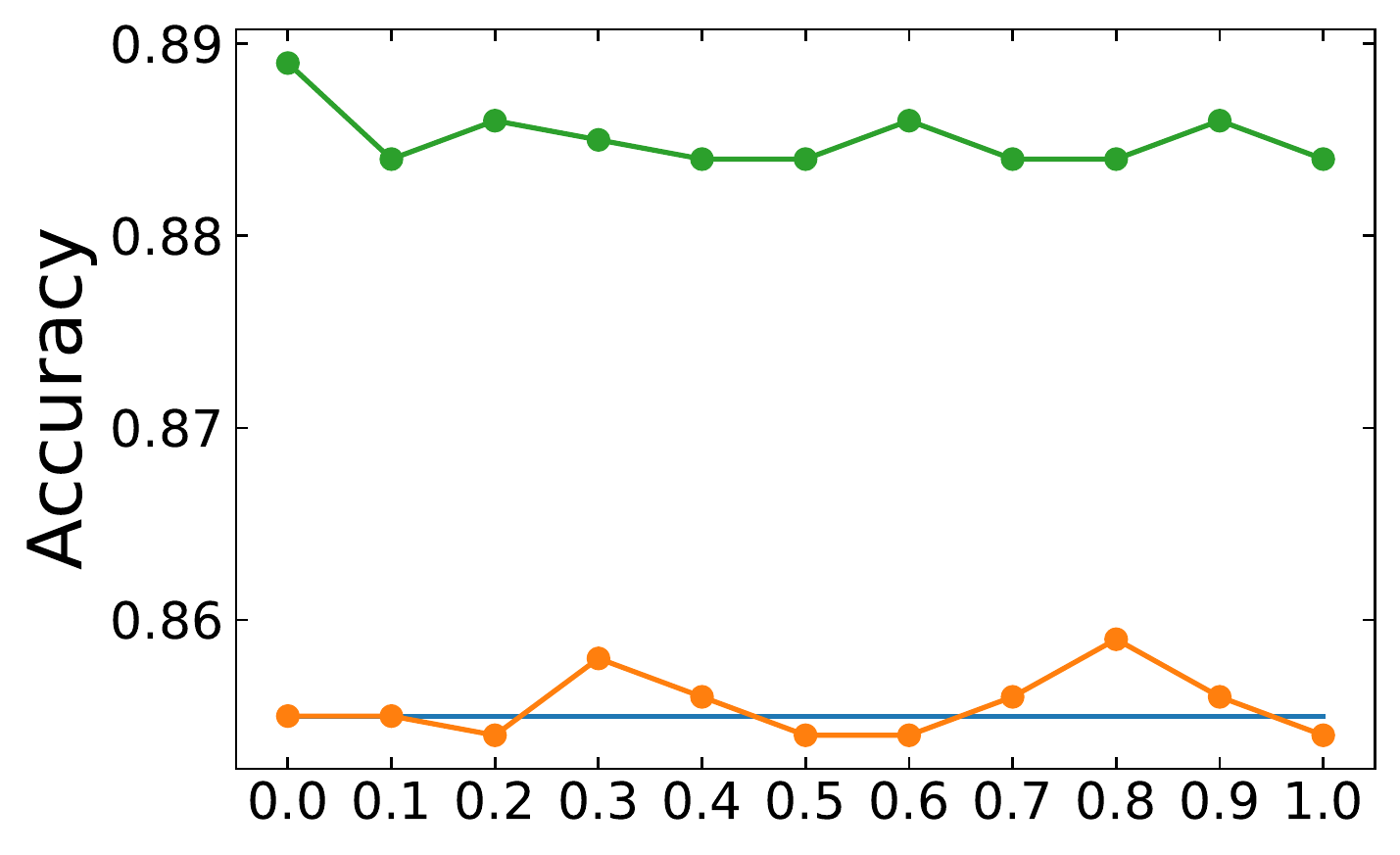}
    }
    \subfloat[Wiki]{
        \includegraphics[width=.24\textwidth]{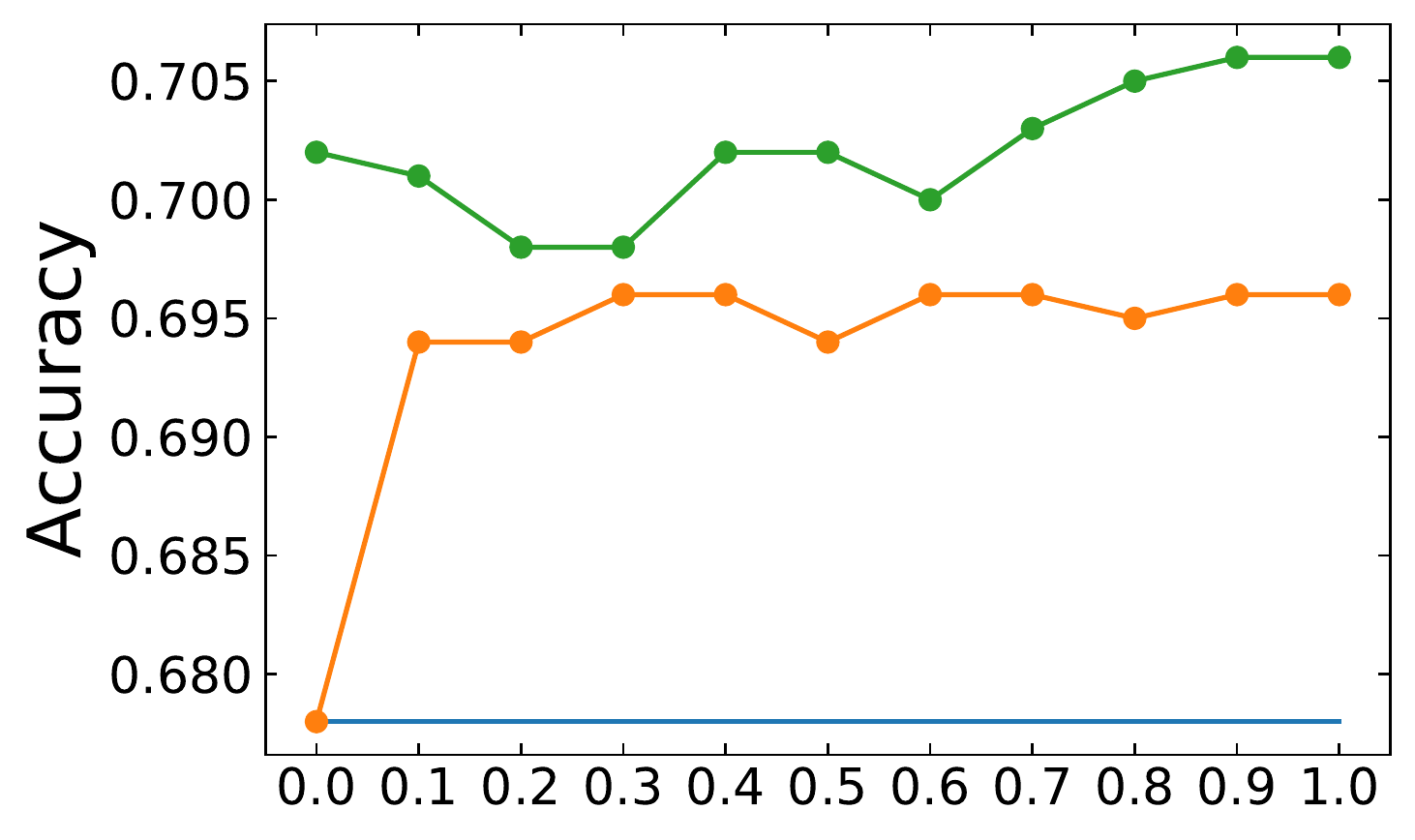}
    }
    \caption{Node classification accuracy under different global pseudo graph coefficient $\beta$.}
    \label{fig:pg_weight_acc}
\end{figure}

\begin{figure*}[t]
    \centering
    \subfloat[Cora]{
        \includegraphics[width=.24\textwidth]{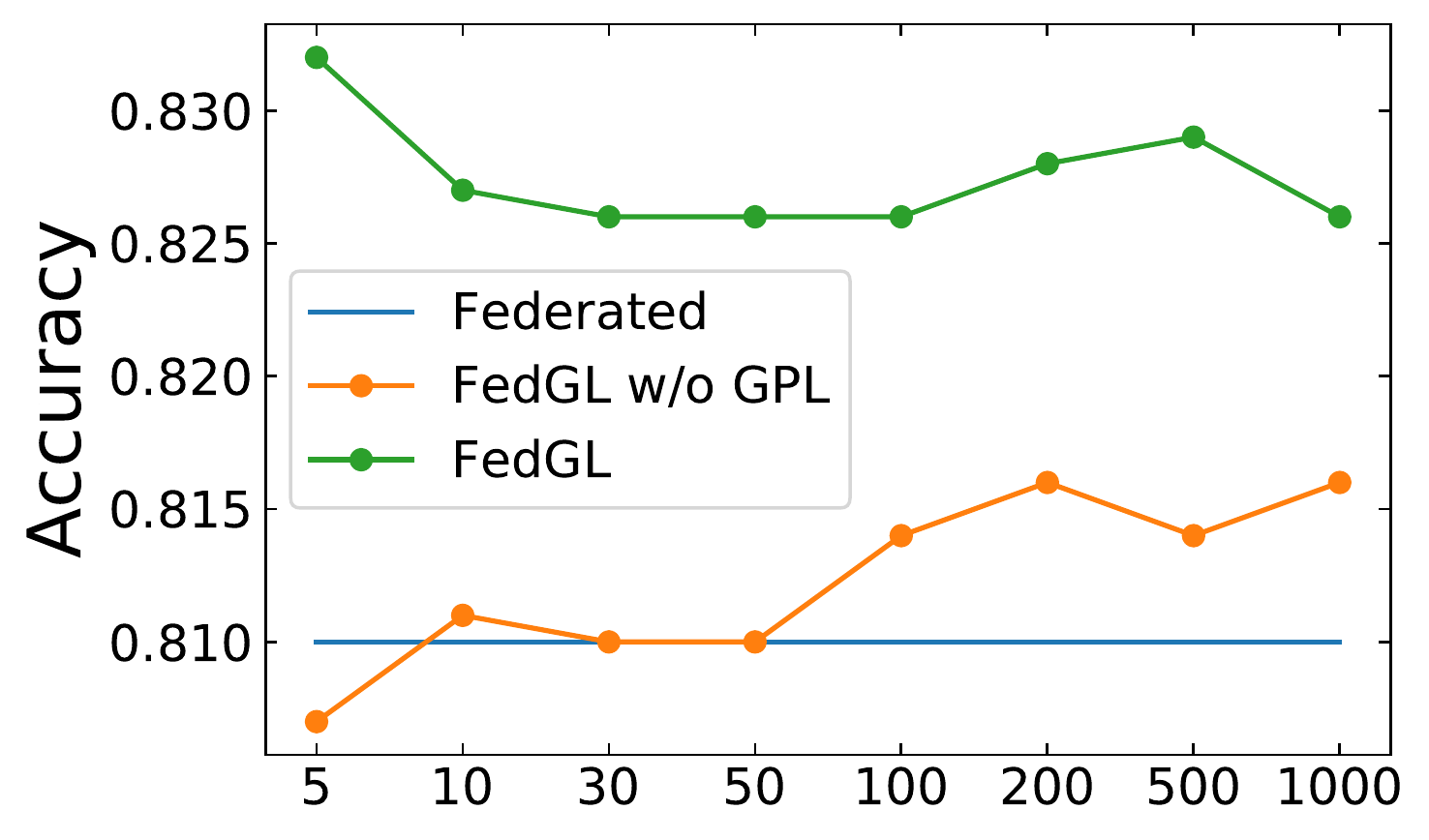}
    }
    \subfloat[Citeseer]{
        \includegraphics[width=.24\textwidth]{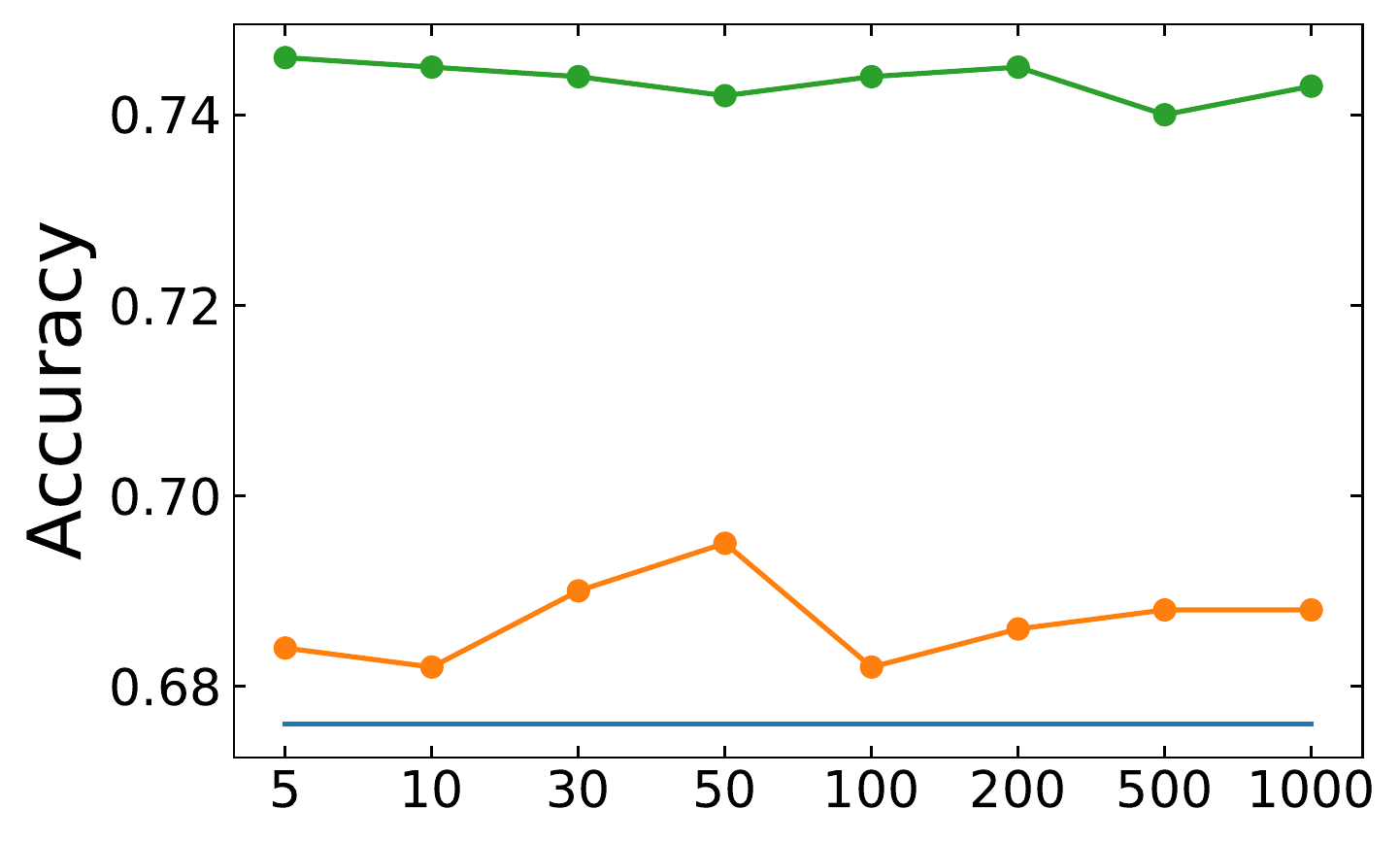}
    }
    \subfloat[ACM]{
        \includegraphics[width=.24\textwidth]{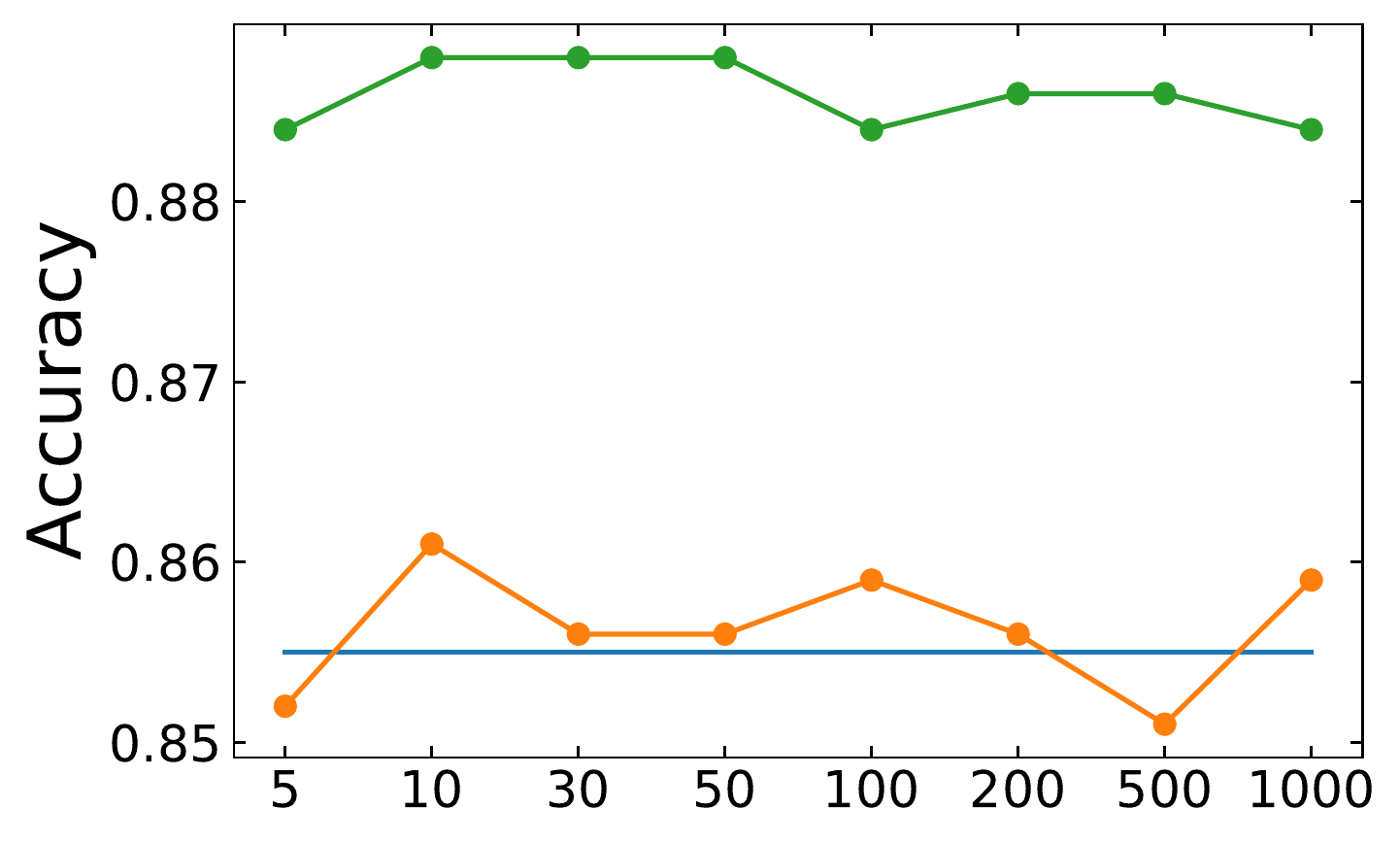}
    }
    \subfloat[Wiki]{
        \includegraphics[width=.24\textwidth]{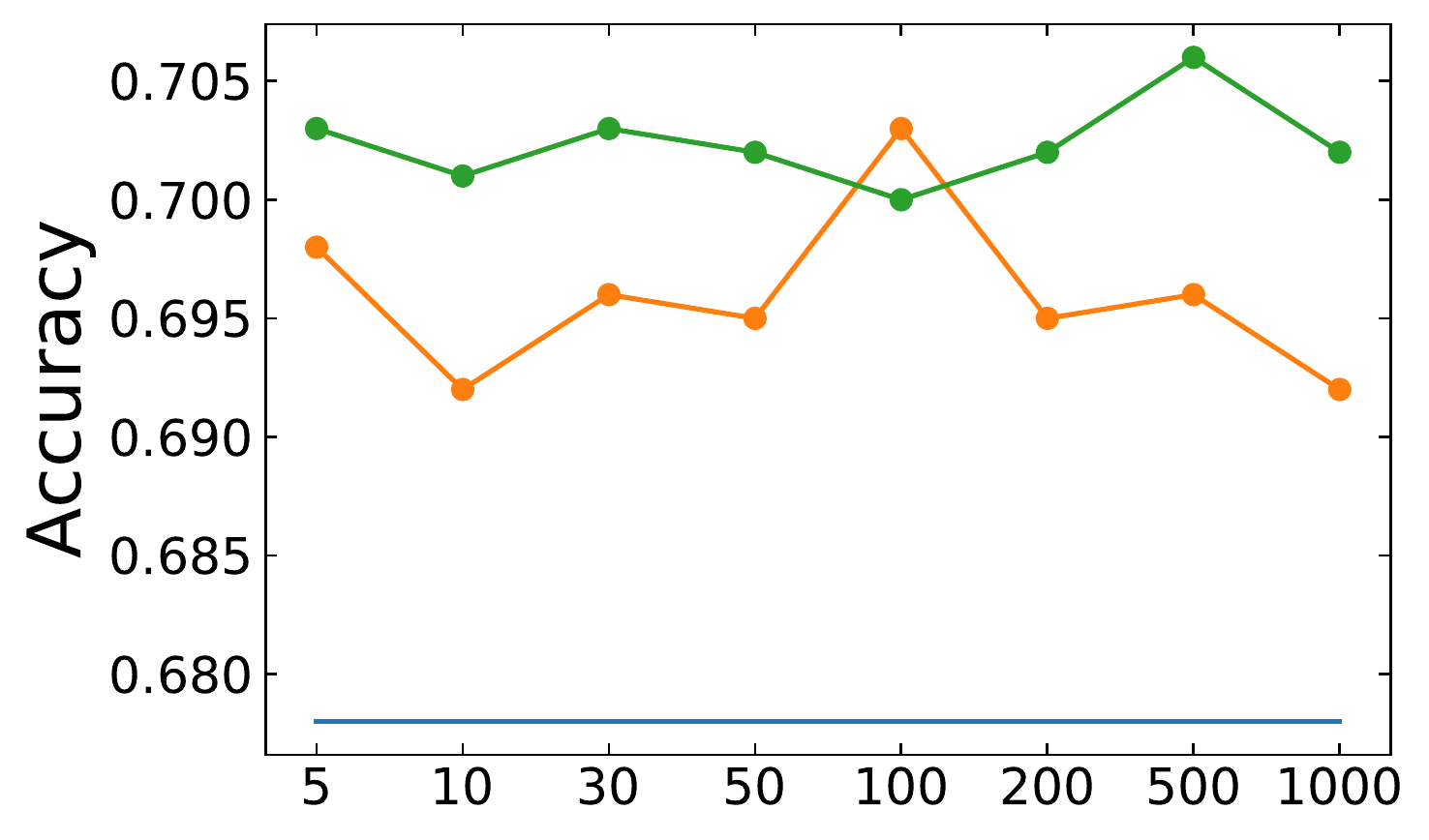}
    }
    \caption{Node classification accuracy under different neighbor number $s$ of global pseudo graph.}
    \label{fig:neighbor_number_acc}
\end{figure*}

\subsubsection{Global Pseudo Graph Coefficient} The global pseudo graph coefficient $\beta$ in Eq. \eqref{update_adj} is used to control the strength of complementing the graph structure. To explore the impact of $\beta$ on FedGL, we tune $\beta$ in [0, 1] with step size 0.1. Note that $\beta=0$ means without using the global pseudo graph, and $\beta=1$ means that the global pseudo graph is as important as the original graph structure. The results are shown in Fig. \ref{fig:pg_weight_acc}. As the value of $\beta$ increases, the performance of FedGL and FedGL w/o GPG have some small fluctuations, but the overall performance is relatively stable and higher than Federated, which shows the effectiveness and stability of the proposed global pseudo graph.

\subsubsection{Neighbor Number of Global Pseudo Graph} The neighbor number $s$ is used to control the sparsity of the global pseudo graph. To explore the impact of $s$ on FedGL, we set $s$ to 5, 10, 30, 50, 100, 200, 500, 1000, respectively. The results are shown in Fig. \ref{fig:neighbor_number_acc}. As the value of $s$ increases, the performance of FedGL and FedGL w/o GPG have some small fluctuations, but the overall performance is relatively stable and higher than Federated, which shows the effectiveness and stability of the proposed global pseudo graph.

\section{Conclusion}\label{Con}
In this paper, we propose a general federated graph learning framework FedGL, which can collaborate the graph data stored in different clients to train a high-quality graph model while protecting data privacy. To tackle the \textit{heterogeneity} and \textit{complementarity} of graph data between clients, we propose to discover and exploit the global self-supervision information. Concretely, clients additionally upload prediction results and node embeddings to the server for discovering global pseudo label and global pseudo graph. Server then distributes them to each client to enrich the training labels and complement the graph structure respectively, thereby improving the quality of each local model and obtaining a high-quality global model. More importantly, the process of global self-supervision discovery and using enables the information of each client to flow and share in a privacy-preserving manner, thus mitigating the \textit{heterogeneity} and utilizing the \textit{complementarity}. Finally, extensive experimental results on the node classification task show that FedGL significantly outperforms the centralized method, simple federated method, and local method, which fully verifies the effectiveness of FedGL.

Recall that FedGL is a general federated graph learning framework, which is not limited to specific graph models. Therefore, in the future, we are interested in exploring the effectiveness of FedGL on more graph models such as GAT \cite{velivckovic2017graph} and FastGCN \cite{chen2018fastgcn}. Besides, we are also interested in extending FedGL to the scenarios where clients use different graph models and even clients have multimodal data.

\bibliographystyle{ACM-Reference-Format}
\bibliography{FedGNN_Reference}

\end{document}